\documentclass{article}

\usepackage{microtype}
\usepackage{graphicx}
\usepackage{subcaption}
\usepackage{booktabs} % for professional tables
\usepackage{xcolor}
\usepackage{lipsum}

\usepackage{hyperref}

\usepackage[preprint]{icml2026}

\usepackage{amsmath}
\usepackage{amssymb}
\usepackage{mathtools}
\usepackage{amsthm}

\usepackage[capitalize,noabbrev]{cleveref}

\theoremstyle{plain}

\theoremstyle{definition}

\theoremstyle{remark}

\usepackage[most]{tcolorbox}
\newtcolorbox{thmbox}{
  colback=blue!5,        % Light blue background
  colframe=blue!75!black,% Dark blue frame
}
\newtcolorbox{defnbox}{
  colback=black!5,        % Light background
  colframe=black!75!black,% Dark frame
}
\definecolor{promptbox}{RGB}{216, 224, 231}

\newtcolorbox[auto counter=promptbox,number within=section]{promptbox}[3]{
  enhanced,
  breakable,
  verbatim,    % <-- works on all Overleaf versions
  attach boxed title to top left={xshift=3mm, yshift=-3mm,yshifttext=-1mm},
  fonttitle=\itshape\color{black},
  boxed title style={boxrule=0.3mm, colback=promptbox!50, colframe=promptbox!60!black},
  title={#2},
  colback=white,
  frame hidden,
  rounded corners,
  arc=3pt,
  fontupper=\ttfamily\linespread{1}\selectfont\small,
  borderline={0.75pt}{0pt}{promptbox!60!black},
  borderline={0.3pt}{2pt}{promptbox!75!black},
  before upper=\ignorespaces #3,
  label=#1,
}

\newtcolorbox{defnboxmaintext}{
  colback=black!5,        % Light background
  colframe=black!75!black,% Dark frame
  left=4pt,right=4pt,top=1pt,bottom=1pt
}

\newtcolorbox{remarkbox}{
  colback=orange!5,        % Light background
  colframe=orange!75!black,% Dark frame
}

\newtcolorbox{remarkboxmaintext}{
    colback=orange!5,        % Light background
  colframe=orange!75!black,% Dark frame
  left=4pt,right=4pt,top=4pt,bottom=4pt
}

\newcommand{\direct}{\textsc{Direct}}
\newcommand{\onef}{\textsc{1F}}
\newcommand{\onet}{\textsc{1T}}
\newcommand{\oneT}{\textsc{1T}}
\newcommand{\twof}{\textsc{2F}}
\newcommand{\pos}{\textsc{POS}}

\newcommand{\iv}{\textsc{VER}}
\newcommand{\aime}[1]{AIME#1$^*$}
\newcommand{\hmmt}[1]{HMMT#1$^*$}
\newcommand{\gpqa}{GPQA-Diamond$^*$}
\newcommand{\mmlu}{MMLU-Redux$^*$}
\newcommand{\crux}{CRUXEval-I$^*$}
\newcommand{\game}{Game of 24$^*$}
\newcommand{\revise}{\textsc{Revise}}
\newcommand{\filter}{\textsc{Filter}}

\definecolor{tablegreen}{RGB}{0,128,0}
\definecolor{tablered}{RGB}{178,34,34}

\definecolor{perf_increase}{HTML}{519ABA}  % Red for improvements
\definecolor{perf_decrease}{HTML}{FF5A78}  % Blue for declines

\newcommand{\perfincrease}[1]{\textcolor{perf_increase}{#1}}
\newcommand{\perfdecrease}[1]{\textcolor{perf_decrease}{#1}}

\usepackage{adjustbox}

\usepackage{makecell}           
\usepackage{colortbl}
\usepackage[dvipsnames]{xcolor}
\usepackage{listings}
\usepackage{datetime}
\newdateformat{monthyeardate}{\monthname[\THEMONTH] \THEYEAR}
\usepackage{soul}

\definecolor{cb_red}{RGB}{213,94,0}
\definecolor{cb_blue}{RGB}{0,114,178}
\definecolor{cb_yellow}{RGB}{240,228,66}
\definecolor{cb_gray}{RGB}{204,204,204}
\definecolor{cb_orange}{RGB}{230,159,0}
\definecolor{cb_skyblue}{RGB}{86,180,233}
\definecolor{cb_green}{RGB}{0,158,115}
\definecolor{cb_purple}{RGB}{204,121,167}

\newcommand{\hlcella}{\cellcolor{cb_gray!30}}
\newcommand{\hlcellb}{\cellcolor{cb_gray!15}}

\usepackage[textsize=tiny]{todonotes}

\icmltitlerunning{Contextual Drag: How Errors in the Context Affect LLM Reasoning}

\begin{document}

\twocolumn[
  \icmltitle{Contextual Drag: How Errors in the Context Affect LLM Reasoning}



  \icmlsetsymbol{equal}{*}

  \begin{icmlauthorlist}
    \icmlauthor{Yun Cheng}{pli}
    \icmlauthor{Xingyu Zhu}{pli}
    \icmlauthor{Haoyu Zhao}{pli}
    \icmlauthor{Sanjeev Arora}{pli}
  \end{icmlauthorlist}

  \icmlaffiliation{pli}{Princeton Language and Intelligence, Princeton University}

  \icmlcorrespondingauthor{}{\{yuncheng, xingyu.zhu\}@princeton.edu}

  \icmlkeywords{Machine Learning, ICML}

  \vskip 0.3in
]



\printAffiliationsAndNotice{}  % no special notice (required even if empty)

\begin{abstract}

Central to many self-improvement pipelines for large language models (LLMs) is the assumption that models can improve by reflecting on past mistakes. We study a phenomenon termed {\em contextual drag}: the presence of failed attempts in the context biases subsequent generations toward structurally similar errors. Across evaluations of $11$ proprietary and open-weight models on $8$ reasoning tasks, contextual drag induces $10$–$20$\% performance drops, and iterative self-refinement in models with severe contextual drag can collapse into self-deterioration. Structural analysis using tree edit distance reveals that subsequent reasoning trajectories inherit structurally similar error patterns from the context. We demonstrate that neither external feedback nor successful self-verification suffices to eliminate this effect. While mitigation strategies such as fallback-behavior fine-tuning and context denoising yield partial improvements, they fail to fully restore baseline performance, positioning contextual drag as a persistent failure mode in current reasoning architectures\footnote{Code is available at \href{https://github.com/princeton-pli/contextual-drag}{github.com/princeton-pli/contextual-drag}.}.

\end{abstract}

\section{Introduction}

\begin{figure*}[t]
    \centering
    \includegraphics[width=0.85\linewidth]{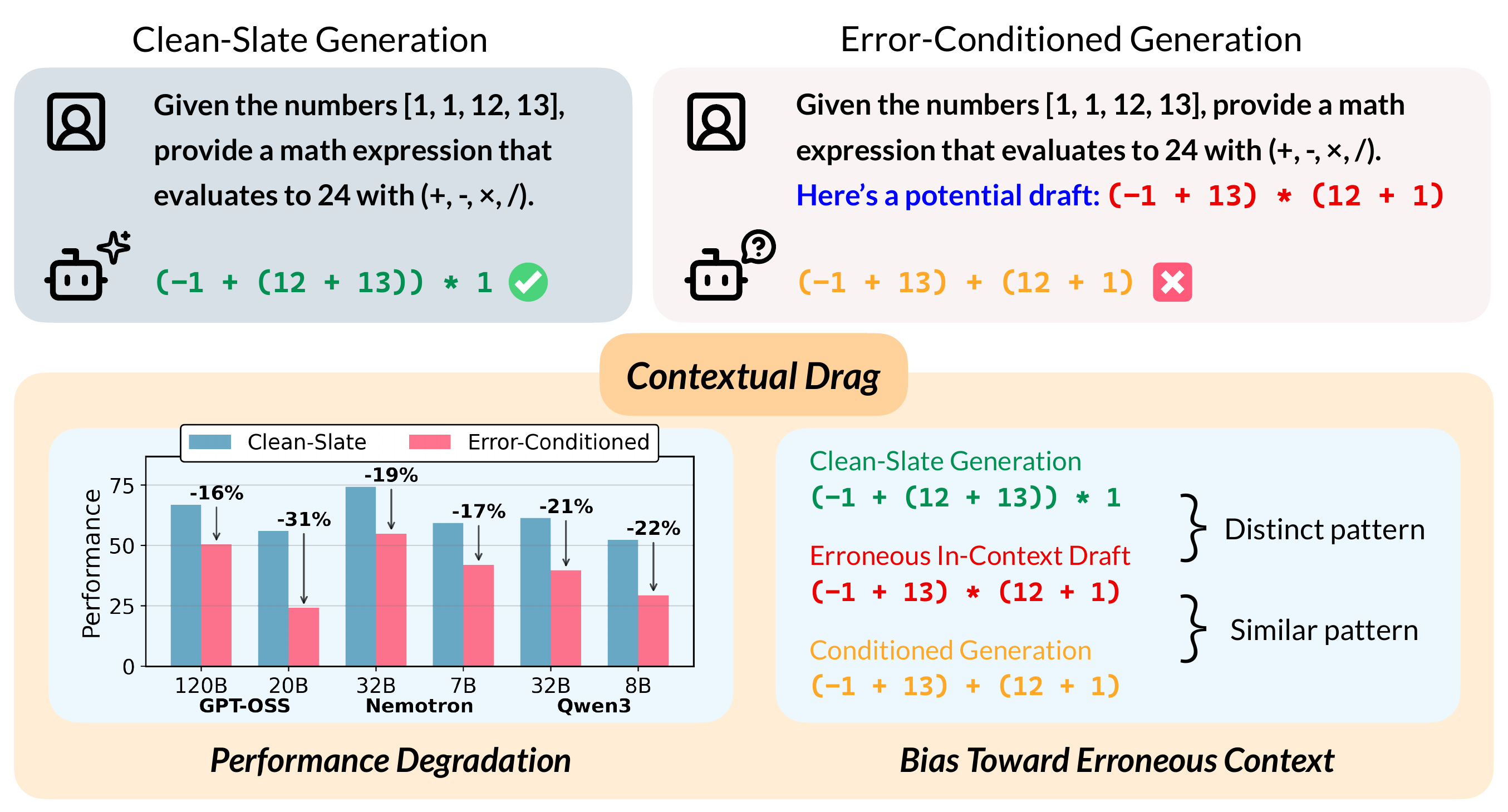}
    \caption{\textbf{Contextual drag} is characterized by the performance drop from clean-slate generation, where the model generates with no additional context, to error-conditioned generation, where the model generates conditioned on incorrect draft solutions in the context. Contextual drag not only manifests as performance degradation but also as a bias in reasoning patterns toward erroneous context.}
    \label{fig:overview}
\end{figure*}

Modern Large Language Model (LLM) workflows for training and inference increasingly rely on critique--verify--revise loops, where a model evaluates and refines another model’s output or its own \cite{madaan2023self, venkatraman2025recursive, madaan2025rethinking}. These pipelines mirror a familiar human strategy: inspecting past mistakes to improve the next attempt.

A potential obstacle is sensitivity to the history of failed attempts building up in the context. Cognitive science shows\\that humans suffer from various forms of {\em anchoring bias}, whereby presented information exerts a lingering influence on subsequent decision-making even after it has been explicitly discredited \cite{tversky1974judgment}. LLMs similarly exhibit strong context sensitivity: misleading hints, user errors, or framing can override parametric knowledge and induce sycophantic behavior \cite{turpin2023language, sharma2024towards, Chen2025, zhu2025cancer}.

 Mitigating anchoring bias in humans usually involves making them mobilize ``System-2 thinking'':  slower, effortful reasoning that enables deliberate checking and correction \cite{kahneman2011thinking}. In parallel, LLM reasoning has shifted from classic chain-of-thought (CoT) prompting \cite{wei2022chain} toward extended thinking with explicit verification structures that more closely resemble System~2-style deliberation \cite{openai2024learning, Guo2025, gemini3-pro}.

Prior work suggests that regular (i.e., short)  CoT reasoning can mitigate anchoring-like effects, for example, by rethinking misleading hints \cite{huang2025empirical} or denoising noisy rationales \cite{zhou2024can} (see more related works in \cref{sec:related-works}). Yet it remains unclear whether today's  reasoning models are robust to biases induced by erroneous solutions that appear in context, as commonly encountered in refinement settings.

With the ability to pause, verify, and backtrack from errors, large reasoning models may be able to identify errors in context and avoid repeating them. However, a contrasting possibility remains: verification  of errors in in-context material might function as a ``local judgment'' that fails to fully undo the downstream influence of contextual errors. That is, a draft that is  accurately rejected  could still shape the model's later reasoning trajectory, \emph{dragging} it toward similar errors. Therefore, the following research questions are of interest: 

\newtcolorbox{rqbox}{
  enhanced,
  colback=white,
  colframe=black!40,
  boxrule=0.6pt,
  arc=2pt,
  left=6pt,right=6pt,top=4pt,bottom=4pt,
  borderline={0.75pt}{0pt}{promptbox!60!black},
  borderline={0.3pt}{2pt}{promptbox!75!black},
  arc=3pt
}

\begin{rqbox}

(RQ1) Are reasoning models prompted with explicit verification robust to the influence of incorrect draft solutions in the context? If not, how does the draft shape the model's reasoning?

\vspace{0.06in}

(RQ2) Do \emph{error signals} (explicit indications that the in-context draft is wrong) from external feedback or self-verification prevent the model from repeating similar errors? If not, what methods can reduce the influence?

\end{rqbox}

\textbf{Main Findings}

We identify \emph{contextual drag} in large reasoning models, whereby conditioning on failed reasoning attempts in the context can bias subsequent generations toward errors.

We evaluate $11$ proprietary and open-weight reasoning models across $8$ reasoning tasks, covering math, science, code, and puzzle solving (\cref{sec:main-eval}). Contextual drag persists even when models execute an explicit verification process, producing consistent $10$-$20\%$ performance drops. The effect is strongest for smaller open-weight models, but frontier models also degrade (to a smaller extent). In a refinement pipeline in which models iteratively generate conditioned on their previous attempt in context, we observe that GPT-OSS-20B, which severely suffers from contextual drag, actually collapses into \emph{self-deterioration} (\cref{sec:self-deterioration}).

To study how incorrect drafts in context shape models' subsequent responses, we conduct a case study on the Game of 24, where solutions can be uniquely parsed into trees and pairwise compared via tree edit distance (\cref{sec:ted-analysis}). Under contextual drag, models' new solutions remain noticeably closer to the erroneous in-context reasoning, providing quantitative evidence of the ``dragging'' behavior. 

To see if contextual drag is just a verification failure, we test whether an explicit \emph{error signal} is sufficient to counter it. In particular, we consider \emph{external} signals injected in the prompt and \emph{self-detected} signals produced by the model’s own verification. In our evaluations, external signals are shown to be insufficient: even when the draft is explicitly labeled as incorrect, models generally remain biased toward the contextual reasoning patterns and repeat structurally similar errors (\cref{sec:conditioned-prompted}). Self-detected signals, however, give model-dependent outcomes: some families (e.g., OpenReasoning-Nemotron) partially recover and can sometimes benefit from the erroneous context, while others remain strongly degraded (\cref{sec:conditioned-selfdetected}).

Finally, we investigate two mitigations: (i) test-time context denoising via multi-turn prompting to rewrite or filter incorrect drafts (\cref{sec:context_denoising}) and (ii) targeted supervised fine-tuning that trains a fallback behavior: resetting to a clean-slate reasoning trajectory once a draft is judged incorrect (\cref{sec:training}). Both methods improve over the baseline, but neither fully recovers the clean-slate performance, positioning contextual drag as a persistent challenge for reliable self-improvement.

\section{Empirical Evaluation of Contextual Drag in Reasoning Models}
\label{sec:evaluation}

In this section, we present a comprehensive evaluation of contextual drag; its impact on current reasoning models (\cref{sec:main-eval}) and iterative pipelines (\cref{sec:self-deterioration}), and a case study on the Game of 24 puzzle that shows how contextual drag shapes subsequent reasoning behavior (\cref{sec:ted-analysis}).

\begin{table*}[t]
    \centering
    \caption{\textbf{Contextual drag in current models}: Existing models exhibit consistent, significant performance drops with incorrect drafts in context. The \underline{anchor models} are underlined for each task. We denote the task by $^*$ to indicate the evaluation is on a \emph{reasonably hard} subset of the original data, and present the size of the subset under the names of the datasets. Note that since the anchor models are chosen based on their performance on the original dataset, they are not necessarily the top-3 performing in the {\direct} columns below. {\direct} refers to clean-slate evaluation with no additional context, and {\onef}/{\twof} refers to models conditioned on $1$ or $2$ incorrect drafts from the anchor models in context. To ensure the statistical significance of observations, we sample 16 generations per question. The full results including $95$\% confidence intervals are in Appendix \cref{tab:main-table-full-pass1}.}
    \label{tab:main-table}

\begin{adjustbox}{width=\textwidth}
    \begin{tabular}{l | c c c c c c c c c c c c}

    \toprule

    \toprule

     & \multicolumn{3}{c}{\textsc{\aime{24}}}  & \multicolumn{3}{c}{\textsc{\aime{25}}} & \multicolumn{3}{c}{\textsc{\hmmt{24}}} & \multicolumn{3}{c}{\textsc{\hmmt{25}}} \\
     & \multicolumn{3}{c}{(10/30)} & \multicolumn{3}{c}{(13/30)} & \multicolumn{3}{c}{(16/30)} & \multicolumn{3}{c}{(21/30)} \\
     \cmidrule(lr){2-4} \cmidrule(lr){5-7} \cmidrule(lr){8-10} \cmidrule(lr){11-13}
     & \textsc{Direct} & {\hlcellb \textsc{1F}} & {\hlcella \textsc{2F}} & \textsc{Direct} & {\hlcellb \textsc{1F}} & {\hlcella \textsc{2F}} & \textsc{Direct} & {\hlcellb \textsc{1F}} & {\hlcella \textsc{2F}} & \textsc{Direct} & {\hlcellb \textsc{1F}} & {\hlcella \textsc{2F}} \\

        \midrule

        GPT-5 & $88.75$ & \hlcellb $88.13$ & \hlcella $86.25$ & $87.50$ & \hlcellb $81.73$ & \hlcella $82.69$ & $76.56$ & \hlcellb $77.73$ & \hlcella $76.56$ & $91.07$ & \hlcellb $84.82$ & \hlcella $85.71$ \\

        Gemini-3 Pro & $98.75$ & \hlcellb $98.75$ & \hlcella $95.00$ & $94.23$ & \hlcellb $90.38$ & \hlcella $92.31$ & $90.62$ & \hlcellb $87.50$ & \hlcella $91.41$ & $99.40$ & \hlcellb $99.40$ & \hlcella $98.21$ \\

        Gemini-2.5 Pro & $100.00$ & \hlcellb $95.00$ & \hlcella $92.50$ & $94.23$ & \hlcellb $90.38$ & \hlcella $92.31$ & $90.62$ & \hlcellb $85.94$ & \hlcella $92.19$ & $99.40$ & \hlcellb $100.00$ & \hlcella $98.21$ \\

        \midrule
        
        GPT-OSS-120B & $66.25$ & \hlcellb $43.75$ & \hlcella $43.75$ & \underline{$66.83$} & \hlcellb $55.29$ & \hlcella $40.38$ & \underline{$48.83$} & \hlcellb $39.84$ & \hlcella $45.31$ & \underline{$61.01$} & \hlcellb $48.51$ & \hlcella $49.11$ \\

        GPT-OSS-20B & $51.88$ & \hlcellb $17.50$ & \hlcella $21.25$ & $51.92$ & \hlcellb $18.75$ & \hlcella $20.67$ & $40.23$ & \hlcellb $13.28$ & \hlcella $17.97$ & $58.63$ & \hlcellb $20.54$ & \hlcella $24.70$ \\

        Nemotron-32B & \underline{$83.75$} & \hlcellb $67.50$ & \hlcella $63.13$ & \underline{$78.37$} & \hlcellb $61.54$ & \hlcella $50.96$ & \underline{$61.72$} & \hlcellb $51.95$ & \hlcella $50.78$ & \underline{$80.36$} & \hlcellb $65.48$ & \hlcella $64.58$ \\

        Nemotron-7B & \underline{$67.50$} & \hlcellb $53.12$ & \hlcella $58.75$ & \underline{$67.79$} & \hlcellb $52.40$ & \hlcella $37.02$ & \underline{$49.22$} & \hlcellb $35.94$ & \hlcella $40.62$ & \underline{$63.10$} & \hlcellb $54.17$ & \hlcella $52.38$ \\

        Qwen3-32B & \underline{$65.00$} & \hlcellb $41.25$ & \hlcella $30.00$ & $54.81$ & \hlcellb $31.25$ & \hlcella $31.25$ & $39.06$ & \hlcellb $35.55$ & \hlcella $40.23$ & $49.70$ & \hlcellb $36.90$ & \hlcella $41.37$ \\

        Qwen3-8B & $55.62$ & \hlcellb $30.63$ & \hlcella $20.62$ & $45.67$ & \hlcellb $20.19$ & \hlcella $11.06$ & $21.88$ & \hlcellb $22.27$ & \hlcella $26.95$ & $42.26$ & \hlcellb $34.23$ & \hlcella $32.74$ \\

        LlamaR1-8B & $21.25$ & \hlcellb $8.13$ & \hlcella $4.38$ & $7.21$ & \hlcellb $4.33$ & \hlcella $2.88$ & $5.08$ & \hlcellb $7.42$ & \hlcella $9.77$ & $10.42$ & \hlcellb $7.44$ & \hlcella $10.42$ \\

        QwenR1-7B & $33.13$ & \hlcellb $11.88$ & \hlcella $8.12$ & $9.62$ & \hlcellb $3.85$ & \hlcella $0.96$ & $8.59$ & \hlcellb $6.25$ & \hlcella $6.25$ & $19.64$ & \hlcellb $10.12$ & \hlcella $13.69$ \\

\midrule

     & \multicolumn{3}{c}{\textsc{{\gpqa}}} & \multicolumn{3}{c}{\textsc{{\mmlu}}} & \multicolumn{3}{c}{\textsc{{\crux}}} & \multicolumn{3}{c}{\textsc{{\game}}} \\
     & \multicolumn{3}{c}{(132/198)} & \multicolumn{3}{c}{(569/2629)} & \multicolumn{3}{c}{(211/800)} & \multicolumn{3}{c}{(653/1362)} \\
     \cmidrule(lr){2-4} \cmidrule(lr){5-7} \cmidrule(lr){8-10} \cmidrule(lr){11-13}
     & \textsc{Direct} & {\hlcellb \textsc{1F}} & {\hlcella \textsc{2F}} & \textsc{Direct} & {\hlcellb \textsc{1F}} & {\hlcella \textsc{2F}} & \textsc{Direct} & {\hlcellb \textsc{1F}} & {\hlcella \textsc{2F}} & \textsc{Direct} & {\hlcellb \textsc{1F}} & {\hlcella \textsc{2F}} \\

        \midrule

        GPT-OSS-120B & \underline{$58.24$} & \hlcellb $35.70$ & \hlcella $31.91$ & \underline{$67.59$} & \hlcellb $42.26$ & \hlcella $31.61$ & $89.23$ & \hlcellb $81.02$ & \hlcella $77.89$ & $77.34$ & \hlcellb $56.94$ & \hlcella $57.09$ \\

        GPT-OSS-20B & $46.97$ & \hlcellb $16.86$ & \hlcella $13.16$ & $58.86$ & \hlcellb $22.73$ & \hlcella $12.83$ & $61.34$ & \hlcellb $42.39$ & \hlcella $40.53$ & $78.30$ & \hlcellb $41.61$ & \hlcella $39.31$ \\

        Nemotron-32B & \underline{$66.81$} & \hlcellb $49.43$ & \hlcella $37.22$ & \underline{$67.25$} & \hlcellb $32.74$ & \hlcella $20.09$ & \underline{$77.06$} & \hlcellb $53.69$ & \hlcella $54.19$ & \underline{$79.54$} & \hlcellb $56.68$ & \hlcella $51.53$ \\

        Nemotron-7B & $51.61$ & \hlcellb $35.27$ & \hlcella $29.40$ & $56.46$ & \hlcellb $23.58$ & \hlcella $14.02$ & $43.55$ & \hlcellb $35.27$ & \hlcella $31.32$ & $75.35$ & \hlcellb $45.97$ & \hlcella $34.77$ \\

        Qwen3-32B & \underline{$57.62$} & \hlcellb $34.19$ & \hlcella $23.30$ & \underline{$70.62$} & \hlcellb $40.53$ & \hlcella $21.97$ & \underline{$75.83$} & \hlcellb $54.89$ & \hlcella $50.90$ & \underline{$78.48$} & \hlcellb $43.40$ & \hlcella $25.47$ \\

        Qwen3-8B & $47.77$ & \hlcellb $27.13$ & \hlcella $20.45$ & $61.65$ & \hlcellb $27.76$ & \hlcella $17.21$ & \underline{$67.49$} & \hlcellb $40.09$ & \hlcella $36.20$ & \underline{$76.29$} & \hlcellb $32.92$ & \hlcella $23.26$ \\

        LlamaR1-8B & $31.53$ & \hlcellb $20.69$ & \hlcella $17.85$ & $41.17$ & \hlcellb $13.72$ & \hlcella $8.89$ & $30.32$ & \hlcellb $23.84$ & \hlcella $21.81$ & $35.34$ & \hlcellb $5.44$ & \hlcella $3.37$ \\

        QwenR1-7B & $21.26$ & \hlcellb $14.73$ & \hlcella $15.20$ & $36.60$ & \hlcellb $13.62$ & \hlcella $10.25$ & $36.84$ & \hlcellb $22.74$ & \hlcella $16.82$ & $40.69$ & \hlcellb $15.60$ & \hlcella $8.49$ \\

        % \midrule

        % Llama3.1-70B & $31.39$ & \hlcellb $23.86$ & \hlcella $16.00$ & $62.54$ & \hlcellb $20.84$ & \hlcella $8.58$ & $34.74$ & \hlcellb $22.77$ & \hlcella $18.09$ & $14.83$ & \hlcellb $5.97$ & \hlcella $2.92$ \\

        % Llama3.1-8B & $17.23$ & \hlcellb $14.87$ & \hlcella $7.77$ & $38.01$ & \hlcellb $26.91$ & \hlcella $12.14$ & $16.89$ & \hlcellb $18.85$ & \hlcella $14.16$ & $6.48$ & \hlcellb $4.31$ & \hlcella $2.95$ \\

        % Qwen2.5-32B & $36.32$ & \hlcellb $17.57$ & \hlcella $10.42$ & $57.86$ & \hlcellb $21.23$ & \hlcella $7.26$ & $40.86$ & \hlcellb $22.94$ & \hlcella $13.13$ & $15.63$ & \hlcellb $4.54$ & \hlcella $3.19$ \\

        % Qwen2.5-7B & $22.54$ & \hlcellb $11.41$ & \hlcella $8.29$ & $44.01$ & \hlcellb $16.56$ & \hlcella $6.93$ & $28.19$ & \hlcellb $21.51$ & \hlcella $15.36$ & $5.92$ & \hlcellb $2.20$ & \hlcella $1.81$ \\

        \bottomrule

\end{tabular}

    \end{adjustbox}

\end{table*}

\subsection{Evaluation}
\label{sec:main-eval}

\paragraph{Models and benchmarks}
We evaluate proprietary models GPT-5 \cite{gpt5} and Gemini 2.5/3 Pro \cite{gemini2.5-pro, gemini3-pro}\footnote{We only evaluated proprietary models on competition math benchmarks due to prohibitive API costs in academic settings.} and open-weight reasoning models GPT-OSS-20B/120B \cite{gpt-oss}, OpenReasoning-Nemotron-7B/32B \cite{ahmad2025opencodereasoning, ahmad2025opencodereasoningii, moshkov2025aimo}, Qwen3-8B/32B \cite{yang2025qwen3}, and Deepseek R1-distilled Llama-8B/Qwen-7B \cite{deepseekai2025deepseekr1incentivizingreasoningcapability}.

We evaluate on problems from four competition math datasets (AIME24/25 and HMMT24/25 \cite{balunovic_srimatharena_2025}), two general QA benchmarks (GPQA \cite{rein2024gpqa} and MMLU-Redux 2.0 \cite{gema2024mmlu}), one code reasoning benchmark (CRUXEval-I \cite{gu2024cruxeval}), and one puzzle dataset (Game of 24).

\paragraph{Selecting in-context drafts and evaluation problems}
For each task, we first run clean-slate generation for the set of open-weight models, and select the top-3 performers as \emph{anchor models} (underlined in \cref{tab:main-table}). Their responses are used as realistic in-context drafts and avoid confounding drops from low-quality rationales \cite{li2025rethinking}.

We then restrict evaluation to a \emph{reasonably hard} subset of problems in which anchor models exhibit both successes and failures: each selected question must have at least two correct and two incorrect responses\footnote{We verify correctness by parsing the final answer and compare the model's answer with ground truth via symbolic verifier \cite{kydlicek2024mathverify}. The filter ensures the task is nontrivial yet not impossible to solve for the models being evaluated.}. These subsets are also of the most interest for test-time scaling as they correspond to the gap between pass@$1$ and pass@$k$ (for larger $k$), making them the key area of improvement for refinement pipelines. We provide the size of the filtered subsets in \cref{tab:main-table} under the dataset names.

Using this setup, we measure: (1) {\direct} performance, where models solve the problem with no additional context, and (2) {\onef}/{\twof} performance, in which models are additionally conditioned on one or two incorrect draft solutions (\cref{fig:eval-setup}). In the {\onef}/{\twof} settings, the model is explicitly instructed to verify each draft before producing a new solution. The performance change from {\direct} to {\onef}/{\twof} quantifies contextual drag.

\paragraph{Results} 
\cref{tab:main-table} shows that popular open-weight and even proprietary reasoning models consistently exhibit contextual drag. Across all evaluations, introducing one or two incorrect drafts generally leads to significant performance degradation. On math reasoning tasks (AIME24/25, HMMT24/25), the drop is around $10$-$20$\% for most open-weight reasoning models, with a few exceptions (Qwen3-8B and LlamaR1-8B) on HMMT24, where {\onef} and {\twof} performance slightly surpassed the {\direct} performance.

The drop is more severe for smaller models: GPT-OSS-20B and QwenR1-7B lose almost half of their {\direct} accuracy across multiple datasets. Proprietary models and stronger open-weight models such as GPT-OSS-120B and Nemotron-32B show more robustness but still experience non-negligible degradation. The trends are consistent for GPQA, MMLU, CRUXEval-I, and Game of 24. 

\paragraph{Ablations}
Multiple factors can affect a model's ability to process context information. Therefore, we ablate the following factors affecting contextual drag: (1) \textbf{model size}; (2) varying position of the question and the draft solution (\textbf{Position}); (3) explicit verification (\textbf{Verification}); (4) more drafts with mixed correctness (\textbf{Scaling}). We include the details in Appendix \ref{app:ablations}.

\cref{tab:main-table} shows that model scale does not substantially alleviate contextual drag. \cref{tab:ablations} shows that reordering the question and draft, as well as explicit verification partially help, but both fail to restore the {\direct} performance, suggesting that contextual drag reflects a more fundamental issue in processing contextual information. Meanwhile, performance degrades sharply when the context contains only incorrect drafts, with more severe drops as the number of incorrect drafts grows. Correct drafts in context generally improve performance, but models can still struggle when incorrect drafts dominate the context (\cref{fig:scaling-ablation}).

\subsection{Self-Deterioration in Iterative Refinement}
\label{sec:self-deterioration}

\begin{figure}[t]
    \centering
    \includegraphics[width=\linewidth]{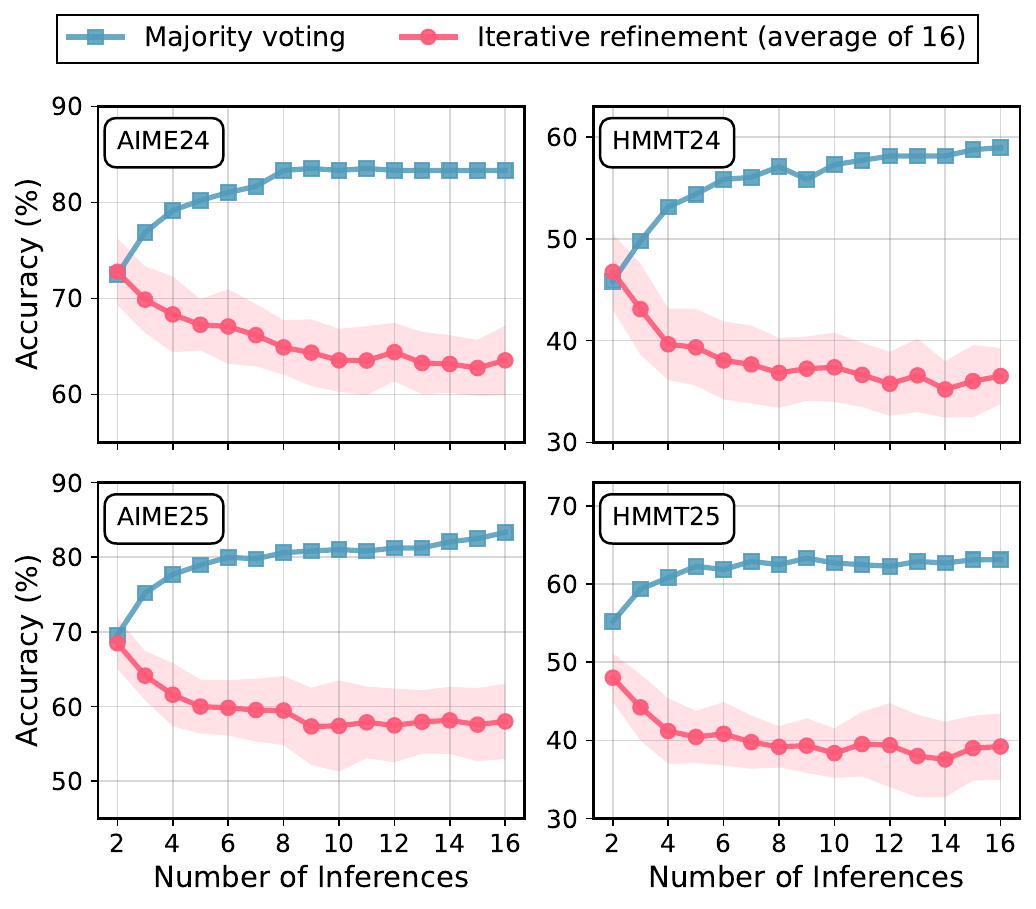}  
    \caption{\textbf{Self-deterioration}: Iterative refinement of models with severe degradation from contextual drag, such as GPT-OSS-20B, collapses in accuracy across iterations, whereas majority voting improves steadily. We sample 16 iterative refinement trajectories and report the average performance across all trajectories.}
    \label{fig:self-deterioration}
\end{figure}

Contextual drag challenges the underlying assumption of many iterative refinement pipelines that models can improve more efficiently by conditioning on their past attempts. \cref{fig:self-deterioration} shows that GPT-OSS-20B, one of the models with the most severe degradation from contextual drag (\cref{tab:main-table}), exhibits \emph{self-deterioration} during iterative refinement\footnote{The model first generates an initial response with no additional context. After that, the output from the previous inference is used as the in-context draft. Note that the draft can be either correct or incorrect in this setting.}, a phenomenon that has also been observed for non-reasoning models by \citet{huang2024large} and \citet{xu2024pride}.

Compared with its majority voting performance, where sampling multiple independent solutions steadily improves accuracy, iterative refinement results in a gradual decline. Therefore, contextual drag exposes such iterative pipelines to a critical vulnerability: without mechanisms to detect, filter, or reset from low-quality drafts in the context, the model can become increasingly biased toward the erroneous context.

\begin{figure}[t]
    \centering
    \includegraphics[width=\linewidth]{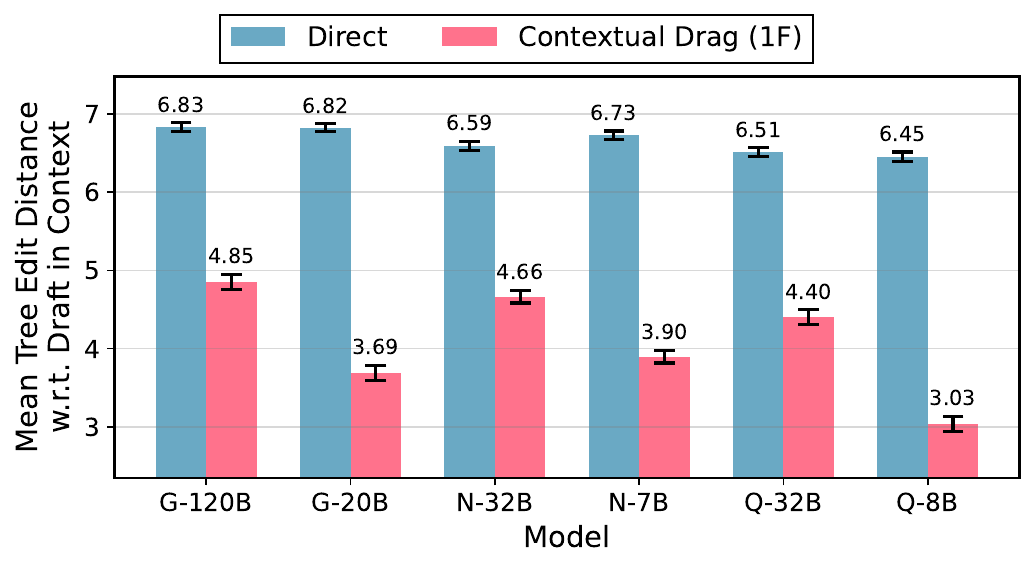}
    \caption{\textbf{Contextual drag arises from copying reasoning patterns}: Measured by mean tree edit distance (TED), models' subsequent solutions under contextual drag ({\onef}) stay significantly closer to the in-context erroneous reasoning than to clean-slate solutions ({\direct}). Lower TED indicates stronger structural similarity to the in-context draft but not necessarily worse performance.}
    \label{fig:edit-distance-base}
\end{figure}

\subsection{Quantifying Contextual Drag:%Shapes Subsequent Reasoning Behavior on
~Game of 24}
\label{sec:ted-analysis}

Given the in-context draft, the model clearly appears to be trying to craft a new solution. What explains the contextual drag? One plausible explanation is that  the underlying architecture (e.g., the attention mechanism) predisposes the model to reuse some reasoning patterns from the draft while falling well short of full copying. To study this possibility, we use a controlled setting, {\em Game of 24} (also often seen as a restricted variant of the \textit{Countdown} problem), where the similarity between answers is quantifiable.

The input to the task is a multiset of four integers drawn from $\{1,\dots,13\}$. The objective is to construct an arithmetic expression using each integer exactly once, with the binary operations $\{+, -, \times, \div\}$ and parentheses, that has a resulting value of exactly $24$. Since each solution can be represented as a tree where the leaf nodes are the provided integers and the non-leaf nodes are operations, we can measure the structural similarity between responses using tree edit distance (TED) \cite{zhang1989simple}. This allows us to quantify the extent to which the models' subsequent answers resemble the incorrect drafts. For each problem, we pair each draft with two sets of responses: the clean-slate responses under {\direct} and the conditioned responses under contextual drag ({\onef}/{\twof}) (\cref{fig:ted}). TED is then computed between the draft and each set of responses. 

As shown in \cref{fig:edit-distance-base}, the conditioned responses under {\onef} remain significantly closer to the incorrect draft than the clean-slate responses under {\direct} across models. This demonstrates that contextual drag operates at the level of internal reasoning structure: models do not merely imitate surface tokens, but subtly follow the erroneous computational pathway suggested by the draft. Additional results under {\twof} exhibit similar trends, suggesting that diversity of responses in context does not substantially alleviate the bias in structure (Appendix \cref{fig:ted-2f}).

Here we highlight that TED is not a performance metric; it serves only as a qualitative indicator of structural similarity between conditioned generations and in-context drafts under contextual drag. The results complement earlier sections by showing that contextual drag is not only a performance phenomenon but a systematic structural distortion of reasoning.

Overall, contextual drag represents a systematic failure mode of current LLMs with implications for multi-step reasoning workflows. Beyond iterative refinement, incorrect hypotheses and failed intermediate branches in extended CoTs can similarly bias subsequent exploration \cite{feng2025characterizes}, resembling a form of contextual drag in extended reasoning. We leave further investigation to future work.

\section{Contextual Drag Under Error Signals}
\label{sec:evaluation-conditioned-drop}

In \cref{sec:evaluation}, we showed that contextual drag persists across a wide range of models under different settings. Next, we take a deeper dive into the failure mode of contextual drag that leads to performance degradation.

A natural hypothesis is that contextual drag is mainly a verification failure: once the model is given a reliable signal that the in-context draft is wrong, it should at least avoid similar errors when producing a new answer. In this section, we show that even with explicit, accurate error signals, either provided externally in the prompt or produced by the model itself, conditioning on an incorrect draft can still bias subsequent reasoning.

We study the effect of error signals on contextual drag in the simplest setting with a single incorrect draft in context (\onef). In \cref{sec:conditioned-prompted}, we still observe severe contextual drag despite clear signals of error in the prompt. In \cref{sec:conditioned-selfdetected}, our post-hoc analysis on reasoning traces with correct self-verification reveals that accurate verification can help some models recover from the bias and even benefit from the erroneous context. However, the improvement varies across models, indicating that verification ability is not the only barrier to eliminating contextual drag.

\subsection{External Error Signal}
\label{sec:conditioned-prompted}

\begin{figure}[t]
    \centering
    \includegraphics[width=\linewidth]{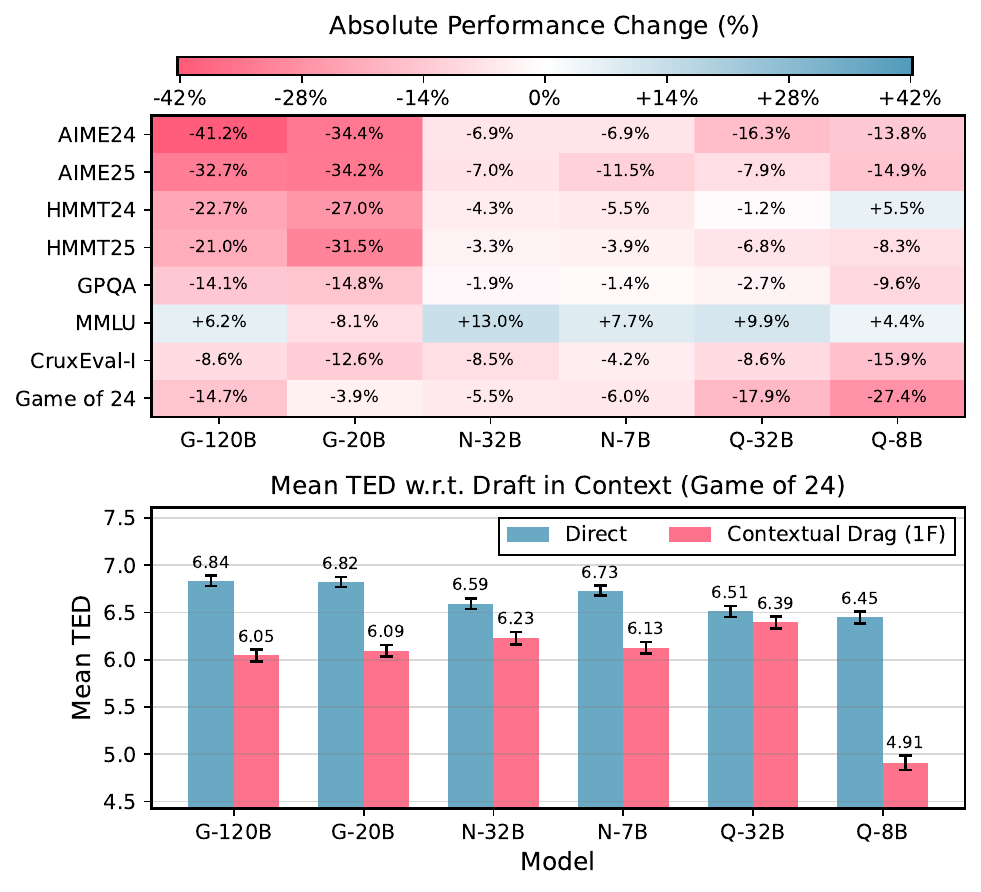}
    \caption{\textbf{External error signal does not recover subsequent reasoning from contextual drag}: Models still experience significant drops from {\direct} to {\onef} across benchmarks with the exception of MMLU, which is usually considered more knowledge-intensive than reasoning-intensive. Full results are provided in \cref{tab:appd-external-error}.}
    \label{fig:conditioned-prompted-error}
\end{figure}

To test whether contextual drag is mainly caused by the lack of error signals, we reuse the {\onef} evaluation setup in \cref{sec:main-eval} with a \hyperref[template-ablation-frm]{modified prompt} that explicitly highlights the incorrectness of the draft and warns against directly copying the answer (Appendix \cref{app:evals-conditioned-drop}). This provides an external, reliable error signal independent of the model's own verification capability.

As shown in \cref{fig:conditioned-prompted-error}, benchmark performance under {\onef} still drops substantially relative to \textsc{Direct} across tasks and models. Moreover, the generations under {\onef} on Game of 24 remain structurally closer to the incorrect draft than the clean-slate counterparts under {\direct}. Even with unambiguous error signals in the prompt, the draft continues to bias downstream reasoning. External error signals alone are insufficient to mitigate contextual drag.

\subsection{Self-Detected Error Signal (Post-hoc)}
\label{sec:conditioned-selfdetected}

\begin{figure}[t]
    \centering
    \includegraphics[width=\linewidth]{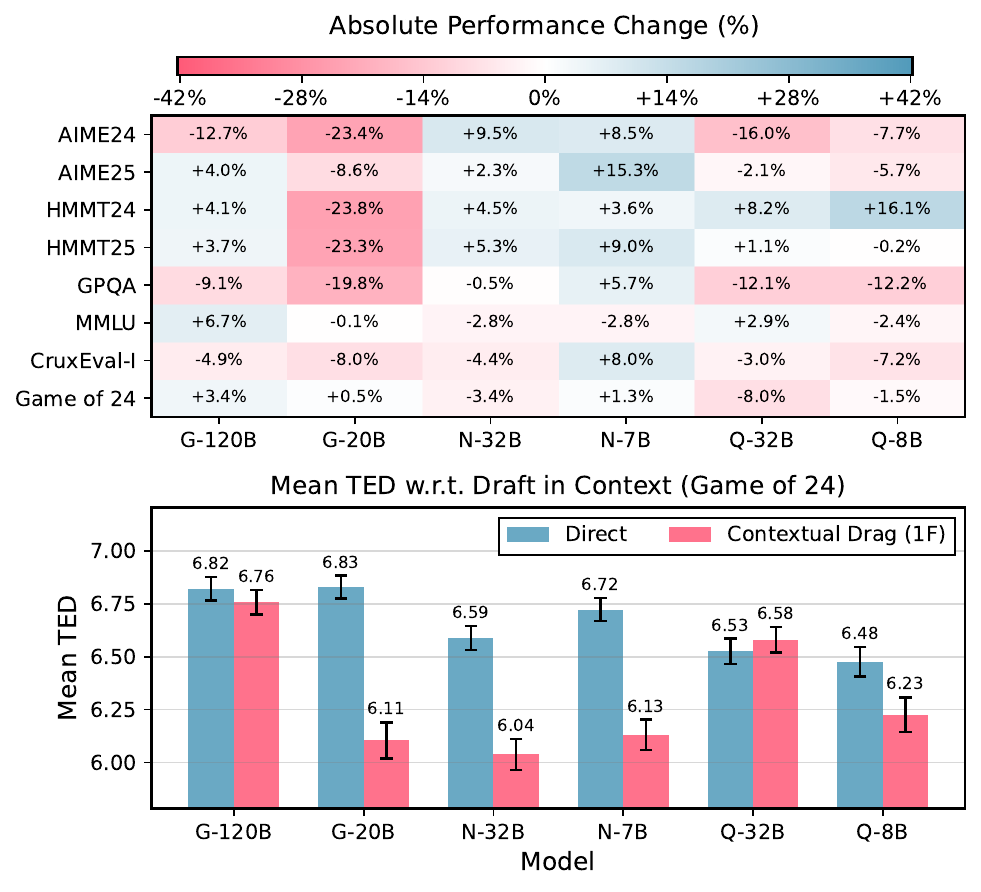}
    \caption{\textbf{Correct self-verification gives rise to varying robustness to contextual drag}: Conditioning on self-detected error signals, contextual drag still persists with varied performance changes: Nemotron-7B/32B recover toward (and even surpass) {\direct} performance, while GPT-OSS-20B remains strongly degraded. Full results are provided in \cref{tab:appd-self-detected-error}.}
    \label{fig:conditioned-self-detected-error}
\end{figure}

We next examine the effect of successful self-verification, when the model itself explicitly recognizes that the draft is wrong, on contextual drag. We filter the {\onef} generations in \cref{sec:main-eval} to only retain trajectories with an explicit correct verdict of the draft's incorrectness\footnote{We verify verdicts via parsable structured tags in the model output. See Appendix \cref{app:verifier_generation} for the exact format.}. We note that this analysis is purely post-hoc and effectively assumes access to an oracle filter: it selectively keeps ``good'' trajectories and therefore biases toward stronger performance. We also note that since the retained subset differs by model and task, the resulting accuracy and TED are not directly comparable to the full {\onef} results in \cref{sec:main-eval}.

As shown in \cref{fig:conditioned-self-detected-error}, when conditioned on correct self-verification, structural contextual drag still generally persists on Game of 24. However, the impact on performance is no longer uniform across models: some models, such as the Nemotron family partially recover and can even exceed {\direct} performance on the retained subset, while others, especially GPT-OSS-20B, remain strongly degraded despite successfully producing an ``Incorrect'' verdict for the draft.

The improved behaviors of the Nemotron family are promising and of interest to future self-improvement pipelines: it is possible for accurate self-verification to alleviate contextual drag and, in favorable cases, even surpass {\direct} performance. On the other hand, the varying performance change also indicates that verification may not be the only bottleneck for full recovery from contextual drag, presenting a great challenge to self-improvement pipelines that rely on external or self-generated feedback loops.

\section{Mitigating Contextual Drag}
\label{sec:mitigation}

As shown in \cref{sec:evaluation}, explicitly verifying the contextual content using extended thinking is not sufficient to prevent contextual drag. In particular, contextual drag can bypass both external error signals and internal error recognition (\cref{sec:evaluation-conditioned-drop}). These findings motivate mitigation beyond error detection and require changes in the model's reasoning state to correct biases induced by erroneous context. 

In this section, we investigate methods that partially mitigate the impact of contextual drag from two perspectives. First, models can be prompted to modify the context content before using it via test-time context denoising (\cref{sec:context_denoising}). Second, models can be trained via targeted supervised fine-tuning, for example, to ignore the in-context draft and switch to clean-slate reasoning upon error detection (\cref{sec:training}).

\subsection{Context Denoising}
\label{sec:context_denoising}

\begin{table*}[t]
    \centering
    \caption{\textbf{Context denoising mitigates but fails to eliminate contextual drag}: \textsc{Rev.}/{\revise} and \textsc{Filt.}/{\filter} are shown. We include the {\iv} ablation under {\onef} ({\onef}-V) for fair comparison (\cref{tab:ablations}), as our implementation omits explicit verification in downstream inference.}
    \label{tab:context-denoising-main-table}
    \begin{adjustbox}{width=\textwidth}
    \begin{tabular}{l | c c c c c c c c c c c c c c c}

    \toprule

    \toprule

     & \multicolumn{5}{c}{\textsc{\hmmt{25}}} & \multicolumn{5}{c}{\textsc{{\gpqa}}} & \multicolumn{5}{c}{\textsc{{\crux}}} \\
     \cmidrule(lr){2-6} \cmidrule(lr){7-11} \cmidrule(lr){12-16}
     & {\direct} & \textsc{Rev.} & \textsc{Filt.} & {\hlcellb {\onef}} & {\hlcellb {\onef}-V} & {\direct} & \textsc{Rev.} & \textsc{Filt.} & {\hlcellb {\onef}} & {\hlcellb {\onef}-V} & {\direct} & \textsc{Rev.} & \textsc{Filt.} & {\hlcellb {\onef}} & {\hlcellb {\onef}-V} \\

        \midrule

        GPT-OSS-120B & $61.01$ & $68.75$ & $59.23$ & \hlcellb $48.51$ & \hlcellb $52.98$ & $58.24$ & $46.88$ & $54.88$ & \hlcellb $35.70$ & \hlcellb $38.16$ & $89.23$ & $81.98$ & $87.80$ & \hlcellb $81.02$ & \hlcellb $87.13$ \\

        GPT-OSS-20B & $58.63$ & $62.20$ & $45.83$ & \hlcellb $20.54$ & \hlcellb $42.26$ & $46.97$ & $33.33$ & $43.70$ & \hlcellb $16.86$ & \hlcellb $26.66$ & $61.34$ & $61.24$ & $49.73$ & \hlcellb $42.39$ & \hlcellb $45.11$ \\

        Nemotron-32B & $80.36$ & $74.40$ & $91.96$ & \hlcellb $65.48$ & \hlcellb $67.56$ & $66.81$ & $48.15$ & $68.47$ & \hlcellb $49.43$ & \hlcellb $39.44$ & $77.06$ & $67.49$ & $73.80$ & \hlcellb $53.69$ & \hlcellb $52.59$ \\

        Nemotron-7B & $63.10$ & $63.10$ & $62.20$ & \hlcellb $54.17$ & \hlcellb $51.19$ & $51.61$ & $35.27$ & $55.73$ & \hlcellb $35.27$ & \hlcellb $27.46$ & $43.55$ & $48.34$ & $39.13$ & \hlcellb $35.27$ & \hlcellb $35.34$ \\

        Qwen3-32B & $49.70$ & $48.81$ & $43.15$ & \hlcellb $36.90$ & \hlcellb $40.77$ & $57.62$ & $27.04$ & $58.38$ & \hlcellb $34.19$ & \hlcellb $22.68$ & $75.83$ & $65.36$ & $62.90$ & \hlcellb $54.89$ & \hlcellb $53.96$ \\

        Qwen3-8B & $42.26$ & $38.69$ & $42.86$ & \hlcellb $34.23$ & \hlcellb $36.61$ & $47.77$ & $25.24$ & $45.27$ & \hlcellb $27.13$ & \hlcellb $18.75$ & $67.49$ & $49.00$ & $51.80$ & \hlcellb $40.09$ & \hlcellb $38.33$ \\

        \bottomrule

\end{tabular}

    \end{adjustbox}

\end{table*}

One mitigation strategy is to prompt the model to denoise the context at test time, thereby reducing the impact of incorrect intermediate reasoning on subsequent inference.

\subsubsection{Methods}
We evaluate two test-time context denoising pipelines: the model is asked to either (1) revise individual drafts to improve quality ({\revise}) or; (2) filter the drafts to only keep the correct and useful intermediate steps given a high-level strategy produced by teh model itself ({\filter}). We include the complete details of the implementation in \cref{app:context-denoising-implementation}.

\subsubsection{Results}

\textbf{Both methods mitigate, but do not eliminate, contextual drag, with {\filter} being more robust.} Across tasks and models, {\revise} and {\filter} outperform the {\onef} baseline (\cref{tab:context-denoising-main-table}). {\filter} more consistently recovers the performance gap relative to {\direct} performance, while gains from {\revise} are mixed. We attribute this to contextual drag persisting during the revision process, which limits {\revise}'s ability to correct deep conceptual errors, whereas {\filter} leverages a clean-slate-generated high-level strategy to guide draft denoising. However, because both methods rely on the model's own, unreliable judgment of in-context drafts, neither fully recovers from the performance drop in all settings (e.g., GPT-OSS-120B/20B on GPQA and Qwen3-32B/8B on CRUXEval-I). We include the full results in \cref{app:context-denoising-full-results}.

\textbf{Context denoising incurs higher inference costs at scale.} Since both pipelines perform multiple inference passes per draft, test-time cost increases by at least $2\times$, scaling with the number of denoising steps and drafts. In practice, batching drafts can partially amortize this cost, but often at the expense of denoising quality.

\subsection{Training Models to Switch to Clean-Slate Reasoning upon Error Detection}
\label{sec:training}

\begin{table}[t]
    \centering
    \caption{\textbf{Targeted SFT generally improves robustness under erroneous drafts}: Accuracy on competitive math benchmarks under {\onef}, and on the post-hoc subset where the model self-verifies the draft as ``Incorrect'' (\cref{sec:conditioned-selfdetected}). ``Finetuned'' denotes GPT-OSS-20B after our conditional fallback training; deltas in parentheses are absolute percentage-point changes vs. baseline.}
\begin{adjustbox}{width=\linewidth}
\begin{tabular}{l|cccc}
\toprule
\toprule
\textsc{Setting} & \textsc{\aime{24}} & \textsc{\aime{25}} & \textsc{\hmmt{24}} & \textsc{\hmmt{25}} \\
\midrule
\multicolumn{5}{l}{\textsc{1F}} \\
\midrule
Baseline & $17.5$ & $18.8$ & $13.3$ & $21.2$ \\
Finetuned & $40.6$ (\perfincrease{+$23.1$}) & $20.7$ (\perfincrease{+$1.92$}) & $19.2$ (\perfincrease{+$5.89$}) & $27.8$ (\perfincrease{+$6.56$}) \\
\midrule
\multicolumn{5}{l}{1F (Conditioned on Self-Detected Error Signal)} \\
\midrule
Baseline & $28.5$ & $43.3$ & $16.4$ & $33.9$ \\
Finetuned & $52.9$ (\perfincrease{+$24.4$}) & $38.7$ (\perfdecrease{-$4.65$}) & $30.5$ (\perfincrease{+$14.0$}) & $45.3$ (\perfincrease{+$11.4$}) \\
\bottomrule
\end{tabular}
\end{adjustbox}
    \label{tab:sft_result}
\end{table}

Motivated by our finding that contextual drag persists even after error recognition (\cref{sec:evaluation-conditioned-drop}), we study a simple supervised mitigation that trains an explicit \emph{fallback behavior}: after step-by-step verification, the model either reuses the draft that is deemed correct, or switches to a clean-slate solution trajectory when it detects an error in the draft's reasoning.

We train GPT-OSS-20B, because it exhibits the largest degradation even under error signals. We sample 40k (problem, correct-draft) pairs and 40k (problem, incorrect-draft) pairs from a moderately challenging subset of Big-Math-RL-Verified \cite{albalak2025bigmathlargescalehighqualitymath}. Drafts are drawn from a pool of open-weight reasoning models, and draft correctness is determined using verified ground-truth answers. 

For each (problem, incorrect-draft) pair, we construct a synthetic reasoning trajectory by concatenating (1) a verification reasoning trajectory generated by GPT-OSS-20B that successfully verifies the draft as ``Incorrect'' and (2) a clean-slate trajectory generated by GPT-OSS-20B for the same problem (see the {\onef} case in \cref{fig:mitigation-datasynth}). This provides an explicit fallback trace that does not reference the draft. 

For each pair of problem and correct draft, we concatenate a verification trajectory that successfully verifies the draft as ``Correct'' with a fixed template guiding the model to reuse the answer from the draft in context (see the {\onet} case in \cref{fig:mitigation-datasynth}). This teaches a conditional reuse of helpful context rather than a uniform behavior of ignoring the in-context draft, which would be uninteresting.
We fine-tune GPT-OSS-20B on these synthetic trajectories with standard supervised learning (token-level next-token prediction). Complete details on data curation, prompt templates, and training are provided in \cref{app:training-details}, including a comparison with GRPO in \cref{sec:apx-grpo}.

\subsubsection{Results}

\textbf{Targeted SFT consistently mitigates contextual drag under erroneous context, but does not fully restore clean-slate performance.} As shown in \cref{tab:sft_result}, targeted SFT yields consistent gains in the {\onef} setting, indicating that even a lightweight synthesis rule can measurably reduce the degradation induced by erroneous context. However, the fine-tuned model does not fully recover {\direct} performance: although the gap narrows, {\onef} accuracy remains below {\direct}, suggesting that residual contextual drag persists even after training the reset mechanism.

\textbf{Targeted SFT improves the model’s ability to act on self-detected error signals.} We repeat the post-hoc conditioned analysis from \cref{sec:conditioned-selfdetected}, restricting the analysis to cases where the model correctly identifies the draft as incorrect before producing its final answer. Under this evaluation, the fine-tuned model generally improves relative to its pre-tuning counterpart across benchmarks (with the exception of AIME25). This provides a complementary view of training effectiveness: beyond improving unconditional {\onef} accuracy, targeted SFT strengthens the model’s capacity to translate error awareness into improved downstream reasoning.

\textbf{Improved robustness comes at the cost of reduced utilization of correct context.} Despite gains under erroneous context, targeted SFT incurs a notable performance drop when the in-context draft is correct (\cref{app:sft_results_1T}). This points to a robustness-utilization tradeoff: training the model to reset upon detecting errors can also weaken its ability to exploit helpful context when the draft is reliable. Taken together, these results position targeted SFT as a proof-of-concept mitigation that enhances robustness to incorrect context, but not yet a drop-in solution without sacrificing performance under correct context.

\section{Related Work}
\label{sec:related-works}

\textbf{Anchoring bias}\quad
LLMs exhibit anchoring behavior, which mirrors the anchoring effect in human judgment and decision-making \cite{tversky1974judgment, FURNHAM201135}. LLM outputs can be systematically biased by numerical hints or framing information embedded in prompts \cite{NGUYEN2024100971, huang2025empirical, o2025anchoring, o2025anchoring2, lou2026anchoring}. These works establish anchoring as a reproducible phenomenon in modern LLMs.

\textbf{Context sensitivity and noisy CoTs}\quad
Prior studies have documented that LLMs are sensitive to positional biases \cite{liu2024lost}, framing \cite{Lorè2024}, and verification \cite{asai2024self} of contextual information. Meanwhile, providing noisy rationales in context can significantly degrade model performance, even when models are instructed to critique the flawed reasoning \cite{zhou2024can, li-etal-2025-patterns, lee2026lost}. These findings suggest that LLMs are subject to strong influence from incorrect information in the context.

\textbf{Self-improvement and multi-agent systems}\quad
Sequential refinement \cite{madaan2023self} and multi-agent aggregation \cite{madaan2025rethinking, venkatraman2025recursive, wang2024mixture, li2025rethinking} pipelines implicitly assume that exposing the model to prior solutions provides informative guidance that facilitates the search for correct answers.

\textbf{Conflicts between context and parametric knowledge}\quad
Recent research examines how LLMs resolve discrepancies between internal parametric knowledge and non-parametric information in the context \cite{zhao2025understanding, goyal2025contextparametric}, exploring interventions through head pruning \cite{jin-etal-2024-cutting}, fine-tuning \cite{wang2024resolving}, and attention interventions \cite{li2025taming}. Context-parametric knowledge conflicts are primarily defined at the level of factual or semantic correctness.

Contextual drag in our study can be viewed as a form of anchoring in reasoning, but we distinguish our work from prior work in several key ways. We are the first to show that the influence of erroneous context persists even after explicit error signals, and manifests as structural distortions in reasoning rather than solely as performance degradation. Models under severe contextual drag exhibit self-deterioration, posing\\a significant challenge to reliable self-improvement and multi-agent systems. Our findings also complement work on belief persistence in knowledge conflicts \cite{he2025martingale}.

\section{Limitations and Future Work}
\label{sec:future-directions}

Contextual drag reflects a failure mode of in-context learning where models over-exploit coherent but incorrect context. This reliance, which emerges during pretraining on structured, high-quality data \cite{xie2022an}, explains our observation that correct drafts yield large performance gains, yet incorrect drafts cause persistent degradation. Notably, the partial recovery results of our targeted training reveal a tradeoff between coherence-driven context utilization and robustness to errors: contextual drag is often reduced at the cost of diminished benefits from correct context. It would be interesting to investigate how this tradeoff arises from training design choices such as data composition.

We also note model-dependent differences in recovery under correct self-verification between the Nemotron models and models like GPT-OSS-20B (\cref{sec:conditioned-selfdetected}). A more fine-grained comparison between these models may reveal mechanisms or training choices, such as reasoning-trace distillation, that improve robustness against contextual drag. Finally, the persistence of contextual drag suggests a fundamental limitation of current attention-based architectures, which lack mechanisms to truly reset the reasoning state regardless of any verification effort. Fully resolving contextual drag may require architectural or training-level changes that enable more selective use of context.

\section{Conclusion}
We identify contextual drag as a  failure mode in current reasoning models, where erroneous solutions in context bias subsequent reasoning toward structurally similar errors. Across models and tasks, this effect persists even when explicit error signals are provided, indicating that contextual drag is not merely a verification failure and presents a significant challenge to iterative refinement pipelines. Structural analysis further shows that incorrect drafts distort the reasoning patterns of subsequent generations under contextual drag. While targeted fine-tuning and context denoising partially mitigate this effect, neither fully restores clean-slate performance. These findings highlight the need for principled mechanisms to reset or discount unreliable context in multi-step reasoning and self-improvement workflows.

\section*{Acknowledgements}

We acknowledge the support from NSF, Schmidt Foundation, DARPA  AIQ Program, OpenAI and Google Inc. CY is additionally supported by the Francis Robbins Upton Fellowship in Engineering. XZ is additionally supported by the Gordon Y.S. Wu Fellowship in Engineering. We thank Abhishek Panigrahi, Yong Lin, Simon Park, Xingyu Fu, Zixuan Wang, Xingyu Dang, Liam Fowl, Nikunj Saunshi, Tomer Wolfson, Evan Russek, Ionatan Kuperwajs, Danqi Chen, Tom Griffiths, and Zhiyuan Li for discussions, suggestions, and proof-reading at various stages of the paper.

\section*{Impact Statement}
This paper studies \emph{contextual drag}: a failure mode where conditioning on an incorrect prior attempt can bias subsequent reasoning toward the same errors, even when incorrectness is signaled or recognized. By characterizing the error mode and by exploring mitigations (e.g., training and test-time strategies to reduce reliance on erroneous context), our work may improve the reliability of iterative self-improvement pipelines and downstream systems that depend on multi-step reasoning, verification, or tool use, potentially reducing silent error propagation in applications.

The primary potential risk would be dual-use: our analysis may help adversaries more deliberately steer models using misleading in-context “solutions” or craft stronger prompt-injection style attacks by exploiting anchoring dynamics. To mitigate this, we emphasize robustness-oriented evaluation, document limitations and failure cases, and frame proposed techniques as defenses that can be incorporated into safer prompting and training practices. Our experiments use standard benchmarks and model-generated traces, without collecting personal data.


\bibliography{main}
\bibliographystyle{icml2026}

%%%%%%%%%%%%%%%%%%%%%%%%%%%%%%%%%%%%%%%%%%%%%%%%%%%%%%%%%%%%%%%%%%%%%%%%%%%%%%%
%%%%%%%%%%%%%%%%%%%%%%%%%%%%%%%%%%%%%%%%%%%%%%%%%%%%%%%%%%%%%%%%%%%%%%%%%%%%%%%
% APPENDIX
%%%%%%%%%%%%%%%%%%%%%%%%%%%%%%%%%%%%%%%%%%%%%%%%%%%%%%%%%%%%%%%%%%%%%%%%%%%%%%%
%%%%%%%%%%%%%%%%%%%%%%%%%%%%%%%%%%%%%%%%%%%%%%%%%%%%%%%%%%%%%%%%%%%%%%%%%%%%%%%
\newpage
\appendix
\onecolumn

\section{Additional Details of Contextual Drag Evaluation}
\label{app:evals}

\begin{figure}[H]
    \centering
    \includegraphics[width=0.95\linewidth]{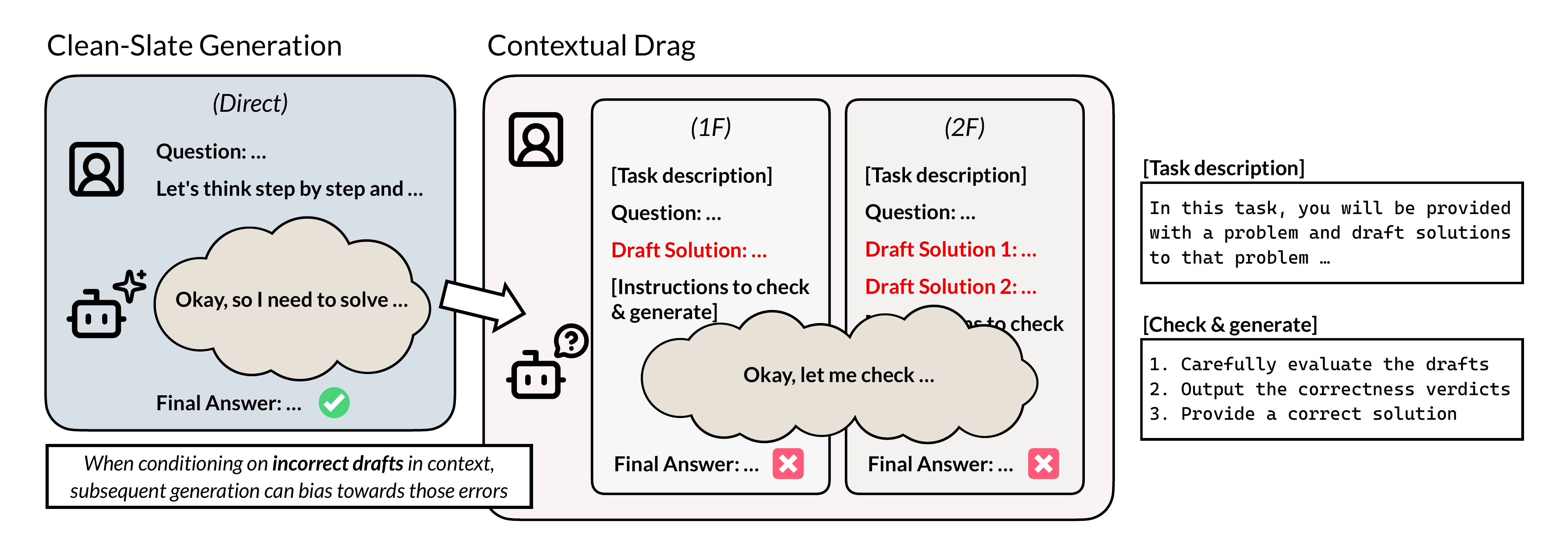}
    \caption{\textbf{Contextual drag evaluation setup}: We measure (1) {\direct} performance of the model's clean-slate generation, and (2) {\onef}/{\twof} performance of the model's generation conditioned on $1$ or $2$ incorrect draft solutions. {\direct} evaluation uses the standard chain-of-thought prompting. For {\onef}/{\twof} evaluations, the models is instructed to perform a step-by-step verification then produce a correctness verdict for each in-context draft, then produce a new solution.}
    \label{fig:eval-setup}
\end{figure}

\cref{fig:eval-setup} illustrates the evaluation setup for contextual drag in \cref{tab:main-table}. 

\subsection{Prompt Templates}
\label{app:evals_prompt_templates}
We provide the prompt templates used to evaluate contextual drag in \cref{tab:main-table} below. We perform $2$ types of inference:
\begin{enumerate}
    \item {\direct}: given only the problem, the model is asked to solve the problem with chain-of-thoughts (\hyperref[template-direct]{{\direct} template});\\
    \begin{promptbox}{template-direct}{{\direct}}
    ~\\
    \textbf{\{problem\}} Let's think step by step and please wrap your final answer in \textbackslash{}boxed\{\}.\\
\end{promptbox}
    \item {\onef}/{\twof}: given the problem and one or two incorrect model-generated draft solutions, the model is asked to check the correctness of the draft and generate a new solution (\hyperref[template-1f]{{\onef} template}, \hyperref[template-2f]{{\twof} template}). We verify correctness by parsing the final answer and compare the model's answer with ground truth via symbolic verifier \cite{kydlicek2024mathverify}. We remove all the thinking traces wrapped in \texttt{<think></think>} from the drafts to keep the context length manageable.\\
    \begin{promptbox}{template-1f}{{\onef}}
    ~\\
    In this task, you will be provided with a problem and one draft solution to that problem. For the draft, walk through the solution step-by-step and check for any fatal mistakes in the reasoning (for example: incorrect use of theorems or facts, incorrect calculations, etc.). Please note that once you spot a fatal error in a draft's reasoning, you **MUST** stop checking the remaining steps of that draft and move on.\\

    For the draft, output an overall verdict on the correctness of the draft using the following tag format:\\

    <overall\_verdict>[Correct / Incorrect]</overall\_verdict>\\
    
    While evaluating individual steps, be strict about algebra, logic, and correct theorem application. If the draft is incorrect, write a complete, correct solution (you may reuse sound ideas from the draft). If the draft is correct, you may adopt and polish it as the final solution.\\

    Conclude by presenting the final answer only inside \textbackslash{}boxed\{\} (i.e., the final answer should appear only inside a LaTeX `\textbackslash{}boxed\{\}`).\\
    \\
    The problem is as follows:\\
    -- beginning of problem --\\
    \textbf{\{problem\}}\\
    -- end of problem --\\

    Here is the draft solution you need to consider:\\

    \#\#\# Solution:\\
    -- beginning of the draft --\\
    \textbf{\{draft1\}}\\
    -- end of the draft --\\

    Please:\\
    1. Carefully evaluate the draft solution step-by-step. Once you spot a fatal error in a draft's reasoning, you **MUST** stop checking the remaining steps of that draft and move on. \\
    2. Output the correctness verdicts for the draft.\\
    3. Provide a correct solution to the problem with complete reasoning steps that lead to the answer.\\
    \\
    Remember: conclude with the final answer only in \textbackslash{}boxed\{\}.
\end{promptbox}

\begin{promptbox}{template-2f}{{\twof}}
    ~\\
    In this task, you will be provided with a problem and two draft solutions to that problem. For each draft, walk through the solution step-by-step and check for any fatal mistakes in the reasoning (for example: incorrect use of theorems or facts, incorrect calculations, etc.). Please note that once you spot a fatal error in a draft's reasoning, you **MUST** stop checking the remaining steps of that draft and move on to the next draft.\\

    For each draft (Solution 1 and Solution 2), output an overall verdict on the correctness of the draft using the following tag format:\\
    
    <overall\_verdict\_1>[Correct / Incorrect]</overall\_verdict\_1>\\
    <overall\_verdict\_2>[Correct / Incorrect]</overall\_verdict\_2>\\
    
    While evaluating individual steps, be strict about algebra, logic, and correct theorem application. If both drafts are incorrect, write a complete, correct solution (you may reuse sound ideas from the drafts). If a draft is correct, you may adopt and polish it as the final solution.\\
    
    Conclude by presenting the final answer only inside \textbackslash{}boxed\{\} (i.e., the final answer should appear only inside a LaTeX `\textbackslash{}boxed\{\}`).\\
    
    The problem is as follows:\\
    -- beginning of problem --\\
    \textbf{\{problem\}}\\
    -- end of problem --\\
    
    Here are the two draft solutions you need to consider:\\
    
    \#\#\# Solution 1:\\
    -- beginning of the first draft --\\
    \textbf{\{draft1\}}\\
    -- end of the first draft --\\
    
    \#\#\# Solution 2:\\
    -- beginning of the second draft --\\
    \textbf{\{draft2\}}\\
    -- end of the second draft --\\
    
    Please:\\
    1. Carefully evaluate the two draft solutions step-by-step. Once you spot a fatal error in a draft's reasoning, you **MUST** stop checking the remaining steps of that draft and move on to the next draft. \\
    2. Output the correctness verdicts for the drafts, respectively.\\
    3. Provide a correct solution to the problem with complete reasoning steps that lead to the answer.\\
    
    Remember: conclude with the final answer only in \textbackslash{}boxed\{\}.
\end{promptbox}
\end{enumerate} 

\subsection{Inference Configurations}

We followed the official recommended generation configurations for model inference, as listed in \cref{tab:generation_config}. For efficient inference, we use vLLM project version 0.10.2 \cite{kwon2023efficient}. We use a total context size of $32{,}768$ for all models except for the Nemotron family, which has a recommended max output token limit of $65{,}536$. As later shown in \cref{tab:initial-sampling}, this does not give extra advantages to the Nemotron models as all open-weight models are able to finish most of generations within the specified max token limits with comparable completion rates.

\begin{table}[H]
    \centering\small
    \caption{Inference configurations for all evaluated models. We report the maximum context length and the sampling parameters used during generation. For models where \texttt{top\_k} sampling is not being used, we denote it as ``--''.}
    \begin{tabular}{lcccl}
\toprule
\toprule
Model & \#tokens & Temp & Top-$p$ / Top-$k$ & Huggingface Path \\
\midrule
Qwen3-8B (Thinking)  & $32{,}768$ & $0.6$ & $0.95$ / $20$ &
\href{https://huggingface.co/Qwen/Qwen3-8B}{\texttt{Qwen/Qwen3-8B}} \\

Qwen3-32B (Thinking) & $32{,}768$ & $0.6$ & $0.95$ / $20$ &
\href{https://huggingface.co/Qwen/Qwen3-32B}{\texttt{Qwen/Qwen3-32B}} \\

Nemotron-7B         & $65{,}536$ & $0.6$ & $0.95$ / $-$ &
\href{https://huggingface.co/nvidia/OpenReasoning-Nemotron-7B}{\texttt{nvidia/OpenReasoning-Nemotron-7B}} \\

Nemotron-32B        & $65{,}536$ & $0.6$ & $0.95$ / $-$ &
\href{https://huggingface.co/nvidia/OpenReasoning-Nemotron-32B}{\texttt{nvidia/OpenReasoning-Nemotron-32B}} \\

GPT-OSS-20B         & $32{,}768$ & $1.0$ & $1.00$ / $40$ &
\href{https://huggingface.co/openai/gpt-oss-20b}{\texttt{openai/gpt-oss-20b}} \\

GPT-OSS-120B        & $32{,}768$ & $1.0$ & $1.00$ / $40$ &
\href{https://huggingface.co/openai/gpt-oss-120b}{\texttt{openai/gpt-oss-120b}} \\

LlamaR1-8B          & $32{,}768$ & $0.6$ & $0.95$ / $-$ &
\href{https://huggingface.co/deepseek-ai/DeepSeek-R1-Distill-Llama-8B}{\texttt{deepseek-ai/DeepSeek-R1-Distill-Llama-8B}} \\

QwenR1-7B           & $32{,}768$ & $0.6$ & $0.95$ / $-$ &
\href{https://huggingface.co/deepseek-ai/DeepSeek-R1-Distill-Qwen-7B}{\texttt{deepseek-ai/DeepSeek-R1-Distill-Qwen-7B}} \\
\bottomrule
\end{tabular}
    \label{tab:generation_config}
\end{table}

\subsection{Initial Evaluation: Selection of Anchor Models, In-Context Drafts, and Evaluation Problems}
Our criteria for constructing the evaluation set for contextual drag is to avoid confounding performance drops caused by trivial, low-quality in-context drafts observed by \citet{li2025rethinking}. To this end, we choose a set of three \emph{anchor models}, which have the top-$3$ clean-slate performance among all open-weight models for each task. 

Note that under this setup, the in-context draft used in the contextual drag evaluation may either come from the evaluated model itself or from other models, depending on whether the evaluated model is chosen as one of the anchor models. Therefore, the evaluation reflects contextual drag in both self-refinement and multi-agent aggregation settings.

Finally, we select the evaluation problems based on their \emph{reasonable hardness} to represent the most realistic use case in test-time scaling, such as iterative refinement. Therefore, our criterion for this evaluation subset is the set of questions where the anchor models exhibit both successes and failures: each selected question has at least two correct and at least two incorrect clean-slate responses from the anchor models. We note that these subsets are also of the most interest for test-time scaling as they correspond to the gap between pass@$1$ and pass@$k$ (for larger $k$).

\cref{tab:initial-sampling} shows the initial evaluation results of the clean-slate performance, which gives the following lists of anchor models:

\begin{table}[H]
    \centering
    \caption{\textbf{Initial evaluation results of open-weight models on all evaluation benchmarks}: We first run clean-slate generation for the set of open-weight models on the original data of all benchmarks to select the \underline{top-3 performant models} (underlined), whose responses are used as the in-context drafts for later evaluation of contextual drag.}\small

    \begin{tabular}{l | c c c c}
    \toprule
    \toprule
    Model & AIME24 & AIME25 & HMMT24 & HMMT25 \\
    \midrule
    GPT-OSS-120B & $79.79$ \tiny($\pm0.45$) & \underline{$78.54$ \tiny($\pm0.41$)} & \underline{$52.92$ \tiny($\pm0.48$)} & \underline{$56.04$ \tiny($\pm0.41$)} \\
    GPT-OSS-20B & $72.71$ \tiny($\pm0.46$) & $68.33$ \tiny($\pm0.60$) & $46.88$ \tiny($\pm0.41$) & $54.37$ \tiny($\pm0.46$) \\
    Nemotron-32B & \underline{$87.92$ \tiny($\pm0.35$)} & \underline{$83.75$ \tiny($\pm0.35$)} & \underline{$59.79$ \tiny($\pm0.54$)} & \underline{$69.79$ \tiny($\pm0.34$)} \\
    Nemotron-7B & \underline{$81.88$ \tiny($\pm0.30$)} & \underline{$79.17$ \tiny($\pm0.36$)} & \underline{$52.71$ \tiny($\pm0.49$)} & \underline{$57.50$ \tiny($\pm0.55$)} \\
    Qwen3-32B & \underline{$80.00$ \tiny($\pm0.33$)} & $71.67$ \tiny($\pm0.45$) & $47.29$ \tiny($\pm0.51$) & $48.12$ \tiny($\pm0.48$) \\
    Qwen3-8B & $75.00$ \tiny($\pm0.28$) & $67.08$ \tiny($\pm0.39$) & $38.12$ \tiny($\pm0.28$) & $42.92$ \tiny($\pm0.39$) \\
    LlamaR1-8B & $43.33$ \tiny($\pm0.71$) & $32.29$ \tiny($\pm0.42$) & $23.54$ \tiny($\pm0.31$) & $20.21$ \tiny($\pm0.39$) \\
    QwenR1-7B & $54.17$ \tiny($\pm0.71$) & $36.25$ \tiny($\pm0.48$) & $28.96$ \tiny($\pm0.46$) & $26.67$ \tiny($\pm0.47$) \\
    \midrule
    Model & \textsc{GPQA-Diamond} & \textsc{MMLU-Redux} & \textsc{CRUXEval-I} & \textsc{Game of 24} \\
    \midrule
    GPT-OSS-120B & \underline{$69.35$ \tiny($\pm0.07$)} & \underline{$90.90$ \tiny($\pm0.00$)} & $89.51$ \tiny($\pm0.07$) & $85.95$ \tiny($\pm0.01$) \\
    GPT-OSS-20B & $58.59$ \tiny($\pm0.08$) & $87.05$ \tiny($\pm0.00$) & $71.97$ \tiny($\pm0.21$) & $86.28$ \tiny($\pm0.02$) \\
    Nemotron-32B & \underline{$75.19$ \tiny($\pm0.04$)} & \underline{$90.87$ \tiny($\pm0.00$)} & \underline{$93.27$ \tiny($\pm0.01$)} & \underline{$89.96$ \tiny($\pm0.01$)} \\
    Nemotron-7B & $62.63$ \tiny($\pm0.07$) & $84.24$ \tiny($\pm0.00$) & $55.13$ \tiny($\pm0.11$) & $87.02$ \tiny($\pm0.01$) \\
    Qwen3-32B & \underline{$69.16$ \tiny($\pm0.06$)} & \underline{$91.64$ \tiny($\pm0.00$)} & \underline{$94.11$ \tiny($\pm0.01$)} & \underline{$89.15$ \tiny($\pm0.01$)} \\
    Qwen3-8B & $60.64$ \tiny($\pm0.07$) & $88.66$ \tiny($\pm0.00$) & \underline{$90.91$ \tiny($\pm0.01$)} & \underline{$88.11$ \tiny($\pm0.01$)} \\
    LlamaR1-8B & $42.87$ \tiny($\pm0.08$) & $66.72$ \tiny($\pm0.01$) & $56.16$ \tiny($\pm0.02$) & $52.41$ \tiny($\pm0.02$) \\
    QwenR1-7B & $34.69$ \tiny($\pm0.04$) & $61.19$ \tiny($\pm0.01$) & $67.57$ \tiny($\pm0.02$) & $75.19$ \tiny($\pm0.01$) \\
    \bottomrule
    \end{tabular}
    
    \label{tab:initial-sampling}
\end{table}

\begin{table}[H]
    \centering\small
    \caption{List of anchor models whose clean-slate responses are used as in-context drafts in the evaluation of contextual drag in \cref{tab:main-table}.}
    \label{tab:anchor-models}
    \begin{tabular}{l|l}
    \toprule
    \toprule
    Task & Anchor Models \\
    \midrule
    \textsc{AIME24} & Nemotron-7B/32B, Qwen3-32B \\
    \textsc{AIME25} & GPT-OSS-120B, Nemotron-7B/32B \\
    \textsc{HMMT24} & GPT-OSS-120B, Nemotron-7B/32B \\
    \textsc{HMMT25} & GPT-OSS-120B, Nemotron-7B/32B \\
    \textsc{GPQA-Diamond} & GPT-OSS-120B, Nemotron-32B, Qwen3-32B \\
    \textsc{MMLU-Redux} & GPT-OSS-120B, Nemotron-32B, Qwen3-32B \\
    \textsc{CRUXEval-I} & Nemotron-32B, Qwen3-8B/32B \\
    \textsc{Game of 24} & Nemotron-32B, Qwen3-8B/32B \\
    \bottomrule
    \end{tabular}
    
\end{table}

\newpage
\subsection{Full Results of Contextual Drag Evaluation}

\cref{tab:main-table-full-pass1} and \cref{tab:main-eval-full-pass5} present the full pass@$1$ and pass@$5$ evaluation results for \cref{tab:main-table}. We conduct $16$ independent inferences for each problem instance. We report the relative performance changes with respect to {\direct} performance and provide the confidence intervals for the mean performance.

We note that there are no substantial differences between the pass@$1$ and pass@$5$ results, demonstrating that contextual drag leads to strong, consistent performance degradation that is hard to recover from increasing number of independent samplings.

\begin{table}[H]
    \centering
    \caption{Full average pass@$1$ results (with $95$\% confidence interval in subscripts) for contextual drag evaluation in \cref{tab:main-table}.}
    
    \begin{adjustbox}{width=\textwidth}
    \begin{tabular}{l | c c c c c c}
    \toprule
    \toprule
         & \multicolumn{3}{c}{\aime{24}} & \multicolumn{3}{c}{\aime{25}} \\
\cmidrule(lr){2-4} \cmidrule(lr){5-7}
     & {\direct} & {\hlcellb {\onef}} & {\hlcella {\twof}} & {\direct} & {\hlcellb {\onef}} & {\hlcella {\twof}} \\
        \midrule

        GPT-OSS-120B & $66.25$ {\tiny$\pm1.72$}  & \hlcellb $43.75$ {\tiny$\pm0.51$} (\perfdecrease{-$22.50$}) & \hlcella $43.75$ {\tiny$\pm2.04$} (\perfdecrease{-$22.50$}) & $66.83$ {\tiny$\pm1.47$}  & \hlcellb $55.29$ {\tiny$\pm1.04$} (\perfdecrease{-$11.54$}) & \hlcella $40.38$ {\tiny$\pm1.31$} (\perfdecrease{-$26.44$}) \\
        GPT-OSS-20B & $51.88$ {\tiny$\pm1.47$}  & \hlcellb $17.50$ {\tiny$\pm1.60$} (\perfdecrease{-$34.38$}) & \hlcella $21.25$ {\tiny$\pm2.32$} (\perfdecrease{-$30.63$}) & $51.92$ {\tiny$\pm1.55$}  & \hlcellb $18.75$ {\tiny$\pm0.84$} (\perfdecrease{-$33.17$}) & \hlcella $20.67$ {\tiny$\pm1.51$} (\perfdecrease{-$31.25$}) \\
        Nemotron-32B & $83.75$ {\tiny$\pm1.81$}  & \hlcellb $67.50$ {\tiny$\pm1.28$} (\perfdecrease{-$16.25$}) & \hlcella $63.13$ {\tiny$\pm1.42$} (\perfdecrease{-$20.62$}) & $78.37$ {\tiny$\pm1.18$}  & \hlcellb $61.54$ {\tiny$\pm0.91$} (\perfdecrease{-$16.83$}) & \hlcella $50.96$ {\tiny$\pm1.47$} (\perfdecrease{-$27.40$}) \\
        Nemotron-7B & $67.50$ {\tiny$\pm1.28$}  & \hlcellb $53.12$ {\tiny$\pm1.22$} (\perfdecrease{-$14.38$}) & \hlcella $58.75$ {\tiny$\pm1.81$} (\perfdecrease{-$8.75$}) & $67.79$ {\tiny$\pm1.12$}  & \hlcellb $52.40$ {\tiny$\pm0.69$} (\perfdecrease{-$15.38$}) & \hlcella $37.02$ {\tiny$\pm1.29$} (\perfdecrease{-$30.77$}) \\
        Qwen3-32B & $65.00$ {\tiny$\pm2.05$}  & \hlcellb $41.25$ {\tiny$\pm0.84$} (\perfdecrease{-$23.75$}) & \hlcella $30.00$ {\tiny$\pm1.55$} (\perfdecrease{-$35.00$}) & $54.81$ {\tiny$\pm1.47$}  & \hlcellb $31.25$ {\tiny$\pm0.66$} (\perfdecrease{-$23.56$}) & \hlcella $31.25$ {\tiny$\pm0.78$} (\perfdecrease{-$23.56$}) \\
        Qwen3-8B & $55.62$ {\tiny$\pm2.12$}  & \hlcellb $30.63$ {\tiny$\pm0.90$} (\perfdecrease{-$25.00$}) & \hlcella $20.62$ {\tiny$\pm1.69$} (\perfdecrease{-$35.00$}) & $45.67$ {\tiny$\pm1.20$}  & \hlcellb $20.19$ {\tiny$\pm0.77$} (\perfdecrease{-$25.48$}) & \hlcella $11.06$ {\tiny$\pm1.04$} (\perfdecrease{-$34.62$}) \\
        LlamaR1-8B & $21.25$ {\tiny$\pm1.89$}  & \hlcellb $8.13$ {\tiny$\pm0.94$} (\perfdecrease{-$13.12$}) & \hlcella $4.38$ {\tiny$\pm0.77$} (\perfdecrease{-$16.88$}) & $7.21$ {\tiny$\pm0.86$}  & \hlcellb $4.33$ {\tiny$\pm0.55$} (\perfdecrease{-$2.88$}) & \hlcella $2.88$ {\tiny$\pm0.63$} (\perfdecrease{-$4.33$}) \\
        QwenR1-7B & $33.13$ {\tiny$\pm2.37$}  & \hlcellb $11.88$ {\tiny$\pm1.09$} (\perfdecrease{-$21.25$}) & \hlcella $8.12$ {\tiny$\pm1.25$} (\perfdecrease{-$25.00$}) & $9.62$ {\tiny$\pm0.78$}  & \hlcellb $3.85$ {\tiny$\pm0.37$} (\perfdecrease{-$5.77$}) & \hlcella $0.96$ {\tiny$\pm0.35$} (\perfdecrease{-$8.65$}) \\
        \midrule

     & \multicolumn{3}{c}{\hmmt{24}} & \multicolumn{3}{c}{\hmmt{25}} \\
\cmidrule(lr){2-4} \cmidrule(lr){5-7}
     & {\direct} & {\hlcellb {\onef}} & {\hlcella {\twof}} & {\direct} & {\hlcellb {\onef}} & {\hlcella {\twof}} \\
        \midrule

        GPT-OSS-120B & $48.83$ {\tiny$\pm1.22$}  & \hlcellb $39.84$ {\tiny$\pm0.60$} (\perfdecrease{-$8.98$}) & \hlcella $45.31$ {\tiny$\pm1.07$} (\perfdecrease{-$3.52$}) & $61.01$ {\tiny$\pm0.70$}  & \hlcellb $48.51$ {\tiny$\pm0.55$} (\perfdecrease{-$12.50$}) & \hlcella $49.11$ {\tiny$\pm0.69$} (\perfdecrease{-$11.90$}) \\
        GPT-OSS-20B & $40.23$ {\tiny$\pm1.08$}  & \hlcellb $13.28$ {\tiny$\pm0.83$} (\perfdecrease{-$26.95$}) & \hlcella $17.97$ {\tiny$\pm1.21$} (\perfdecrease{-$22.27$}) & $58.63$ {\tiny$\pm0.78$}  & \hlcellb $20.54$ {\tiny$\pm0.68$} (\perfdecrease{-$38.10$}) & \hlcella $24.70$ {\tiny$\pm1.02$} (\perfdecrease{-$33.93$}) \\
        Nemotron-32B & $61.72$ {\tiny$\pm1.40$}  & \hlcellb $51.95$ {\tiny$\pm0.54$} (\perfdecrease{-$9.77$}) & \hlcella $50.78$ {\tiny$\pm0.85$} (\perfdecrease{-$10.94$}) & $80.36$ {\tiny$\pm0.62$}  & \hlcellb $65.48$ {\tiny$\pm0.55$} (\perfdecrease{-$14.88$}) & \hlcella $64.58$ {\tiny$\pm0.70$} (\perfdecrease{-$15.77$}) \\
        Nemotron-7B & $49.22$ {\tiny$\pm1.14$}  & \hlcellb $35.94$ {\tiny$\pm0.66$} (\perfdecrease{-$13.28$}) & \hlcella $40.62$ {\tiny$\pm0.77$} (\perfdecrease{-$8.59$}) & $63.10$ {\tiny$\pm0.94$}  & \hlcellb $54.17$ {\tiny$\pm0.31$} (\perfdecrease{-$8.93$}) & \hlcella $52.38$ {\tiny$\pm0.74$} (\perfdecrease{-$10.71$}) \\
        Qwen3-32B & $39.06$ {\tiny$\pm1.34$}  & \hlcellb $35.55$ {\tiny$\pm0.43$} (\perfdecrease{-$3.52$}) & \hlcella $40.23$ {\tiny$\pm0.76$} (\perfincrease{+$1.17$}) & $49.70$ {\tiny$\pm0.82$}  & \hlcellb $36.90$ {\tiny$\pm0.34$} (\perfdecrease{-$12.80$}) & \hlcella $41.37$ {\tiny$\pm0.67$} (\perfdecrease{-$8.33$}) \\
        Qwen3-8B & $21.88$ {\tiny$\pm0.72$}  & \hlcellb $22.27$ {\tiny$\pm0.80$} (\perfincrease{+$0.39$}) & \hlcella $26.95$ {\tiny$\pm0.89$} (\perfincrease{+$5.08$}) & $42.26$ {\tiny$\pm0.67$}  & \hlcellb $34.23$ {\tiny$\pm0.46$} (\perfdecrease{-$8.04$}) & \hlcella $32.74$ {\tiny$\pm0.67$} (\perfdecrease{-$9.52$}) \\
        LlamaR1-8B & $5.08$ {\tiny$\pm0.40$}  & \hlcellb $7.42$ {\tiny$\pm0.38$} (\perfincrease{+$2.34$}) & \hlcella $9.77$ {\tiny$\pm0.66$} (\perfincrease{+$4.69$}) & $10.42$ {\tiny$\pm0.68$}  & \hlcellb $7.44$ {\tiny$\pm0.37$} (\perfdecrease{-$2.98$}) & \hlcella $10.42$ {\tiny$\pm0.58$} (\perfincrease{+$0.00$}) \\
        QwenR1-7B & $8.59$ {\tiny$\pm0.81$}  & \hlcellb $6.25$ {\tiny$\pm0.43$} (\perfdecrease{-$2.34$}) & \hlcella $6.25$ {\tiny$\pm0.72$} (\perfdecrease{-$2.34$}) & $19.64$ {\tiny$\pm0.74$}  & \hlcellb $10.12$ {\tiny$\pm0.38$} (\perfdecrease{-$9.52$}) & \hlcella $13.69$ {\tiny$\pm0.69$} (\perfdecrease{-$5.95$}) \\
        \midrule

     & \multicolumn{3}{c}{\textsc{\gpqa}} & \multicolumn{3}{c}{\textsc{\mmlu}} \\
\cmidrule(lr){2-4} \cmidrule(lr){5-7}
     & {\direct} & {\hlcellb {\onef}} & {\hlcella {\twof}} & {\direct} & {\hlcellb {\onef}} & {\hlcella {\twof}} \\
        \midrule

        GPT-OSS-120B & $58.24$ {\tiny$\pm0.13$}  & \hlcellb $35.70$ {\tiny$\pm0.08$} (\perfdecrease{-$22.54$}) & \hlcella $31.91$ {\tiny$\pm0.09$} (\perfdecrease{-$26.33$}) & $67.59$ {\tiny$\pm0.02$}  & \hlcellb $42.26$ {\tiny$\pm0.02$} (\perfdecrease{-$25.33$}) & \hlcella $31.61$ {\tiny$\pm0.03$} (\perfdecrease{-$35.97$}) \\
        GPT-OSS-20B & $46.97$ {\tiny$\pm0.14$}  & \hlcellb $16.86$ {\tiny$\pm0.05$} (\perfdecrease{-$30.11$}) & \hlcella $13.16$ {\tiny$\pm0.08$} (\perfdecrease{-$33.81$}) & $58.86$ {\tiny$\pm0.04$}  & \hlcellb $22.73$ {\tiny$\pm0.01$} (\perfdecrease{-$36.14$}) & \hlcella $12.83$ {\tiny$\pm0.02$} (\perfdecrease{-$46.03$}) \\
        Nemotron-32B & $66.81$ {\tiny$\pm0.08$}  & \hlcellb $49.43$ {\tiny$\pm0.10$} (\perfdecrease{-$17.38$}) & \hlcella $37.22$ {\tiny$\pm0.09$} (\perfdecrease{-$29.59$}) & $67.25$ {\tiny$\pm0.02$}  & \hlcellb $32.74$ {\tiny$\pm0.01$} (\perfdecrease{-$34.50$}) & \hlcella $20.09$ {\tiny$\pm0.02$} (\perfdecrease{-$47.16$}) \\
        Nemotron-7B & $51.61$ {\tiny$\pm0.11$}  & \hlcellb $35.27$ {\tiny$\pm0.07$} (\perfdecrease{-$16.34$}) & \hlcella $29.40$ {\tiny$\pm0.07$} (\perfdecrease{-$22.21$}) & $56.46$ {\tiny$\pm0.02$}  & \hlcellb $23.58$ {\tiny$\pm0.02$} (\perfdecrease{-$32.88$}) & \hlcella $14.02$ {\tiny$\pm0.02$} (\perfdecrease{-$42.44$}) \\
        Qwen3-32B & $57.62$ {\tiny$\pm0.10$}  & \hlcellb $34.19$ {\tiny$\pm0.05$} (\perfdecrease{-$23.44$}) & \hlcella $23.30$ {\tiny$\pm0.10$} (\perfdecrease{-$34.33$}) & $70.62$ {\tiny$\pm0.02$}  & \hlcellb $40.53$ {\tiny$\pm0.02$} (\perfdecrease{-$30.09$}) & \hlcella $21.97$ {\tiny$\pm0.02$} (\perfdecrease{-$48.65$}) \\
        Qwen3-8B & $47.77$ {\tiny$\pm0.09$}  & \hlcellb $27.13$ {\tiny$\pm0.08$} (\perfdecrease{-$20.64$}) & \hlcella $20.45$ {\tiny$\pm0.06$} (\perfdecrease{-$27.32$}) & $61.65$ {\tiny$\pm0.03$}  & \hlcellb $27.76$ {\tiny$\pm0.02$} (\perfdecrease{-$33.90$}) & \hlcella $17.21$ {\tiny$\pm0.01$} (\perfdecrease{-$44.44$}) \\
        LlamaR1-8B & $31.53$ {\tiny$\pm0.16$}  & \hlcellb $20.69$ {\tiny$\pm0.05$} (\perfdecrease{-$10.84$}) & \hlcella $17.85$ {\tiny$\pm0.06$} (\perfdecrease{-$13.68$}) & $41.17$ {\tiny$\pm0.03$}  & \hlcellb $13.72$ {\tiny$\pm0.02$} (\perfdecrease{-$27.45$}) & \hlcella $8.89$ {\tiny$\pm0.02$} (\perfdecrease{-$32.28$}) \\
        QwenR1-7B & $21.26$ {\tiny$\pm0.08$}  & \hlcellb $14.73$ {\tiny$\pm0.09$} (\perfdecrease{-$6.53$}) & \hlcella $15.20$ {\tiny$\pm0.06$} (\perfdecrease{-$6.06$}) & $36.60$ {\tiny$\pm0.04$}  & \hlcellb $13.62$ {\tiny$\pm0.01$} (\perfdecrease{-$22.98$}) & \hlcella $10.25$ {\tiny$\pm0.02$} (\perfdecrease{-$26.35$}) \\
        \midrule

     & \multicolumn{3}{c}{\textsc{\crux}} & \multicolumn{3}{c}{\textsc{\game}} \\
\cmidrule(lr){2-4} \cmidrule(lr){5-7}
     & {\direct} & {\hlcellb {\onef}} & {\hlcella {\twof}} & {\direct} & {\hlcellb {\onef}} & {\hlcella {\twof}} \\
        \midrule

        GPT-OSS-120B & $89.23$ {\tiny$\pm0.09$}  & \hlcellb $81.02$ {\tiny$\pm0.05$} (\perfdecrease{-$8.21$}) & \hlcella $77.89$ {\tiny$\pm0.06$} (\perfdecrease{-$11.34$}) & $77.34$ {\tiny$\pm0.03$}  & \hlcellb $56.94$ {\tiny$\pm0.03$} (\perfdecrease{-$20.40$}) & \hlcella $57.09$ {\tiny$\pm0.05$} (\perfdecrease{-$20.24$}) \\
        GPT-OSS-20B & $61.34$ {\tiny$\pm0.30$}  & \hlcellb $42.39$ {\tiny$\pm0.11$} (\perfdecrease{-$18.95$}) & \hlcella $40.53$ {\tiny$\pm0.13$} (\perfdecrease{-$20.81$}) & $78.30$ {\tiny$\pm0.03$}  & \hlcellb $41.61$ {\tiny$\pm0.05$} (\perfdecrease{-$36.70$}) & \hlcella $39.31$ {\tiny$\pm0.06$} (\perfdecrease{-$38.99$}) \\
        Nemotron-32B & $77.06$ {\tiny$\pm0.09$}  & \hlcellb $53.69$ {\tiny$\pm0.06$} (\perfdecrease{-$23.37$}) & \hlcella $54.19$ {\tiny$\pm0.09$} (\perfdecrease{-$22.87$}) & $79.54$ {\tiny$\pm0.03$}  & \hlcellb $56.68$ {\tiny$\pm0.03$} (\perfdecrease{-$22.86$}) & \hlcella $51.53$ {\tiny$\pm0.03$} (\perfdecrease{-$28.01$}) \\
        Nemotron-7B & $43.55$ {\tiny$\pm0.21$}  & \hlcellb $35.27$ {\tiny$\pm0.07$} (\perfdecrease{-$8.28$}) & \hlcella $31.32$ {\tiny$\pm0.09$} (\perfdecrease{-$12.23$}) & $75.35$ {\tiny$\pm0.02$}  & \hlcellb $45.97$ {\tiny$\pm0.03$} (\perfdecrease{-$29.38$}) & \hlcella $34.77$ {\tiny$\pm0.02$} (\perfdecrease{-$40.58$}) \\
        Qwen3-32B & $75.83$ {\tiny$\pm0.08$}  & \hlcellb $54.89$ {\tiny$\pm0.04$} (\perfdecrease{-$20.94$}) & \hlcella $50.90$ {\tiny$\pm0.09$} (\perfdecrease{-$24.93$}) & $78.48$ {\tiny$\pm0.03$}  & \hlcellb $43.40$ {\tiny$\pm0.03$} (\perfdecrease{-$35.09$}) & \hlcella $25.47$ {\tiny$\pm0.03$} (\perfdecrease{-$53.01$}) \\
        Qwen3-8B & $67.49$ {\tiny$\pm0.08$}  & \hlcellb $40.09$ {\tiny$\pm0.06$} (\perfdecrease{-$27.39$}) & \hlcella $36.20$ {\tiny$\pm0.08$} (\perfdecrease{-$31.28$}) & $76.29$ {\tiny$\pm0.03$}  & \hlcellb $32.92$ {\tiny$\pm0.02$} (\perfdecrease{-$43.38$}) & \hlcella $23.26$ {\tiny$\pm0.02$} (\perfdecrease{-$53.03$}) \\
        LlamaR1-8B & $30.32$ {\tiny$\pm0.08$}  & \hlcellb $23.84$ {\tiny$\pm0.07$} (\perfdecrease{-$6.48$}) & \hlcella $21.81$ {\tiny$\pm0.08$} (\perfdecrease{-$8.51$}) & $35.34$ {\tiny$\pm0.04$}  & \hlcellb $5.44$ {\tiny$\pm0.02$} (\perfdecrease{-$29.90$}) & \hlcella $3.37$ {\tiny$\pm0.01$} (\perfdecrease{-$31.97$}) \\
        QwenR1-7B & $36.84$ {\tiny$\pm0.11$}  & \hlcellb $22.74$ {\tiny$\pm0.06$} (\perfdecrease{-$14.10$}) & \hlcella $16.82$ {\tiny$\pm0.06$} (\perfdecrease{-$20.01$}) & $40.69$ {\tiny$\pm0.03$}  & \hlcellb $15.60$ {\tiny$\pm0.02$} (\perfdecrease{-$25.09$}) & \hlcella $8.49$ {\tiny$\pm0.02$} (\perfdecrease{-$32.20$}) \\
        \bottomrule
\end{tabular}
\end{adjustbox}

    \label{tab:main-table-full-pass1}
\end{table}
\newpage
\begin{table}[H]
    \centering
    \caption{Full average pass@$5$ results (with $95$\% confidence interval in subscripts) for contextual drag evaluation in \cref{tab:main-table}.}
    \begin{adjustbox}{width=\textwidth}
    \begin{tabular}{l | c c c c c c}
    \toprule
    \toprule
         & \multicolumn{3}{c}{\aime{24}} & \multicolumn{3}{c}{\aime{25}} \\
\cmidrule(lr){2-4} \cmidrule(lr){5-7}
     & {\direct} & {\hlcellb {\onef}} & {\hlcella {\twof}} & {\direct} & {\hlcellb {\onef}} & {\hlcella {\twof}} \\
        \midrule

        GPT-OSS-120B & $90.62$ {\tiny$\pm1.02$}  & \hlcellb $68.75$ {\tiny$\pm0.75$} (\perfdecrease{-$21.88$}) & \hlcella $73.12$ {\tiny$\pm0.90$} (\perfdecrease{-$17.50$}) & $92.31$ {\tiny$\pm0.74$}  & \hlcellb $83.65$ {\tiny$\pm0.43$} (\perfdecrease{-$8.65$}) & \hlcella $74.52$ {\tiny$\pm0.88$} (\perfdecrease{-$17.79$}) \\
        GPT-OSS-20B & $83.12$ {\tiny$\pm1.31$}  & \hlcellb $40.00$ {\tiny$\pm0.55$} (\perfdecrease{-$43.12$}) & \hlcella $46.25$ {\tiny$\pm1.44$} (\perfdecrease{-$36.88$}) & $78.85$ {\tiny$\pm0.69$}  & \hlcellb $58.65$ {\tiny$\pm1.33$} (\perfdecrease{-$20.19$}) & \hlcella $62.98$ {\tiny$\pm1.44$} (\perfdecrease{-$15.87$}) \\
        Nemotron-32B & $100.00$ {\tiny$\pm0.00$}  & \hlcellb $94.38$ {\tiny$\pm0.60$} (\perfdecrease{-$5.63$}) & \hlcella $92.50$ {\tiny$\pm1.40$} (\perfdecrease{-$7.50$}) & $91.83$ {\tiny$\pm0.25$}  & \hlcellb $88.94$ {\tiny$\pm0.31$} (\perfdecrease{-$2.88$}) & \hlcella $77.40$ {\tiny$\pm0.78$} (\perfdecrease{-$14.42$}) \\
        Nemotron-7B & $89.38$ {\tiny$\pm0.38$}  & \hlcellb $78.75$ {\tiny$\pm0.51$} (\perfdecrease{-$10.63$}) & \hlcella $88.12$ {\tiny$\pm1.13$} (\perfdecrease{-$1.25$}) & $87.50$ {\tiny$\pm0.63$}  & \hlcellb $86.54$ {\tiny$\pm0.52$} (\perfdecrease{-$0.96$}) & \hlcella $63.46$ {\tiny$\pm0.87$} (\perfdecrease{-$24.04$}) \\
        Qwen3-32B & $98.75$ {\tiny$\pm0.51$}  & \hlcellb $67.50$ {\tiny$\pm0.67$} (\perfdecrease{-$31.25$}) & \hlcella $56.88$ {\tiny$\pm0.72$} (\perfdecrease{-$41.88$}) & $79.33$ {\tiny$\pm0.48$}  & \hlcellb $63.46$ {\tiny$\pm0.83$} (\perfdecrease{-$15.87$}) & \hlcella $54.33$ {\tiny$\pm0.45$} (\perfdecrease{-$25.00$}) \\
        Qwen3-8B & $76.88$ {\tiny$\pm1.06$}  & \hlcellb $63.75$ {\tiny$\pm0.51$} (\perfdecrease{-$13.12$}) & \hlcella $54.37$ {\tiny$\pm1.64$} (\perfdecrease{-$22.50$}) & $68.75$ {\tiny$\pm0.69$}  & \hlcellb $51.44$ {\tiny$\pm0.25$} (\perfdecrease{-$17.31$}) & \hlcella $34.62$ {\tiny$\pm1.17$} (\perfdecrease{-$34.13$}) \\
        LlamaR1-8B & $48.12$ {\tiny$\pm1.57$}  & \hlcellb $26.88$ {\tiny$\pm0.38$} (\perfdecrease{-$21.25$}) & \hlcella $13.75$ {\tiny$\pm0.75$} (\perfdecrease{-$34.38$}) & $23.08$ {\tiny$\pm0.64$}  & \hlcellb $13.46$ {\tiny$\pm0.37$} (\perfdecrease{-$9.62$}) & \hlcella $14.42$ {\tiny$\pm1.38$} (\perfdecrease{-$8.65$}) \\
        QwenR1-7B & $59.38$ {\tiny$\pm0.38$}  & \hlcellb $38.75$ {\tiny$\pm0.64$} (\perfdecrease{-$20.62$}) & \hlcella $31.87$ {\tiny$\pm2.46$} (\perfdecrease{-$27.50$}) & $28.85$ {\tiny$\pm0.69$}  & \hlcellb $18.27$ {\tiny$\pm0.43$} (\perfdecrease{-$10.58$}) & \hlcella $4.81$ {\tiny$\pm0.51$} (\perfdecrease{-$24.04$}) \\
        \midrule

     & \multicolumn{3}{c}{\hmmt{24}} & \multicolumn{3}{c}{\hmmt{25}} \\
\cmidrule(lr){2-4} \cmidrule(lr){5-7}
     & {\direct} & {\hlcellb {\onef}} & {\hlcella {\twof}} & {\direct} & {\hlcellb {\onef}} & {\hlcella {\twof}} \\
        \midrule

        GPT-OSS-120B & $87.11$ {\tiny$\pm0.57$}  & \hlcellb $68.36$ {\tiny$\pm0.40$} (\perfdecrease{-$18.75$}) & \hlcella $75.00$ {\tiny$\pm0.61$} (\perfdecrease{-$12.11$}) & $87.20$ {\tiny$\pm0.50$}  & \hlcellb $72.62$ {\tiny$\pm0.31$} (\perfdecrease{-$14.58$}) & \hlcella $66.67$ {\tiny$\pm0.48$} (\perfdecrease{-$20.54$}) \\
        GPT-OSS-20B & $80.86$ {\tiny$\pm0.50$}  & \hlcellb $40.23$ {\tiny$\pm0.49$} (\perfdecrease{-$40.62$}) & \hlcella $47.66$ {\tiny$\pm1.24$} (\perfdecrease{-$33.20$}) & $86.01$ {\tiny$\pm0.85$}  & \hlcellb $52.98$ {\tiny$\pm0.35$} (\perfdecrease{-$33.04$}) & \hlcella $54.76$ {\tiny$\pm0.48$} (\perfdecrease{-$31.25$}) \\
        Nemotron-32B & $89.84$ {\tiny$\pm0.53$}  & \hlcellb $82.81$ {\tiny$\pm0.33$} (\perfdecrease{-$7.03$}) & \hlcella $85.16$ {\tiny$\pm0.53$} (\perfdecrease{-$4.69$}) & $96.43$ {\tiny$\pm0.22$}  & \hlcellb $86.31$ {\tiny$\pm0.17$} (\perfdecrease{-$10.12$}) & \hlcella $85.12$ {\tiny$\pm0.31$} (\perfdecrease{-$11.31$}) \\
        Nemotron-7B & $87.11$ {\tiny$\pm0.63$}  & \hlcellb $64.84$ {\tiny$\pm0.57$} (\perfdecrease{-$22.27$}) & \hlcella $71.48$ {\tiny$\pm0.71$} (\perfdecrease{-$15.62$}) & $91.37$ {\tiny$\pm0.32$}  & \hlcellb $78.57$ {\tiny$\pm0.25$} (\perfdecrease{-$12.80$}) & \hlcella $78.87$ {\tiny$\pm0.40$} (\perfdecrease{-$12.50$}) \\
        Qwen3-32B & $69.92$ {\tiny$\pm0.56$}  & \hlcellb $63.67$ {\tiny$\pm0.50$} (\perfdecrease{-$6.25$}) & \hlcella $60.94$ {\tiny$\pm0.33$} (\perfdecrease{-$8.98$}) & $72.92$ {\tiny$\pm0.47$}  & \hlcellb $55.95$ {\tiny$\pm0.13$} (\perfdecrease{-$16.96$}) & \hlcella $63.69$ {\tiny$\pm0.40$} (\perfdecrease{-$9.23$}) \\
        Qwen3-8B & $51.95$ {\tiny$\pm0.59$}  & \hlcellb $49.22$ {\tiny$\pm0.57$} (\perfdecrease{-$2.73$}) & \hlcella $51.56$ {\tiny$\pm0.43$} (\perfdecrease{-$0.39$}) & $63.99$ {\tiny$\pm0.74$}  & \hlcellb $51.19$ {\tiny$\pm0.40$} (\perfdecrease{-$12.80$}) & \hlcella $55.95$ {\tiny$\pm0.38$} (\perfdecrease{-$8.04$}) \\
        LlamaR1-8B & $19.14$ {\tiny$\pm0.69$}  & \hlcellb $21.88$ {\tiny$\pm0.27$} (\perfincrease{+$2.73$}) & \hlcella $28.91$ {\tiny$\pm0.46$} (\perfincrease{+$9.77$}) & $24.70$ {\tiny$\pm0.63$}  & \hlcellb $20.83$ {\tiny$\pm0.17$} (\perfdecrease{-$3.87$}) & \hlcella $28.27$ {\tiny$\pm0.55$} (\perfincrease{+$3.57$}) \\
        QwenR1-7B & $27.73$ {\tiny$\pm0.94$}  & \hlcellb $26.56$ {\tiny$\pm0.51$} (\perfdecrease{-$1.17$}) & \hlcella $25.00$ {\tiny$\pm0.81$} (\perfdecrease{-$2.73$}) & $48.81$ {\tiny$\pm0.64$}  & \hlcellb $29.46$ {\tiny$\pm0.22$} (\perfdecrease{-$19.35$}) & \hlcella $36.61$ {\tiny$\pm0.74$} (\perfdecrease{-$12.20$}) \\
        \midrule

     & \multicolumn{3}{c}{\textsc{\gpqa}} & \multicolumn{3}{c}{\textsc{\mmlu}} \\
\cmidrule(lr){2-4} \cmidrule(lr){5-7}
     & {\direct} & {\hlcellb {\onef}} & {\hlcella {\twof}} & {\direct} & {\hlcellb {\onef}} & {\hlcella {\twof}} \\
        \midrule

        GPT-OSS-120B & $84.23$ {\tiny$\pm0.08$}  & \hlcellb $60.27$ {\tiny$\pm0.05$} (\perfdecrease{-$23.96$}) & \hlcella $53.93$ {\tiny$\pm0.08$} (\perfdecrease{-$30.30$}) & $87.52$ {\tiny$\pm0.01$}  & \hlcellb $62.34$ {\tiny$\pm0.01$} (\perfdecrease{-$25.19$}) & \hlcella $49.91$ {\tiny$\pm0.02$} (\perfdecrease{-$37.61$}) \\
        GPT-OSS-20B & $80.07$ {\tiny$\pm0.05$}  & \hlcellb $42.90$ {\tiny$\pm0.08$} (\perfdecrease{-$37.17$}) & \hlcella $32.10$ {\tiny$\pm0.11$} (\perfdecrease{-$47.96$}) & $85.31$ {\tiny$\pm0.02$}  & \hlcellb $47.24$ {\tiny$\pm0.02$} (\perfdecrease{-$38.07$}) & \hlcella $30.00$ {\tiny$\pm0.02$} (\perfdecrease{-$55.32$}) \\
        Nemotron-32B & $90.34$ {\tiny$\pm0.04$}  & \hlcellb $69.93$ {\tiny$\pm0.07$} (\perfdecrease{-$20.41$}) & \hlcella $55.26$ {\tiny$\pm0.09$} (\perfdecrease{-$35.09$}) & $87.83$ {\tiny$\pm0.01$}  & \hlcellb $50.74$ {\tiny$\pm0.01$} (\perfdecrease{-$37.09$}) & \hlcella $32.28$ {\tiny$\pm0.01$} (\perfdecrease{-$55.55$}) \\
        Nemotron-7B & $79.64$ {\tiny$\pm0.08$}  & \hlcellb $59.14$ {\tiny$\pm0.03$} (\perfdecrease{-$20.50$}) & \hlcella $46.97$ {\tiny$\pm0.10$} (\perfdecrease{-$32.67$}) & $80.72$ {\tiny$\pm0.02$}  & \hlcellb $43.15$ {\tiny$\pm0.01$} (\perfdecrease{-$37.58$}) & \hlcella $26.93$ {\tiny$\pm0.01$} (\perfdecrease{-$53.79$}) \\
        Qwen3-32B & $79.97$ {\tiny$\pm0.07$}  & \hlcellb $54.31$ {\tiny$\pm0.04$} (\perfdecrease{-$25.66$}) & \hlcella $37.22$ {\tiny$\pm0.05$} (\perfdecrease{-$42.76$}) & $86.91$ {\tiny$\pm0.01$}  & \hlcellb $61.70$ {\tiny$\pm0.01$} (\perfdecrease{-$25.21$}) & \hlcella $36.99$ {\tiny$\pm0.02$} (\perfdecrease{-$49.91$}) \\
        Qwen3-8B & $66.76$ {\tiny$\pm0.09$}  & \hlcellb $45.03$ {\tiny$\pm0.05$} (\perfdecrease{-$21.73$}) & \hlcella $30.40$ {\tiny$\pm0.05$} (\perfdecrease{-$36.36$}) & $77.58$ {\tiny$\pm0.01$}  & \hlcellb $42.18$ {\tiny$\pm0.01$} (\perfdecrease{-$35.40$}) & \hlcella $29.82$ {\tiny$\pm0.01$} (\perfdecrease{-$47.76$}) \\
        LlamaR1-8B & $64.82$ {\tiny$\pm0.12$}  & \hlcellb $35.94$ {\tiny$\pm0.07$} (\perfdecrease{-$28.88$}) & \hlcella $27.89$ {\tiny$\pm0.06$} (\perfdecrease{-$36.93$}) & $71.67$ {\tiny$\pm0.02$}  & \hlcellb $30.23$ {\tiny$\pm0.02$} (\perfdecrease{-$41.44$}) & \hlcella $18.50$ {\tiny$\pm0.01$} (\perfdecrease{-$53.17$}) \\
        QwenR1-7B & $41.71$ {\tiny$\pm0.08$}  & \hlcellb $24.62$ {\tiny$\pm0.05$} (\perfdecrease{-$17.09$}) & \hlcella $23.82$ {\tiny$\pm0.08$} (\perfdecrease{-$17.90$}) & $64.59$ {\tiny$\pm0.03$}  & \hlcellb $29.54$ {\tiny$\pm0.02$} (\perfdecrease{-$35.05$}) & \hlcella $21.30$ {\tiny$\pm0.01$} (\perfdecrease{-$43.29$}) \\
        \midrule

     & \multicolumn{3}{c}{\textsc{\crux}} & \multicolumn{3}{c}{\textsc{\game}} \\
\cmidrule(lr){2-4} \cmidrule(lr){5-7}
     & {\direct} & {\hlcellb {\onef}} & {\hlcella {\twof}} & {\direct} & {\hlcellb {\onef}} & {\hlcella {\twof}} \\
        \midrule

        GPT-OSS-120B & $99.14$ {\tiny$\pm0.01$}  & \hlcellb $94.41$ {\tiny$\pm0.02$} (\perfdecrease{-$4.72$}) & \hlcella $93.28$ {\tiny$\pm0.03$} (\perfdecrease{-$5.85$}) & $97.43$ {\tiny$\pm0.01$}  & \hlcellb $83.65$ {\tiny$\pm0.02$} (\perfdecrease{-$13.77$}) & \hlcella $84.52$ {\tiny$\pm0.02$} (\perfdecrease{-$12.90$}) \\
        GPT-OSS-20B & $92.15$ {\tiny$\pm0.03$}  & \hlcellb $75.00$ {\tiny$\pm0.02$} (\perfdecrease{-$17.15$}) & \hlcella $72.31$ {\tiny$\pm0.05$} (\perfdecrease{-$19.85$}) & $98.76$ {\tiny$\pm0.01$}  & \hlcellb $72.78$ {\tiny$\pm0.03$} (\perfdecrease{-$25.98$}) & \hlcella $73.31$ {\tiny$\pm0.02$} (\perfdecrease{-$25.45$}) \\
        Nemotron-32B & $93.35$ {\tiny$\pm0.03$}  & \hlcellb $78.56$ {\tiny$\pm0.01$} (\perfdecrease{-$14.79$}) & \hlcella $74.83$ {\tiny$\pm0.04$} (\perfdecrease{-$18.52$}) & $95.65$ {\tiny$\pm0.01$}  & \hlcellb $86.54$ {\tiny$\pm0.01$} (\perfdecrease{-$9.10$}) & \hlcella $83.43$ {\tiny$\pm0.01$} (\perfdecrease{-$12.21$}) \\
        Nemotron-7B & $80.52$ {\tiny$\pm0.06$}  & \hlcellb $64.49$ {\tiny$\pm0.06$} (\perfdecrease{-$16.02$}) & \hlcella $60.94$ {\tiny$\pm0.08$} (\perfdecrease{-$19.58$}) & $94.71$ {\tiny$\pm0.01$}  & \hlcellb $80.84$ {\tiny$\pm0.01$} (\perfdecrease{-$13.87$}) & \hlcella $72.16$ {\tiny$\pm0.02$} (\perfdecrease{-$22.55$}) \\
        Qwen3-32B & $94.05$ {\tiny$\pm0.02$}  & \hlcellb $78.49$ {\tiny$\pm0.02$} (\perfdecrease{-$15.56$}) & \hlcella $74.27$ {\tiny$\pm0.04$} (\perfdecrease{-$19.78$}) & $97.51$ {\tiny$\pm0.01$}  & \hlcellb $77.56$ {\tiny$\pm0.02$} (\perfdecrease{-$19.96$}) & \hlcella $51.71$ {\tiny$\pm0.03$} (\perfdecrease{-$45.80$}) \\
        Qwen3-8B & $85.17$ {\tiny$\pm0.03$}  & \hlcellb $61.10$ {\tiny$\pm0.03$} (\perfdecrease{-$24.07$}) & \hlcella $59.74$ {\tiny$\pm0.09$} (\perfdecrease{-$25.43$}) & $95.33$ {\tiny$\pm0.01$}  & \hlcellb $59.60$ {\tiny$\pm0.01$} (\perfdecrease{-$35.73$}) & \hlcella $46.41$ {\tiny$\pm0.03$} (\perfdecrease{-$48.92$}) \\
        LlamaR1-8B & $56.85$ {\tiny$\pm0.06$}  & \hlcellb $45.15$ {\tiny$\pm0.03$} (\perfdecrease{-$11.70$}) & \hlcella $41.32$ {\tiny$\pm0.06$} (\perfdecrease{-$15.53$}) & $69.55$ {\tiny$\pm0.03$}  & \hlcellb $15.97$ {\tiny$\pm0.01$} (\perfdecrease{-$53.58$}) & \hlcella $9.37$ {\tiny$\pm0.02$} (\perfdecrease{-$60.18$}) \\
        QwenR1-7B & $66.39$ {\tiny$\pm0.09$}  & \hlcellb $43.55$ {\tiny$\pm0.03$} (\perfdecrease{-$22.84$}) & \hlcella $34.97$ {\tiny$\pm0.03$} (\perfdecrease{-$31.42$}) & $74.44$ {\tiny$\pm0.03$}  & \hlcellb $39.18$ {\tiny$\pm0.02$} (\perfdecrease{-$35.26$}) & \hlcella $23.43$ {\tiny$\pm0.02$} (\perfdecrease{-$51.01$}) \\
        \bottomrule
\end{tabular}
\end{adjustbox}
    \label{tab:main-eval-full-pass5}
\end{table}

\newpage
\section{Ablation Evaluation of Contextual Drag}
\label{app:ablations}

\subsection{Position and Verification}
\label{app:ablations-implementation}

For the \textbf{Position} ablation ({\pos}), we adjust (1) the position of the question to the beginning for more model attention and (2) the draft to the middle of the prompt for less distraction. For the \textbf{Verification} ablation ({\iv}), we remove the instruction to explicitly verify drafts. 

We provide the full details of implementation of the \textbf{Position} ({\pos}) and \textbf{Verification} ({\iv}) ablations below. To enable a fair comparison with the {\onef} results in \cref{tab:main-table}, all ablation prompt templates are adapted from \hyperref[template-1f]{1F template} with minimal changes. For easy reference, we \textcolor{blue}{highlight those changes in blue color}.

\paragraph{Position} This ablation investigates the effect of varying the position of the question and the draft solution. \citet{liu2024lost} points out that the model tends to make more use of the information at the beginning or end of the input context, which is known as the ``lost-in-the-middle'' effect. We adjust the position of (1) the question to the beginning and (2) the draft to the middle of the prompt. We expect that this change in position can reduce the use of the in-context draft, therefore alleviating contextual drag.\\

\begin{promptbox}{template-ablation-pos}{{\pos} ablation ({\onef})}
    ~\\
    In this task, you will be provided with a problem and one draft solution to that problem. \textcolor{blue}{The problem is as follows:\\
    -- beginning of problem --\\
    \textbf{\{problem\}}\\
    -- end of problem --\\}

    For the draft, walk through the solution step-by-step and check for any fatal mistakes in the reasoning (for example: incorrect use of theorems or facts, incorrect calculations, etc.). Please note that once you spot a fatal error in a draft's reasoning, you **MUST** stop checking the remaining steps of that draft and move on.\\

    For the draft, output an overall verdict on the correctness of the draft using the following tag format:\\

    <overall\_verdict>[Correct / Incorrect]</overall\_verdict>\\
    \\
    While evaluating individual steps, be strict about algebra, logic, and correct theorem application. \textcolor{blue}{Here is the draft solution you need to consider:\\}

    \textcolor{blue}{\#\#\# Solution:\\
    -- beginning of the draft --\\
    \textbf{\{draft1\}}\\
    -- end of the draft --\\}

    If the draft is incorrect, write a complete, correct solution (you may reuse sound ideas from the draft). If the draft is correct, you may adopt and polish it as the final solution.\\

    Conclude by presenting the final answer only inside \textbackslash{}boxed\{\} (i.e., the final answer should appear only inside a LaTeX `\textbackslash{}boxed\{\}`).\\

    Please:\\
    1. Carefully evaluate the draft solution step-by-step. Once you spot a fatal error in a draft's reasoning, you **MUST** stop checking the remaining steps of that draft and move on. \\
    2. Output the correctness verdicts for the draft.\\
    3. Provide a correct solution to the problem with complete reasoning steps that lead to the answer.\\

    Remember: conclude with the final answer only in \textbackslash{}boxed\{\}.
\end{promptbox}

\paragraph{Verification}

This ablation investigates the effect of performing an explicit check of the draft before generating the solution. We remove the corresponding instructions in the input prompt so that the model does not need to perform the explicit verification. While an explicit verification can help the model become aware of the mistakes contained in the draft, this also results in more verbose responses and therefore longer contexts that can lead to performance degradation.\\

\begin{promptbox}{template-ablation-iv}{{\iv} ablation (single draft)}
    ~\\
    \textcolor{blue}{In this task, you will be provided with a problem and one draft solution to that problem. If the draft is incorrect, write a complete, correct solution (you may reuse sound ideas from the draft). If the draft is correct, you may adopt and polish it as the final solution.}\\

    Conclude by presenting the final answer only inside \textbackslash{}boxed\{\} (i.e., the final answer should appear only inside a LaTeX `\textbackslash{}boxed\{\}`).\\

    The problem is as follows:\\
    -- beginning of problem --\\
    \textbf{\{problem\}}\\
    -- end of problem --\\

    Here is the draft solution you need to consider:\\

    \#\#\# Solution:\\
    -- beginning of the draft --\\
    \textbf{\{draft1\}}\\
    -- end of the draft --\\

    Please provide a correct solution to the problem with complete reasoning steps that lead to the answer.\\

    Remember: conclude with the final answer only in \textbackslash{}boxed\{\}.
\end{promptbox}

\cref{tab:ablations} shows the results of the \textbf{Position} ({\pos}) and \textbf{Verification} ({\iv}) ablation evaluation of GPT-OSS-120B and Nemotron-7B/32B, which exhibit more robustness to contextual drag in \cref{tab:main-table}, on HMMT25, GPQA, and CRUXEval-I.

\begin{table}[H]
    \centering
    \caption{\textbf{Position and Verification ablations}: \textsc{Direct} and \textsc{1F} statistics are copied from \cref{tab:main-table} for comparison. {\pos} and {\iv} denote to the \textbf{Position} and \textbf{Verification} ablations.}
    \label{tab:ablations}
    \small
    \begin{tabular}{l | c c c c}

    \toprule

    \toprule

     & \multicolumn{4}{c}{\hmmt{25}} \\
     \cmidrule(lr){2-5}
     & {\direct} & {\pos} & {\iv} & {\hlcellb 1F} \\

        \midrule

        GPT-OSS-120B & $61.01$\tiny{$\pm0.70$} & $53.57$\tiny{$\pm0.77$} & $52.98$\tiny{$\pm0.67$} & \hlcellb $48.51$\tiny{$\pm0.78$} \\

        Nemotron-32B & $80.36$\tiny{$\pm0.62$} & $73.81$\tiny{$\pm0.88$} & $67.56$\tiny{$\pm0.36$} & \hlcellb $65.48$\tiny{$\pm0.78$} \\

        Nemotron-7B & $63.10$\tiny{$\pm0.95$} & $52.38$\tiny{$\pm0.36$} & $51.19$\tiny{$\pm1.30$} & \hlcellb $54.17$\tiny{$\pm0.43$} \\

        \midrule

     & \multicolumn{4}{c}{\textsc{\gpqa}} \\
     \cmidrule(lr){2-5}
     & {\direct} & {\pos} & {\iv} & {\hlcellb 1F} \\

        \midrule

        GPT-OSS-120B & $58.24$\tiny{$\pm0.13$} & $42.52$\tiny{$\pm0.09$} & $38.16$\tiny{$\pm0.16$} & \hlcellb $35.70$\tiny{$\pm0.11$} \\

        Nemotron-32B & $66.81$\tiny{$\pm0.08$} & $49.95$\tiny{$\pm0.08$} & $39.44$\tiny{$\pm0.08$} & \hlcellb $49.43$\tiny{$\pm0.14$} \\

        Nemotron-7B & $51.61$\tiny{$\pm0.11$} & $37.59$\tiny{$\pm0.12$} & $27.46$\tiny{$\pm0.05$} & \hlcellb $35.27$\tiny{$\pm0.10$} \\

\midrule
     & \multicolumn{4}{c}{\textsc{\crux}} \\
     \cmidrule(lr){2-5}
     & {\direct} & {\pos} & {\iv} & {\hlcellb 1F} \\

        \midrule

        GPT-OSS-120B & $89.23$\tiny{$\pm0.08$} & $55.49$\tiny{$\pm0.14$} & $87.13$\tiny{$\pm0.07$} & \hlcellb $81.02$\tiny{$\pm0.06$} \\

        Nemotron-32B & $77.06$\tiny{$\pm0.09$} & $52.76$\tiny{$\pm0.09$} & $52.59$\tiny{$\pm0.13$} & \hlcellb $53.69$\tiny{$\pm0.08$} \\

        Nemotron-7B & $43.55$\tiny{$\pm0.20$} & $32.81$\tiny{$\pm0.13$} & $35.34$\tiny{$\pm0.09$} & \hlcellb $35.27$\tiny{$\pm0.09$} \\

        \bottomrule

\end{tabular}

\end{table}

\newpage
\subsection{Scaling: Varying the Numbers of In-Context Drafts with Mixed Correctness}

We also ablate the number of drafts in context and their correctness and observe the corresponding effects on contextual drag. Specifically, we repeat the same evaluation in \cref{sec:main-eval} with $1$-$4$ in-context drafts with mixed correctness (1T, 1F; 2T, 1T1F, 2F; 3T, 2T1F, 1T2F, 3F; and 4T, 3T1F, 2T2F, 1T3F, 4F). 

\begin{figure}[H]
    \centering
    \includegraphics[width=0.8\linewidth]{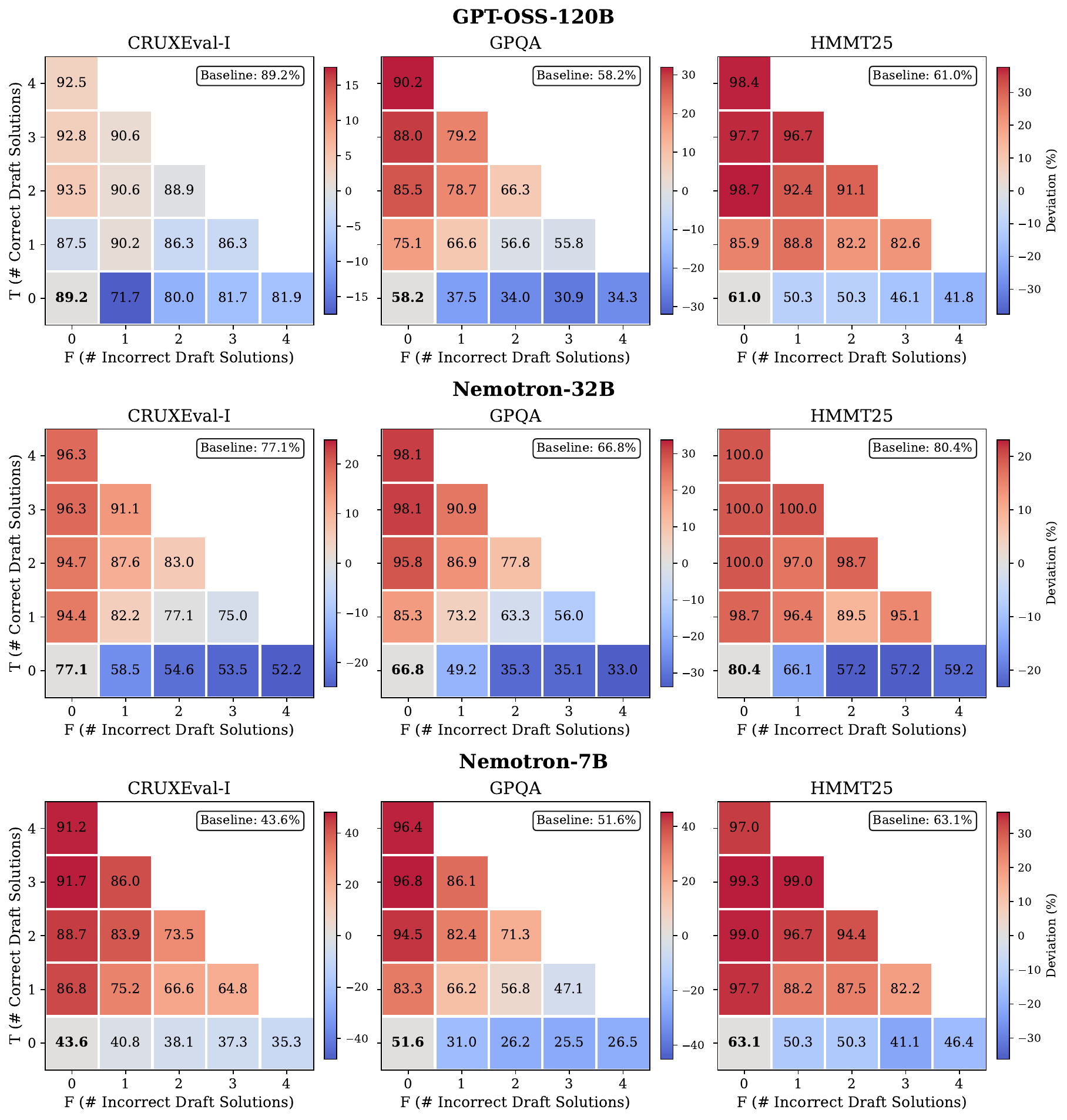}
    \caption{\textbf{\textbf{Scaling} ablation}: Performance changes relative to the {\direct} performance as the number of drafts with mixed correctness in context increases.}
    \label{fig:scaling-ablation}
\end{figure}

As shown in \cref{fig:scaling-ablation}, performance degrades sharply when the context contains only incorrect drafts, with more severe drops as the number of incorrect drafts increases. Correct drafts in context generally result in large performance gains, but models can still struggle when incorrect drafts dominate the context (\cref{fig:scaling-ablation}), notably for GPT-OSS-120B and Nemotron-32B on CRUXEval-I and GPQA.

\newpage
\section{Tree-Edit Distance (TED) Analysis}
\label{app:ted}

\begin{figure}[H]
    \centering
    \includegraphics[width=0.9\linewidth]{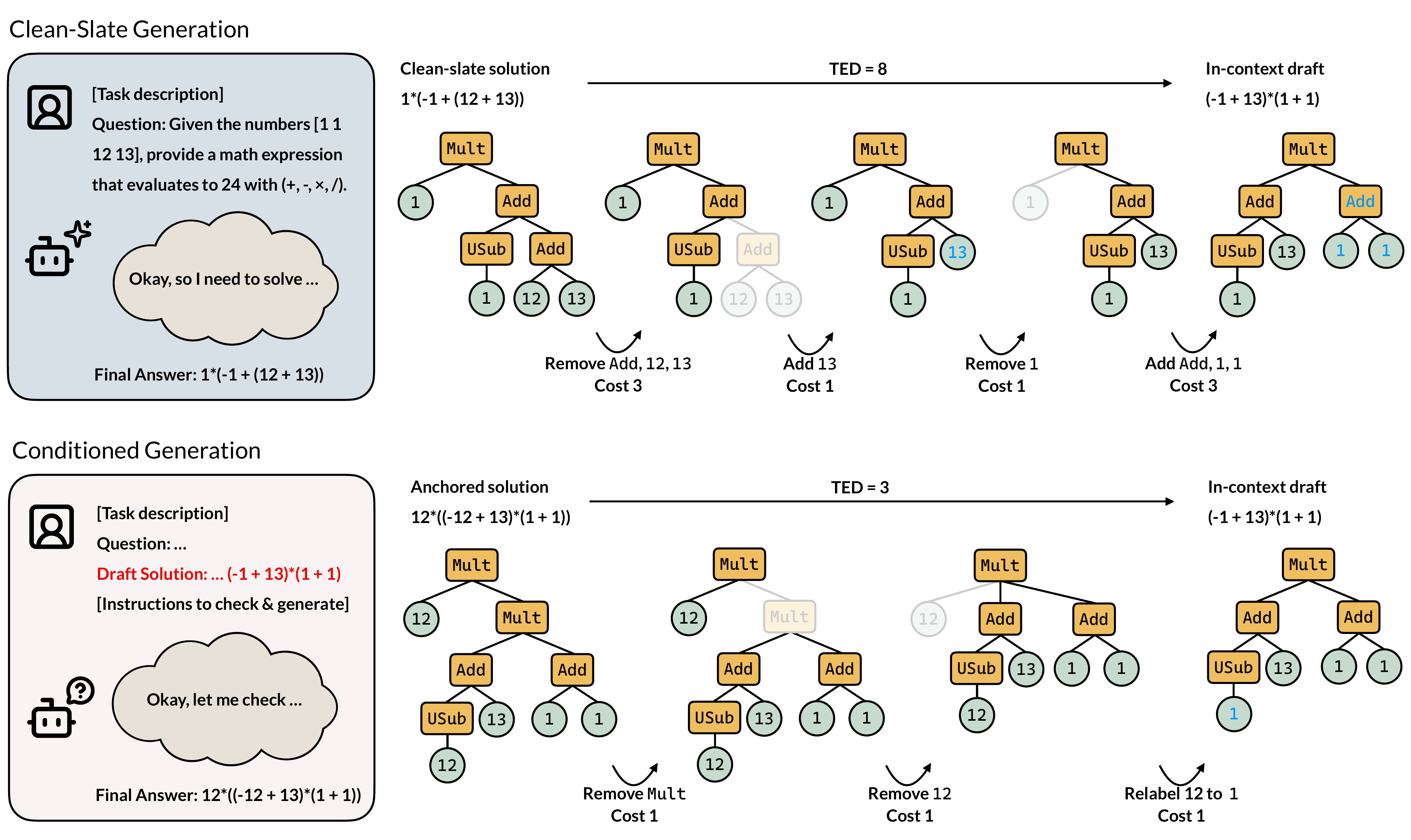}
    \caption{\textbf{Overview of TED analysis}: We measure the similarity in reasoning structure of the clean-slate generation, in-context draft, and the conditioned generation using TED. For each pair of clean-slate/conditioned generation and in-context draft, we extract the solution and parse it into a tree, then compute the TED metric between each pair.}
    \label{fig:ted}
\end{figure}

\subsection{Implementation}

We conduct tree-edit-distance (TED) analysis on the synthetic Game of 24 task, which is commonly referred to as a restricted variant of the \emph{Countdown} problem. Since each solution to the Game of 24 puzzle can be parsed into a tree, this allows us to quantify the structural similarity of the reasoning and error patterns underlying the solution. 

We take the set of clean-slate generations under {\direct} and the conditioned generations under {\onef} in \cref{sec:main-eval}. As shown in \cref{fig:ted}, for each problem, we form two pairs of responses: (1) a clean-slate generation and an in-context draft solution, and (2) a conditioned generation and an in-context draft solution. For each pair, we extract the solutions from the reasoning traces and parse them into trees. Then we compute the TED between each pair.

In the original formulation, the TED metric measures the dissimilarity between two trees by the minimum-cost sequence of edit operations needed to transform one tree into the other \cite{zhang1989simple}. The computation of TED involves three types of operations: (1) node insertion, (2) node deletion, and (3) node relabeling (or substitution). \cref{fig:ted} provides an example of TED computation used in our analysis.

\subsection{Additional Analysis of Contextual Drag under {\twof}}

\begin{figure}[H]
    \centering
    \includegraphics[width=0.5\linewidth]{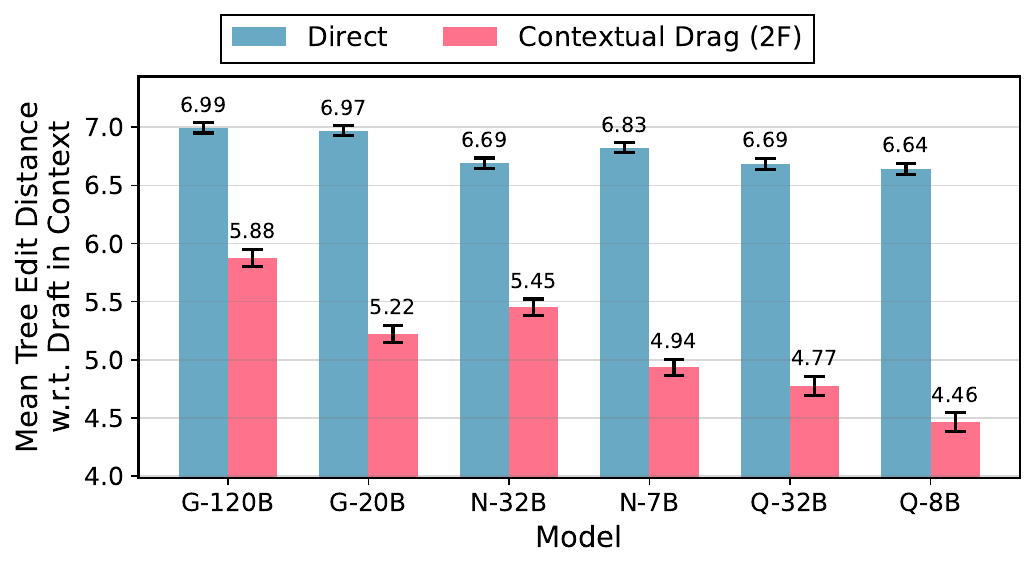}
    \caption{\textbf{TED analysis of conditioned generations under {\twof}}}
    \label{fig:ted-2f}
\end{figure}

We also repeat the same analysis from \cref{sec:ted-analysis} for conditioned generation under {\twof}. Since there are two in-context drafts, we define
\[
\text{TED}'(\text{generation}, \{\text{draft1},\text{draft2}\})=\min(\text{TED}(\text{generation},\text{draft1}),\text{TED}(\text{generation},\text{draft2}))
\]
Note that this formulation necessarily results in lower absolute TED values, but does not imply that contextual drag is more severe under {\twof} than under {\onef}. Since we define this for both clean-slate and conditioned generation, the comparison of their TED values remains fair. \cref{fig:ted-2f} reveals a similar trend to \cref{fig:edit-distance-base}, suggesting that increased diversity of drafts in context does not substantially alleviate contextual drag.

\vspace{-0.4in}

\subsection{Ablation of Answer Reuse Instruction}

Another potential confound for the reduced TED distance is the explicit instruction that prompts the model to reuse parts of the solution if it finds them correct. In this ablation setting, we show that removing such reuse instruction does not eliminate the edit distance gap. Below we \textcolor{blue}{highlight the removal of \st{the reuse instruction in the prompt}}.

\begin{promptbox}{template-1f-no-reuse}{{\onef} (no reuse instruction)}
    ~\\
    In this task, you will be provided with a problem and one draft solution to that problem. For the draft, walk through the solution step-by-step and check for any fatal mistakes in the reasoning (for example: incorrect use of theorems or facts, incorrect calculations, etc.). Please note that once you spot a fatal error in a draft's reasoning, you **MUST** stop checking the remaining steps of that draft and move on.\\

    For the draft, output an overall verdict on the correctness of the draft using the following tag format:\\

    <overall\_verdict>[Correct / Incorrect]</overall\_verdict>\\
    
    While evaluating individual steps, be strict about algebra, logic, and correct theorem application. If the draft is incorrect, write a complete, correct solution\textcolor{blue}{\st{ (you may reuse sound ideas from the draft)}}. If the draft is correct, you may adopt and polish it as the final solution.\\

    Conclude by presenting the final answer only inside \textbackslash{}boxed\{\} (i.e., the final answer should appear only inside a LaTeX `\textbackslash{}boxed\{\}`).\\
    \\
    The problem is as follows:\\
    -- beginning of problem --\\
    \textbf{\{problem\}}\\
    -- end of problem --\\

    Here is the draft solution you need to consider:\\

    \#\#\# Solution:\\
    -- beginning of the draft --\\
    \textbf{\{draft1\}}\\
    -- end of the draft --\\

    Please:\\
    1. Carefully evaluate the draft solution step-by-step. Once you spot a fatal error in a draft's reasoning, you **MUST** stop checking the remaining steps of that draft and move on. \\
    2. Output the correctness verdicts for the draft.\\
    3. Provide a correct solution to the problem with complete reasoning steps that lead to the answer.\\
    \\
    Remember: conclude with the final answer only in \textbackslash{}boxed\{\}.
\end{promptbox}

\begin{figure}[H]
    \centering
    \includegraphics[width=0.95\linewidth]{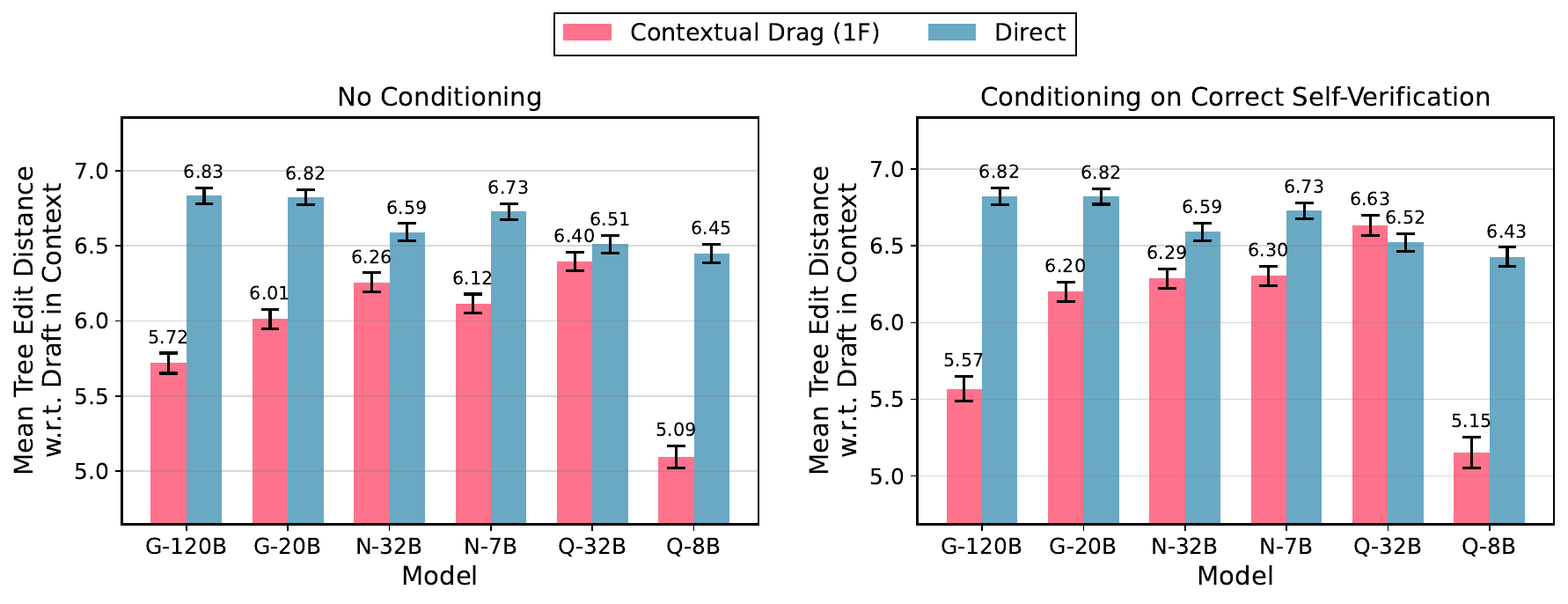}
    \caption{\textbf{TED analysis of conditioned generations under 1F without reuse instruction in the prompt}}
    \label{fig:ted-1f-no-reuse}
\end{figure}

As shown in \cref{fig:ted-1f-no-reuse}, removing the reuse instruction does not significantly change the trend, which remains consistent with \cref{fig:edit-distance-base} and \cref{fig:conditioned-self-detected-error}. This further supports our interpretation that the reduced TED distance is driven by contextual drag rather than instruction-following behavior.

\newpage
\section{Additional Analysis of Contextual Drag Despite Error Awareness}
\label{app:evals-conditioned-drop}

\paragraph{External Error Signal} Here we provide the modified prompt used in \cref{sec:conditioned-prompted} which explicitly and repeatedly stated the incorrectness of the draft in context.

\begin{promptbox}{template-ablation-frm}{External Error Signal ({\onef})}
    ~\\
    In this task, you will be provided with a problem and one draft solution \textcolor{blue}{with **INCORRECT** final answer} to that problem. For the \textcolor{blue}{**INCORRECT**} draft, walk through the solution step-by-step and check for any fatal mistakes in the reasoning (for example: incorrect use of theorems or facts, incorrect calculations, etc.). Please note that once you spot a fatal error in a draft's reasoning, you **MUST** stop checking the remaining steps of that draft and move on.\\

    For the \textcolor{blue}{**INCORRECT**} draft, output an overall verdict on the correctness of the draft using the following tag format:\\

    <overall\_verdict>\textcolor{blue}{Incorrect}</overall\_verdict>\\

    While evaluating individual steps, be strict about algebra, logic, and correct theorem application. Finally, write a complete, correct solution (you may reuse sound ideas from the draft\textcolor{blue}{, but **DO NOT COPY** the final answer because it is **INCORRECT**}).\\

    Conclude by presenting the final answer only inside \textbackslash{}boxed\{\} (i.e., the final answer should appear only inside a LaTeX `\textbackslash{}boxed\{\}`).\\

    The problem is as follows:\\
    -- beginning of problem --\\
    \textbf{\{problem\}}\\
    -- end of problem --\\

    Here is the draft solution \textcolor{blue}{with **INCORRECT** final answer} you need to consider:\\

    \#\#\# Solution:\\
    -- beginning of the draft --\\
    \textbf{\{draft1\}}\\
    -- end of the draft --\\

    Please:\\
    1. Carefully evaluate the draft solution step-by-step. Once you spot a fatal error in a draft's reasoning, you **MUST** stop checking the remaining steps of that draft and move on. \\
    2. Output the correctness verdicts for the draft.\\
    3. Provide a correct solution to the problem with complete reasoning steps that lead to the answer.\\

    Remember: \textcolor{blue}{the draft solution has **INCORRECT** final answer.} Please conclude with the final answer only in \textbackslash{}boxed\{\}.
\end{promptbox}

\begin{table}[H]
    \centering
    \caption{Full results for \cref{fig:conditioned-prompted-error}, where E-Error refers to the performance under external error signal.}
\begin{adjustbox}{width=\textwidth}
\begin{tabular}{l |  c c c c c c c c c c c c c c c c}
\toprule\toprule
\textbf{Model} & \multicolumn{2}{c}{\textsc{\aime{24}}} & \multicolumn{2}{c}{\textsc{\aime{25}}} & \multicolumn{2}{c}{\textsc{\hmmt{24}}} & \multicolumn{2}{c}{\textsc{\hmmt{25}}} \\
\cmidrule(lr){2-3} \cmidrule(lr){4-5} \cmidrule(lr){6-7} \cmidrule(lr){8-9}
& \textsc{Direct} & \textsc{E-Error} & \textsc{Direct} & \textsc{E-Error} & \textsc{Direct} & \textsc{E-Error} & \textsc{Direct} & \textsc{E-Error} \\
\midrule
GPT-OSS-120B & $66.2$ & $25.0$ (\perfdecrease{-$41.2$}) & $66.8$ & $34.1$ (\perfdecrease{-$32.7$}) & $48.8$ & $26.2$ (\perfdecrease{-$22.7$}) & $61.0$ & $40.0$ (\perfdecrease{-$21.0$}) \\
GPT-OSS-20B & $51.9$ & $17.5$ (\perfdecrease{-$34.4$}) & $51.9$ & $17.7$ (\perfdecrease{-$34.2$}) & $40.2$ & $13.3$ (\perfdecrease{-$27.0$}) & $58.6$ & $27.1$ (\perfdecrease{-$31.5$}) \\
Nemotron-32B & $83.8$ & $76.9$ (\perfdecrease{-$6.9$}) & $78.4$ & $71.4$ (\perfdecrease{-$7.0$}) & $61.7$ & $57.4$ (\perfdecrease{-$4.3$}) & $80.4$ & $77.1$ (\perfdecrease{-$3.3$}) \\
Nemotron-7B & $67.5$ & $60.6$ (\perfdecrease{-$6.9$}) & $67.8$ & $56.2$ (\perfdecrease{-$11.5$}) & $49.2$ & $43.8$ (\perfdecrease{-$5.5$}) & $63.1$ & $59.2$ (\perfdecrease{-$3.9$}) \\
Qwen3-32B & $65.0$ & $48.8$ (\perfdecrease{-$16.2$}) & $54.8$ & $46.9$ (\perfdecrease{-$7.9$}) & $39.1$ & $37.9$ (\perfdecrease{-$1.2$}) & $49.7$ & $42.9$ (\perfdecrease{-$6.8$}) \\
Qwen3-8B & $55.6$ & $41.9$ (\perfdecrease{-$13.8$}) & $45.7$ & $30.8$ (\perfdecrease{-$14.9$}) & $21.9$ & $27.3$ (\perfincrease{+$5.5$}) & $42.3$ & $33.9$ (\perfdecrease{-$8.3$}) \\
\midrule
\textbf{Model} & \multicolumn{2}{c}{\textsc{\gpqa}} & \multicolumn{2}{c}{\textsc{\mmlu}} & \multicolumn{2}{c}{\textsc{\crux}} & \multicolumn{2}{c}{\textsc{\game}} \\
\cmidrule(lr){2-3} \cmidrule(lr){4-5} \cmidrule(lr){6-7} \cmidrule(lr){8-9}
& \textsc{Direct} & \textsc{E-Error} & \textsc{Direct} & \textsc{E-Error} & \textsc{Direct} & \textsc{E-Error} & \textsc{Direct} & \textsc{E-Error} \\
\midrule
GPT-OSS-120B & $58.2$ & $44.1$ (\perfdecrease{-$14.1$}) & $67.6$ & $73.8$ (\perfincrease{+$6.2$}) & $89.2$ & $80.6$ (\perfdecrease{-$8.6$}) & $77.3$ & $62.6$ (\perfdecrease{-$14.7$}) \\
GPT-OSS-20B & $47.0$ & $32.1$ (\perfdecrease{-$14.8$}) & $58.9$ & $50.8$ (\perfdecrease{-$8.1$}) & $61.3$ & $48.7$ (\perfdecrease{-$12.6$}) & $78.3$ & $74.4$ (\perfdecrease{-$3.9$}) \\
Nemotron-32B & $66.8$ & $64.9$ (\perfdecrease{-$1.9$}) & $67.2$ & $80.3$ (\perfincrease{+$13.0$}) & $77.1$ & $68.6$ (\perfdecrease{-$8.5$}) & $79.5$ & $74.1$ (\perfdecrease{-$5.5$}) \\
Nemotron-7B & $51.6$ & $50.3$ (\perfdecrease{-$1.4$}) & $56.5$ & $64.2$ (\perfincrease{+$7.7$}) & $43.6$ & $39.4$ (\perfdecrease{-$4.2$}) & $75.4$ & $69.4$ (\perfdecrease{-$6.0$}) \\
Qwen3-32B & $57.6$ & $54.9$ (\perfdecrease{-$2.7$}) & $70.6$ & $80.5$ (\perfincrease{+$9.9$}) & $75.8$ & $67.2$ (\perfdecrease{-$8.6$}) & $78.5$ & $60.5$ (\perfdecrease{-$17.9$}) \\
Qwen3-8B & $47.8$ & $38.2$ (\perfdecrease{-$9.6$}) & $61.7$ & $66.1$ (\perfincrease{+$4.4$}) & $67.5$ & $51.6$ (\perfdecrease{-$15.9$}) & $76.3$ & $48.9$ (\perfdecrease{-$27.4$}) \\
\bottomrule
\end{tabular}
\end{adjustbox}
    \label{tab:appd-external-error}
\end{table}

\begin{table}[H]
    \centering
    \caption{Full results for \cref{fig:conditioned-self-detected-error}, where SD-Error refers to the performance under self-detected error signal. We also show the number of problems remaining after the post-hoc filtering next to the direct performance.}
\begin{adjustbox}{width=\textwidth}
\begin{tabular}{l |  c c c c c c c c c c c c c c c c}
\toprule\toprule
\textbf{Model} & \multicolumn{2}{c}{\textsc{\aime{24}}} & \multicolumn{2}{c}{\textsc{\aime{25}}} & \multicolumn{2}{c}{\textsc{\hmmt{24}}} & \multicolumn{2}{c}{\textsc{\hmmt{25}}} \\
\cmidrule(lr){2-3} \cmidrule(lr){4-5} \cmidrule(lr){6-7} \cmidrule(lr){8-9}
& \textsc{Direct} & \textsc{SD-Error} & \textsc{Direct} & \textsc{SD-Error} & \textsc{Direct} & \textsc{SD-Error} & \textsc{Direct} & \textsc{SD-Error} \\
\midrule
GPT-OSS-120B & $66.2$ ($10$/$10$) & $53.6$ (\perfdecrease{-$12.7$}) & $66.8$ ($13$/$13$) & $70.9$ (\perfincrease{+$4.0$}) & $48.8$ ($16$/$16$) & $52.9$ (\perfincrease{+$4.1$}) & $60.5$ ($19$/$21$) & $64.3$ (\perfincrease{+$3.7$}) \\
GPT-OSS-20B & $51.9$ ($10$/$10$) & $28.5$ (\perfdecrease{-$23.4$}) & $51.9$ ($13$/$13$) & $43.3$ (\perfdecrease{-$8.6$}) & $40.2$ ($16$/$16$) & $16.4$ (\perfdecrease{-$23.8$}) & $57.2$ ($20$/$21$) & $33.9$ (\perfdecrease{-$23.3$}) \\
Nemotron-32B & $83.8$ ($10$/$10$) & $93.3$ (\perfincrease{+$9.5$}) & $78.4$ ($13$/$13$) & $80.6$ (\perfincrease{+$2.3$}) & $61.7$ ($16$/$16$) & $66.2$ (\perfincrease{+$4.5$}) & $80.4$ ($21$/$21$) & $85.7$ (\perfincrease{+$5.3$}) \\
Nemotron-7B & $67.5$ ($10$/$10$) & $76.0$ (\perfincrease{+$8.5$}) & $67.8$ ($13$/$13$) & $83.1$ (\perfincrease{+$15.3$}) & $50.8$ ($15$/$16$) & $54.4$ (\perfincrease{+$3.6$}) & $63.1$ ($21$/$21$) & $72.1$ (\perfincrease{+$9.0$}) \\
Qwen3-32B & $65.0$ ($10$/$10$) & $49.0$ (\perfdecrease{-$16.0$}) & $55.7$ ($12$/$13$) & $53.7$ (\perfdecrease{-$2.1$}) & $39.1$ ($16$/$16$) & $47.3$ (\perfincrease{+$8.2$}) & $50.3$ ($20$/$21$) & $51.4$ (\perfincrease{+$1.1$}) \\
Qwen3-8B & $61.8$ ($9$/$10$) & $54.1$ (\perfdecrease{-$7.7$}) & $50.0$ ($8$/$13$) & $44.3$ (\perfdecrease{-$5.7$}) & $20.1$ ($14$/$16$) & $36.2$ (\perfincrease{+$16.1$}) & $46.5$ ($18$/$21$) & $46.3$ (\perfdecrease{-$0.2$}) \\
\midrule
\textbf{Model} & \multicolumn{2}{c}{\textsc{\gpqa}} & \multicolumn{2}{c}{\textsc{\mmlu}} & \multicolumn{2}{c}{\textsc{\crux}} & \multicolumn{2}{c}{\textsc{\game}} \\
\cmidrule(lr){2-3} \cmidrule(lr){4-5} \cmidrule(lr){6-7} \cmidrule(lr){8-9}
& \textsc{Direct} & \textsc{SD-Error} & \textsc{Direct} & \textsc{SD-Error} & \textsc{Direct} & \textsc{SD-Error} & \textsc{Direct} & \textsc{SD-Error} \\
\midrule
GPT-OSS-120B & $57.9$ ($112$/$132$) & $48.9$ (\perfdecrease{-$9.1$}) & $73.0$ ($450$/$569$) & $79.7$ (\perfincrease{+$6.7$}) & $90.6$ ($173$/$188$) & $85.7$ (\perfdecrease{-$4.9$}) & $77.5$ ($634$/$653$) & $80.8$ (\perfincrease{+$3.4$}) \\
GPT-OSS-20B & $44.8$ ($115$/$132$) & $25.0$ (\perfdecrease{-$19.8$}) & $61.4$ ($458$/$569$) & $61.3$ (\perfdecrease{-$0.1$}) & $61.5$ ($175$/$188$) & $53.5$ (\perfdecrease{-$8.0$}) & $78.5$ ($603$/$653$) & $79.0$ (\perfincrease{+$0.5$}) \\
Nemotron-32B & $66.5$ ($104$/$132$) & $66.0$ (\perfdecrease{-$0.5$}) & $77.1$ ($293$/$569$) & $74.2$ (\perfdecrease{-$2.8$}) & $79.1$ ($164$/$188$) & $74.7$ (\perfdecrease{-$4.4$}) & $79.4$ ($649$/$653$) & $76.0$ (\perfdecrease{-$3.4$}) \\
Nemotron-7B & $50.8$ ($103$/$132$) & $56.5$ (\perfincrease{+$5.7$}) & $62.7$ ($266$/$569$) & $59.9$ (\perfdecrease{-$2.8$}) & $44.4$ ($163$/$188$) & $52.4$ (\perfincrease{+$8.0$}) & $75.6$ ($641$/$653$) & $76.8$ (\perfincrease{+$1.3$}) \\
Qwen3-32B & $55.3$ ($110$/$132$) & $43.3$ (\perfdecrease{-$12.1$}) & $74.6$ ($426$/$569$) & $77.4$ (\perfincrease{+$2.9$}) & $75.5$ ($161$/$188$) & $72.5$ (\perfdecrease{-$3.0$}) & $78.3$ ($645$/$653$) & $70.3$ (\perfdecrease{-$8.0$}) \\
Qwen3-8B & $44.5$ ($104$/$132$) & $32.3$ (\perfdecrease{-$12.2$}) & $70.4$ ($332$/$569$) & $67.9$ (\perfdecrease{-$2.4$}) & $67.3$ ($158$/$188$) & $60.1$ (\perfdecrease{-$7.2$}) & $76.2$ ($576$/$653$) & $74.7$ (\perfdecrease{-$1.5$}) \\
\bottomrule
\end{tabular}
\end{adjustbox}
    \label{tab:appd-self-detected-error}
\end{table}

\newpage
\section{Additional Details of Context Manipulation Experiments}
\label{app:context-denoising}

\subsection{Implementation of {\revise} and {\filter}}
\label{app:context-denoising-implementation}
We provide the full implementation details of both methods below. We note that both involve multiple inference steps and when using the output from a previous step, we always remove the thinking traces wrapped in \texttt{<think></think>} for manageable context length. We only present the prompt templates for {\onef} for illustration.

{\revise} consists of $2$ steps of inference:
\begin{enumerate}
    \item (revise): given the \texttt{\textbf{problem}} and one \texttt{\textbf{draft}}, the model is asked to address any mistakes and produce a revised solution. Denote the output as the \texttt{\textbf{revised\_draft}}. This inference needs to be repeated for each draft independently;\\
    \begin{promptbox}{template-revise-revise}{{\revise} (revise step, {\onef})}
        ~\\
        You are given a problem and one draft solution to that problem. Your task is to revise the draft to produce an improved, complete solution that addresses any mistakes or shortcomings in the original draft. Ensure that your revised solution includes all necessary reasoning steps leading to the final answer and conclude by presenting the final answer only inside \textbackslash{}boxed\{\}.\\
    
        The problem is as follows:\\
        -- beginning of problem --\\
        \textbf{\{problem\}}\\
        -- end of problem --\\
        
        Here is the draft solution you need to revise:\\
        -- beginning of the draft --\\
        \textbf{\{draft1\}}\\
        -- end of the draft --\\
        
        Remember: conclude with the final answer only in \textbackslash{}boxed\{\}.
    \end{promptbox}
    \item (solve): given the \texttt{\textbf{problem}} and the \texttt{\textbf{revised\_draft}}, the model is asked to generate a new solution. The prompt template is the same as the \hyperref[template-ablation-iv]{{\iv} ablation template} except that the draft in the context is replaced with its revised version. We note that the instruction to check the draft correctness is removed to optimize model performance.\\
    \begin{promptbox}{template-revise-solve}{{\revise} (solve step, {\onef})}
        ~\\
        In this task, you will be provided with a problem and one draft solution to that problem. If the draft is incorrect, write a complete, correct solution (you may reuse sound ideas from the draft). If the draft is correct, you may adopt and polish it as the final solution.\\
    
        Conclude by presenting the final answer only inside \textbackslash{}boxed\{\} (i.e., the final answer should appear only inside a LaTeX `\textbackslash{}boxed\{\}`).\\
        
        The problem is as follows:\\
        -- beginning of problem --\\
        \textbf{\{problem\}}\\
        -- end of problem --\\
        
        Here is the draft solution you need to consider:\\
        
        \#\#\# Solution:\\
        -- beginning of the draft --\\
        \textcolor{blue}{\textbf{\{revised\_draft1\}}}\\
        -- end of the draft --\\
        
        Please provide a correct solution to the problem with complete reasoning steps that lead to the answer.\\
        
        Remember: conclude with the final answer only in \textbackslash{}boxed\{\}.
    \end{promptbox}
\end{enumerate}

{\filter} consists of $3$ steps of inference:
\begin{enumerate}
    \item (strategy): given the \texttt{\textbf{problem}}, the model is asked to produce a high-level strategy to solve the problem. The model is not allowed to include any intermediate steps or the final answer. Denote the output as the \texttt{\textbf{strategy}};\\
    \begin{promptbox}{template-filter-strategy}{{\filter} (strategy stage, single draft)}
        ~\\
        You are given a problem. Your task is **NOT to solve the problem**. Instead, produce a **high-level strategy or a plan** that you would use to approach the problem.\\

        Requirements for your response:\\
        1. **Do NOT give any step-by-step computation or any final answers**.\\
        2. Produce a concise, structured plan (use numbered steps or short bullet points) of you would take as if you would attempt to solve the given question. You can include any useful heuristics, ways to reduce problem complexity, and common pitfalls to avoid.\\
        3. Finally, include a short checklist to verify a candidate answer.\\

        Now, produce the high-level strategy (no solution) for the following problem:\\
        -- beginning of problem --\\
        \textbf{\{problem\}}\\
        -- end of problem --\\

        Remember that you **do not attempt to solve the question or include any intermediate steps or the final answer in your response**.
    \end{promptbox}
    \item (filter): given the \texttt{\textbf{problem}}, the \texttt{\textbf{strategy}}, and one \texttt{\textbf{draft}}, the model is asked to filter the draft to only keep the correct and useful steps toward solving the problem according to the \texttt{\textbf{strategy}} it produced in the previous step. Similar to the revise step in {\revise}, this step needs to be repeated for each draft independently. Denote the output as the \texttt{\textbf{filtered\_draft}};\\
    \begin{promptbox}{template-filter-filter}{{\filter} (filter step, {\onef})}
        ~\\
        You are given a problem, a draft solution attempt, and a high-level strategy for solving the problem. Your task is to identify the useful components or intermediate steps in the draft solution that align with the given strategy. Focus on selecting steps that contribute meaningfully to progressing toward the final solution, even if the draft solution as a whole may contain mistakes or irrelevant parts. Format your response as numbered steps, followed by a concise (no more than 2 sentences), clear explanation of how each step fits into the strategy. Remember: **only include the useful and correct parts** and **do not include any incorrect or irrelevant parts** in your response.\\

        The problem is as following:\\
        -- beginning of problem --\\
        \textbf{\{problem\}}\\
        -- end of problem --\\

        You should consider the following strategy to solve the problem:\\
        -- beginning of strategy --\\
        \textbf{\{strategy\}}\\
        -- end of strategy --\\

        Please **only output the useful and correct components or intermediate steps that align with the above strategy** from the following draft solution:\\
        -- beginning of draft solution --\\
        \textbf{\{draft1\}}\\
        -- end of draft solution --\\

        Remember to format your response as numbered steps, followed by a concise explanation of why they are useful.
    \end{promptbox}
    \item (solve): given the \texttt{\textbf{problem}} and the \texttt{\textbf{filtered\_draft}}, the model is asked to generate a new solution. Similar to {\revise}, the prompt template used at this step is the same as the \hyperref[template-1f]{{\onef} template} in \cref{app:evals_prompt_templates} except that (1) the draft in the context is replaced with its revised version and (2) the instruction to check the correctness of the draft is removed to optimize model performance.\\
    \begin{promptbox}{template-filter-solve}{{\filter} (solve step, {\onef})}
        ~\\
        In this task, you will be provided with a problem and one draft solution to that problem. If the draft is incorrect, write a complete, correct solution (you may reuse sound ideas from the draft). If the draft is correct, you may adopt and polish it as the final solution.\\
    
        Conclude by presenting the final answer only inside \textbackslash{}boxed\{\} (i.e., the final answer should appear only inside a LaTeX `\textbackslash{}boxed\{\}`).\\
        
        The problem is as follows:\\
        -- beginning of problem --\\
        \textbf{\{problem\}}\\
        -- end of problem --\\
        
        Here is the draft solution you need to consider:\\
        
        \#\#\# Solution:\\
        -- beginning of the draft --\\
        \textcolor{blue}{\textbf{\{filtered\_draft1\}}}\\
        -- end of the draft --\\
        
        Please provide a correct solution to the problem with complete reasoning steps that lead to the answer.\\
        
        Remember: conclude with the final answer only in \textbackslash{}boxed\{\}.
    \end{promptbox}
\end{enumerate}

\newpage

\subsection{Additional Results}
\label{app:context-denoising-full-results}

Here we provide the full results of the two context denoising approaches discussed in \cref{sec:context_denoising}. Results for \onef\ are shown in \cref{tab:context_manipulation_full_1_draft} and results for \twof\ are shown in \cref{tab:context_manipulation_full_2_drafts}.

\begin{table}[H]
    \centering
    \caption{Additional results of context denoising experiments (1F) with 95\% confidence internal. }
    \begin{adjustbox}{width=\textwidth}
    \begin{tabular}{l | c c c c c c c c c c}
    \toprule\toprule
     & \multicolumn{5}{c}{\aime{24}} & \multicolumn{5}{c}{\aime{25}} \\
     \cmidrule(lr){2-6} \cmidrule(lr){7-11}
     & {\direct} & {\revise} & {\filter} & {\hlcellb {\onef}} & {\hlcellb {\onef}-{\iv}} & {\direct} & {\revise} & {\filter} & {\hlcellb {\onef}} & {\hlcellb {\onef}-{\iv}} \\
    \midrule
        GPT-OSS-120B & $66.25$\tiny{$\pm1.72$} & $80.00$\tiny{$\pm0.55$} & $63.75$\tiny{$\pm2.10$} & \hlcellb $43.75$\tiny{$\pm0.73$} & \hlcellb $57.50$\tiny{$\pm0.95$} & $66.83$\tiny{$\pm1.47$} & $68.27$\tiny{$\pm1.03$} & $65.38$\tiny{$\pm1.52$} & \hlcellb $55.29$\tiny{$\pm1.47$} & \hlcellb $58.17$\tiny{$\pm1.50$} \\
        GPT-OSS-20B & $51.88$\tiny{$\pm1.47$} & $52.50$\tiny{$\pm0.77$} & $53.12$\tiny{$\pm1.21$} & \hlcellb $17.50$\tiny{$\pm2.26$} & \hlcellb $41.25$\tiny{$\pm1.80$} & $51.92$\tiny{$\pm1.54$} & $48.56$\tiny{$\pm1.43$} & $47.60$\tiny{$\pm1.22$} & \hlcellb $18.75$\tiny{$\pm1.20$} & \hlcellb $46.63$\tiny{$\pm2.54$} \\
        Nemotron-32B & $83.75$\tiny{$\pm1.80$} & $74.37$\tiny{$\pm1.02$} & $84.38$\tiny{$\pm1.02$} & \hlcellb $67.50$\tiny{$\pm1.82$} & \hlcellb $66.88$\tiny{$\pm1.33$} & $78.37$\tiny{$\pm1.18$} & $73.56$\tiny{$\pm1.07$} & $84.13$\tiny{$\pm0.58$} & \hlcellb $61.54$\tiny{$\pm1.28$} & \hlcellb $62.50$\tiny{$\pm1.21$} \\
        Nemotron-7B & $67.50$\tiny{$\pm1.28$} & $63.13$\tiny{$\pm1.33$} & $68.12$\tiny{$\pm0.94$} & \hlcellb $53.12$\tiny{$\pm1.73$} & \hlcellb $56.25$\tiny{$\pm2.24$} & $67.79$\tiny{$\pm1.12$} & $63.94$\tiny{$\pm0.73$} & $70.19$\tiny{$\pm0.72$} & \hlcellb $52.40$\tiny{$\pm0.97$} & \hlcellb $46.63$\tiny{$\pm0.78$} \\
        Qwen3-32B & $65.00$\tiny{$\pm2.05$} & $46.25$\tiny{$\pm1.31$} & $65.62$\tiny{$\pm1.28$} & \hlcellb $41.25$\tiny{$\pm1.20$} & \hlcellb $43.75$\tiny{$\pm3.04$} & $54.81$\tiny{$\pm1.47$} & $37.98$\tiny{$\pm0.78$} & $53.85$\tiny{$\pm0.98$} & \hlcellb $31.25$\tiny{$\pm0.94$} & \hlcellb $34.13$\tiny{$\pm1.54$} \\
        Qwen3-8B & $55.62$\tiny{$\pm2.12$} & $50.00$\tiny{$\pm1.23$} & $66.25$\tiny{$\pm1.20$} & \hlcellb $30.63$\tiny{$\pm1.28$} & \hlcellb $31.87$\tiny{$\pm1.21$} & $45.67$\tiny{$\pm1.20$} & $39.42$\tiny{$\pm0.96$} & $52.88$\tiny{$\pm1.03$} & \hlcellb $20.19$\tiny{$\pm1.09$} & \hlcellb $19.23$\tiny{$\pm1.23$} \\
    \midrule
     & \multicolumn{5}{c}{\hmmt{24}} & \multicolumn{5}{c}{\hmmt{25}} \\
     \cmidrule(lr){2-6} \cmidrule(lr){7-11}
     & {\direct} & {\revise} & {\filter} & {\hlcellb {\onef}} & {\hlcellb {\onef}-{\iv}} & {\direct} & {\revise} & {\filter} & {\hlcellb {\onef}} & {\hlcellb {\onef}-{\iv}} \\
    \midrule
        GPT-OSS-120B & $48.83$\tiny{$\pm1.22$} & $57.03$\tiny{$\pm0.59$} & $44.92$\tiny{$\pm1.18$} & \hlcellb $39.84$\tiny{$\pm0.84$} & \hlcellb $43.36$\tiny{$\pm1.56$} & $61.01$\tiny{$\pm0.70$} & $68.75$\tiny{$\pm0.52$} & $59.23$\tiny{$\pm0.94$} & \hlcellb $48.51$\tiny{$\pm0.78$} & \hlcellb $52.98$\tiny{$\pm0.67$} \\
        GPT-OSS-20B & $40.23$\tiny{$\pm1.08$} & $50.00$\tiny{$\pm0.47$} & $30.86$\tiny{$\pm0.42$} & \hlcellb $13.28$\tiny{$\pm1.17$} & \hlcellb $30.08$\tiny{$\pm0.98$} & $58.63$\tiny{$\pm0.78$} & $62.20$\tiny{$\pm0.68$} & $45.83$\tiny{$\pm0.50$} & \hlcellb $20.54$\tiny{$\pm0.97$} & \hlcellb $42.26$\tiny{$\pm0.53$} \\
        Nemotron-32B & $61.72$\tiny{$\pm1.40$} & $62.11$\tiny{$\pm0.99$} & $66.02$\tiny{$\pm0.83$} & \hlcellb $51.95$\tiny{$\pm0.76$} & \hlcellb $50.78$\tiny{$\pm1.00$} & $80.36$\tiny{$\pm0.62$} & $74.40$\tiny{$\pm0.56$} & $91.96$\tiny{$\pm0.47$} & \hlcellb $65.48$\tiny{$\pm0.78$} & \hlcellb $67.56$\tiny{$\pm0.36$} \\
        Nemotron-7B & $49.22$\tiny{$\pm1.15$} & $49.61$\tiny{$\pm0.92$} & $43.75$\tiny{$\pm0.76$} & \hlcellb $35.94$\tiny{$\pm0.94$} & \hlcellb $35.55$\tiny{$\pm1.32$} & $63.10$\tiny{$\pm0.95$} & $63.10$\tiny{$\pm0.47$} & $62.20$\tiny{$\pm0.73$} & \hlcellb $54.17$\tiny{$\pm0.43$} & \hlcellb $51.19$\tiny{$\pm1.30$} \\
        Qwen3-32B & $39.06$\tiny{$\pm1.34$} & $41.02$\tiny{$\pm0.57$} & $34.77$\tiny{$\pm0.78$} & \hlcellb $35.55$\tiny{$\pm0.60$} & \hlcellb $29.69$\tiny{$\pm0.54$} & $49.70$\tiny{$\pm0.83$} & $48.81$\tiny{$\pm0.44$} & $43.15$\tiny{$\pm0.80$} & \hlcellb $36.90$\tiny{$\pm0.47$} & \hlcellb $40.77$\tiny{$\pm0.77$} \\
        Qwen3-8B & $21.88$\tiny{$\pm0.72$} & $30.47$\tiny{$\pm0.52$} & $26.56$\tiny{$\pm0.54$} & \hlcellb $22.27$\tiny{$\pm1.13$} & \hlcellb $23.05$\tiny{$\pm0.85$} & $42.26$\tiny{$\pm0.67$} & $38.69$\tiny{$\pm0.62$} & $42.86$\tiny{$\pm0.51$} & \hlcellb $34.23$\tiny{$\pm0.65$} & \hlcellb $36.61$\tiny{$\pm0.72$} \\
    \midrule
     & \multicolumn{5}{c}{\textsc{\gpqa}} & \multicolumn{5}{c}{\textsc{\mmlu}} \\
     \cmidrule(lr){2-6} \cmidrule(lr){7-11}
     & {\direct} & {\revise} & {\filter} & {\hlcellb {\onef}} & {\hlcellb {\onef}-{\iv}} & {\direct} & {\revise} & {\filter} & {\hlcellb {\onef}} & {\hlcellb {\onef}-{\iv}} \\
    \midrule
        GPT-OSS-120B & $58.24$\tiny{$\pm0.13$} & $46.88$\tiny{$\pm0.09$} & $54.88$\tiny{$\pm0.14$} & \hlcellb $35.70$\tiny{$\pm0.11$} & \hlcellb $38.16$\tiny{$\pm0.16$} & $67.59$\tiny{$\pm0.02$} & $48.11$\tiny{$\pm0.02$} & $63.70$\tiny{$\pm0.03$} & \hlcellb $42.26$\tiny{$\pm0.03$} & \hlcellb $40.49$\tiny{$\pm0.03$} \\
        GPT-OSS-20B & $46.97$\tiny{$\pm0.14$} & $33.33$\tiny{$\pm0.07$} & $43.70$\tiny{$\pm0.16$} & \hlcellb $16.86$\tiny{$\pm0.07$} & \hlcellb $26.66$\tiny{$\pm0.12$} & $58.86$\tiny{$\pm0.04$} & $35.54$\tiny{$\pm0.03$} & $55.26$\tiny{$\pm0.03$} & \hlcellb $22.73$\tiny{$\pm0.01$} & \hlcellb $25.48$\tiny{$\pm0.04$} \\
        Nemotron-32B & $66.81$\tiny{$\pm0.08$} & $48.15$\tiny{$\pm0.06$} & $68.47$\tiny{$\pm0.07$} & \hlcellb $49.43$\tiny{$\pm0.14$} & \hlcellb $39.44$\tiny{$\pm0.08$} & $67.25$\tiny{$\pm0.02$} & $31.38$\tiny{$\pm0.01$} & $64.54$\tiny{$\pm0.01$} & \hlcellb $32.74$\tiny{$\pm0.01$} & \hlcellb $24.75$\tiny{$\pm0.02$} \\
        Nemotron-7B & $51.61$\tiny{$\pm0.11$} & $35.27$\tiny{$\pm0.07$} & $55.73$\tiny{$\pm0.11$} & \hlcellb $35.27$\tiny{$\pm0.10$} & \hlcellb $27.46$\tiny{$\pm0.05$} & $56.46$\tiny{$\pm0.02$} & $19.63$\tiny{$\pm0.01$} & $51.58$\tiny{$\pm0.02$} & \hlcellb $23.58$\tiny{$\pm0.03$} & \hlcellb $15.99$\tiny{$\pm0.02$} \\
        Qwen3-32B & $57.62$\tiny{$\pm0.10$} & $27.04$\tiny{$\pm0.05$} & $58.38$\tiny{$\pm0.06$} & \hlcellb $34.19$\tiny{$\pm0.07$} & \hlcellb $22.68$\tiny{$\pm0.08$} & $70.62$\tiny{$\pm0.02$} & $30.23$\tiny{$\pm0.01$} & $39.08$\tiny{$\pm0.02$} & \hlcellb $40.53$\tiny{$\pm0.03$} & \hlcellb $23.91$\tiny{$\pm0.01$} \\
        Qwen3-8B & $47.77$\tiny{$\pm0.09$} & $25.24$\tiny{$\pm0.05$} & $45.27$\tiny{$\pm0.06$} & \hlcellb $27.13$\tiny{$\pm0.11$} & \hlcellb $18.75$\tiny{$\pm0.06$} & $61.65$\tiny{$\pm0.03$} & $28.77$\tiny{$\pm0.01$} & $57.96$\tiny{$\pm0.02$} & \hlcellb $27.76$\tiny{$\pm0.02$} & \hlcellb $20.05$\tiny{$\pm0.03$} \\
    \midrule
     & \multicolumn{5}{c}{\textsc{\crux}} & \multicolumn{5}{c}{\textsc{\game}} \\
     \cmidrule(lr){2-6} \cmidrule(lr){7-11}
     & {\direct} & {\revise} & {\filter} & {\hlcellb {\onef}} & {\hlcellb {\onef}-{\iv}} & {\direct} & {\revise} & {\filter} & {\hlcellb {\onef}} & {\hlcellb {\onef}-{\iv}} \\
    \midrule
        GPT-OSS-120B & $89.23$\tiny{$\pm0.08$} & $81.98$\tiny{$\pm0.05$} & $87.80$\tiny{$\pm0.06$} & \hlcellb $81.02$\tiny{$\pm0.06$} & \hlcellb $87.13$\tiny{$\pm0.07$} & $77.34$\tiny{$\pm0.03$} & $73.19$\tiny{$\pm0.02$} & $78.91$\tiny{$\pm0.01$} & \hlcellb $56.94$\tiny{$\pm0.05$} & \hlcellb $57.20$\tiny{$\pm0.06$} \\
        GPT-OSS-20B & $61.34$\tiny{$\pm0.28$} & $61.24$\tiny{$\pm0.10$} & $49.73$\tiny{$\pm0.14$} & \hlcellb $42.39$\tiny{$\pm0.14$} & \hlcellb $45.11$\tiny{$\pm0.12$} & $78.30$\tiny{$\pm0.03$} & $56.93$\tiny{$\pm0.05$} & $75.20$\tiny{$\pm0.08$} & \hlcellb $41.61$\tiny{$\pm0.07$} & \hlcellb $39.76$\tiny{$\pm0.07$} \\
        Nemotron-32B & $77.06$\tiny{$\pm0.09$} & $67.49$\tiny{$\pm0.06$} & $73.80$\tiny{$\pm0.06$} & \hlcellb $53.69$\tiny{$\pm0.08$} & \hlcellb $52.59$\tiny{$\pm0.13$} & $79.54$\tiny{$\pm0.03$} & $74.62$\tiny{$\pm0.01$} & $82.30$\tiny{$\pm0.01$} & \hlcellb $56.68$\tiny{$\pm0.04$} & \hlcellb $51.00$\tiny{$\pm0.02$} \\
        Nemotron-7B & $43.55$\tiny{$\pm0.20$} & $48.34$\tiny{$\pm0.06$} & $39.13$\tiny{$\pm0.08$} & \hlcellb $35.27$\tiny{$\pm0.09$} & \hlcellb $35.34$\tiny{$\pm0.09$} & $75.35$\tiny{$\pm0.02$} & $71.18$\tiny{$\pm0.02$} & $73.96$\tiny{$\pm0.02$} & \hlcellb $45.97$\tiny{$\pm0.04$} & \hlcellb $38.97$\tiny{$\pm0.03$} \\
        Qwen3-32B & $75.83$\tiny{$\pm0.08$} & $65.36$\tiny{$\pm0.03$} & $62.90$\tiny{$\pm0.06$} & \hlcellb $54.89$\tiny{$\pm0.06$} & \hlcellb $53.96$\tiny{$\pm0.04$} & $78.48$\tiny{$\pm0.03$} & $58.26$\tiny{$\pm0.02$} & $79.39$\tiny{$\pm0.03$} & \hlcellb $43.40$\tiny{$\pm0.05$} & \hlcellb $41.62$\tiny{$\pm0.03$} \\
        Qwen3-8B & $67.49$\tiny{$\pm0.07$} & $49.00$\tiny{$\pm0.06$} & $51.80$\tiny{$\pm0.09$} & \hlcellb $40.09$\tiny{$\pm0.08$} & \hlcellb $38.33$\tiny{$\pm0.04$} & $76.29$\tiny{$\pm0.03$} & $42.71$\tiny{$\pm0.02$} & $76.67$\tiny{$\pm0.02$} & \hlcellb $32.92$\tiny{$\pm0.03$} & \hlcellb $29.34$\tiny{$\pm0.02$} \\
    \bottomrule\bottomrule
\end{tabular}

    \end{adjustbox}

    \label{tab:context_manipulation_full_1_draft}
\end{table}
\newpage
\begin{table}[H]
    \centering
    \caption{Additional results of context denoising experiments (2F) with 95\% confidence internal.}
    \begin{adjustbox}{width=\textwidth}
    \begin{tabular}{l | c c c c c c c c c c}
    \toprule\toprule
     & \multicolumn{5}{c}{\aime{24}} & \multicolumn{5}{c}{\aime{25}} \\
     \cmidrule(lr){2-6} \cmidrule(lr){7-11}
     & {\direct} & {\revise} & {\filter} & {\hlcella {\twof}} & {\hlcella {\twof}-{\iv}} & {\direct} & {\revise} & {\filter} & {\hlcella {\twof}} & {\hlcella {\twof}-{\iv}} \\
    \midrule
        GPT-OSS-120B & $66.25$\tiny{$\pm1.72$} & $61.87$\tiny{$\pm1.25$} & $57.50$\tiny{$\pm1.60$} & \hlcella $43.75$\tiny{$\pm2.04$} & \hlcella $55.00$\tiny{$\pm2.19$} & $66.83$\tiny{$\pm1.47$} & $68.27$\tiny{$\pm0.73$} & $54.33$\tiny{$\pm1.45$} & \hlcella $40.38$\tiny{$\pm1.30$} & \hlcella $56.73$\tiny{$\pm1.52$} \\
        GPT-OSS-20B & $51.88$\tiny{$\pm1.47$} & $47.50$\tiny{$\pm1.16$} & $49.38$\tiny{$\pm2.01$} & \hlcella $21.25$\tiny{$\pm2.32$} & \hlcella $40.62$\tiny{$\pm2.01$} & $51.92$\tiny{$\pm1.54$} & $54.33$\tiny{$\pm1.01$} & $50.48$\tiny{$\pm1.22$} & \hlcella $20.67$\tiny{$\pm1.51$} & \hlcella $46.15$\tiny{$\pm1.11$} \\
        Nemotron-32B & $83.75$\tiny{$\pm1.80$} & $70.62$\tiny{$\pm1.02$} & $83.12$\tiny{$\pm1.31$} & \hlcella $63.13$\tiny{$\pm1.42$} & \hlcella $60.62$\tiny{$\pm1.39$} & $78.37$\tiny{$\pm1.18$} & $71.63$\tiny{$\pm0.80$} & $71.63$\tiny{$\pm0.71$} & \hlcella $50.96$\tiny{$\pm1.47$} & \hlcella $63.46$\tiny{$\pm1.26$} \\
        Nemotron-7B & $67.50$\tiny{$\pm1.28$} & $64.38$\tiny{$\pm0.77$} & $71.25$\tiny{$\pm1.21$} & \hlcella $58.75$\tiny{$\pm1.80$} & \hlcella $51.88$\tiny{$\pm1.25$} & $67.79$\tiny{$\pm1.12$} & $48.56$\tiny{$\pm0.71$} & $75.96$\tiny{$\pm0.97$} & \hlcella $37.02$\tiny{$\pm1.29$} & \hlcella $41.83$\tiny{$\pm1.28$} \\
        Qwen3-32B & $65.00$\tiny{$\pm2.05$} & $35.62$\tiny{$\pm1.34$} & $60.62$\tiny{$\pm1.02$} & \hlcella $30.00$\tiny{$\pm1.55$} & \hlcella $47.50$\tiny{$\pm1.69$} & $54.81$\tiny{$\pm1.47$} & $38.46$\tiny{$\pm0.98$} & $58.65$\tiny{$\pm0.73$} & \hlcella $31.25$\tiny{$\pm0.78$} & \hlcella $30.29$\tiny{$\pm1.01$} \\
        Qwen3-8B & $55.62$\tiny{$\pm2.12$} & $34.38$\tiny{$\pm0.94$} & $65.62$\tiny{$\pm1.22$} & \hlcella $20.62$\tiny{$\pm1.69$} & \hlcella $24.38$\tiny{$\pm1.22$} & $45.67$\tiny{$\pm1.20$} & $15.38$\tiny{$\pm0.64$} & $43.27$\tiny{$\pm0.97$} & \hlcella $11.06$\tiny{$\pm1.04$} & \hlcella $15.38$\tiny{$\pm1.33$} \\
    \midrule
     & \multicolumn{5}{c}{\hmmt{24}} & \multicolumn{5}{c}{\hmmt{25}} \\
     \cmidrule(lr){2-6} \cmidrule(lr){7-11}
     & {\direct} & {\revise} & {\filter} & {\hlcella {\twof}} & {\hlcella {\twof}-{\iv}} & {\direct} & {\revise} & {\filter} & {\hlcella {\twof}} & {\hlcella {\twof}-{\iv}} \\
    \midrule
        GPT-OSS-120B & $48.83$\tiny{$\pm1.22$} & $59.38$\tiny{$\pm0.38$} & $48.05$\tiny{$\pm1.14$} & \hlcella $45.31$\tiny{$\pm1.06$} & \hlcella $45.31$\tiny{$\pm0.99$} & $61.01$\tiny{$\pm0.70$} & $59.82$\tiny{$\pm0.51$} & $57.44$\tiny{$\pm0.66$} & \hlcella $49.11$\tiny{$\pm0.69$} & \hlcella $55.36$\tiny{$\pm0.82$} \\
        GPT-OSS-20B & $40.23$\tiny{$\pm1.08$} & $57.03$\tiny{$\pm1.11$} & $50.39$\tiny{$\pm1.10$} & \hlcella $17.97$\tiny{$\pm1.21$} & \hlcella $35.94$\tiny{$\pm0.88$} & $58.63$\tiny{$\pm0.78$} & $56.85$\tiny{$\pm0.66$} & $52.68$\tiny{$\pm0.77$} & \hlcella $24.70$\tiny{$\pm1.02$} & \hlcella $44.05$\tiny{$\pm0.89$} \\
        Nemotron-32B & $61.72$\tiny{$\pm1.40$} & $59.77$\tiny{$\pm0.47$} & $67.97$\tiny{$\pm1.01$} & \hlcella $50.78$\tiny{$\pm0.85$} & \hlcella $47.66$\tiny{$\pm0.89$} & $80.36$\tiny{$\pm0.62$} & $76.19$\tiny{$\pm0.44$} & $79.46$\tiny{$\pm0.50$} & \hlcella $64.58$\tiny{$\pm0.70$} & \hlcella $62.50$\tiny{$\pm0.69$} \\
        Nemotron-7B & $49.22$\tiny{$\pm1.15$} & $56.64$\tiny{$\pm0.99$} & $57.42$\tiny{$\pm0.87$} & \hlcella $40.62$\tiny{$\pm0.76$} & \hlcella $32.42$\tiny{$\pm0.95$} & $63.10$\tiny{$\pm0.95$} & $68.45$\tiny{$\pm0.30$} & $75.30$\tiny{$\pm0.45$} & \hlcella $52.38$\tiny{$\pm0.74$} & \hlcella $49.11$\tiny{$\pm0.78$} \\
        Qwen3-32B & $39.06$\tiny{$\pm1.34$} & $29.30$\tiny{$\pm0.75$} & $33.20$\tiny{$\pm0.80$} & \hlcella $40.23$\tiny{$\pm0.76$} & \hlcella $26.17$\tiny{$\pm0.82$} & $49.70$\tiny{$\pm0.83$} & $47.02$\tiny{$\pm0.50$} & $50.60$\tiny{$\pm0.62$} & \hlcella $41.37$\tiny{$\pm0.67$} & \hlcella $34.23$\tiny{$\pm0.81$} \\
        Qwen3-8B & $21.88$\tiny{$\pm0.72$} & $26.95$\tiny{$\pm0.45$} & $25.78$\tiny{$\pm0.46$} & \hlcella $26.95$\tiny{$\pm0.89$} & \hlcella $18.36$\tiny{$\pm0.96$} & $42.26$\tiny{$\pm0.67$} & $40.48$\tiny{$\pm0.44$} & $44.35$\tiny{$\pm0.78$} & \hlcella $32.74$\tiny{$\pm0.67$} & \hlcella $32.74$\tiny{$\pm0.72$} \\
    \midrule
     & \multicolumn{5}{c}{\textsc{\gpqa}} & \multicolumn{5}{c}{\textsc{\mmlu}} \\
     \cmidrule(lr){2-6} \cmidrule(lr){7-11}
     & {\direct} & {\revise} & {\filter} & {\hlcella {\twof}} & {\hlcella {\twof}-{\iv}} & {\direct} & {\revise} & {\filter} & {\hlcella {\twof}} & {\hlcella {\twof}-{\iv}} \\
    \midrule
        GPT-OSS-120B & $58.24$\tiny{$\pm0.13$} & $43.56$\tiny{$\pm0.11$} & $53.17$\tiny{$\pm0.15$} & \hlcella $31.91$\tiny{$\pm0.09$} & \hlcella $31.20$\tiny{$\pm0.08$} & $67.59$\tiny{$\pm0.02$} & $46.67$\tiny{$\pm0.02$} & $63.43$\tiny{$\pm0.02$} & \hlcella $31.61$\tiny{$\pm0.03$} & \hlcella $30.12$\tiny{$\pm0.02$} \\
        GPT-OSS-20B & $46.97$\tiny{$\pm0.14$} & $30.87$\tiny{$\pm0.12$} & $42.28$\tiny{$\pm0.18$} & \hlcella $13.16$\tiny{$\pm0.08$} & \hlcella $20.27$\tiny{$\pm0.14$} & $58.86$\tiny{$\pm0.04$} & $35.19$\tiny{$\pm0.01$} & $53.71$\tiny{$\pm0.02$} & \hlcella $12.83$\tiny{$\pm0.02$} & \hlcella $14.95$\tiny{$\pm0.03$} \\
        Nemotron-32B & $66.81$\tiny{$\pm0.08$} & $46.64$\tiny{$\pm0.05$} & $63.87$\tiny{$\pm0.05$} & \hlcella $37.22$\tiny{$\pm0.09$} & \hlcella $34.04$\tiny{$\pm0.09$} & $67.25$\tiny{$\pm0.02$} & $30.71$\tiny{$\pm0.01$} & $67.43$\tiny{$\pm0.01$} & \hlcella $20.09$\tiny{$\pm0.02$} & \hlcella $17.03$\tiny{$\pm0.02$} \\
        Nemotron-7B & $51.61$\tiny{$\pm0.11$} & $35.80$\tiny{$\pm0.05$} & $49.72$\tiny{$\pm0.08$} & \hlcella $29.40$\tiny{$\pm0.07$} & \hlcella $27.56$\tiny{$\pm0.06$} & $56.46$\tiny{$\pm0.02$} & $21.10$\tiny{$\pm0.01$} & $51.56$\tiny{$\pm0.01$} & \hlcella $14.02$\tiny{$\pm0.02$} & \hlcella $11.63$\tiny{$\pm0.02$} \\
        Qwen3-32B & $57.62$\tiny{$\pm0.10$} & $26.09$\tiny{$\pm0.03$} & $31.72$\tiny{$\pm0.07$} & \hlcella $23.30$\tiny{$\pm0.10$} & \hlcella $20.08$\tiny{$\pm0.08$} & $70.62$\tiny{$\pm0.02$} & $26.58$\tiny{$\pm0.01$} & $66.93$\tiny{$\pm0.02$} & \hlcella $21.97$\tiny{$\pm0.02$} & \hlcella $14.69$\tiny{$\pm0.01$} \\
        Qwen3-8B & $47.77$\tiny{$\pm0.09$} & $26.42$\tiny{$\pm0.03$} & $46.59$\tiny{$\pm0.06$} & \hlcella $20.45$\tiny{$\pm0.06$} & \hlcella $17.99$\tiny{$\pm0.06$} & $61.65$\tiny{$\pm0.03$} & $28.15$\tiny{$\pm0.01$} & $57.71$\tiny{$\pm0.02$} & \hlcella $17.21$\tiny{$\pm0.01$} & \hlcella $12.63$\tiny{$\pm0.01$} \\
    \midrule
     & \multicolumn{5}{c}{\textsc{\crux}} & \multicolumn{5}{c}{\textsc{\game}} \\
     \cmidrule(lr){2-6} \cmidrule(lr){7-11}
     & {\direct} & {\revise} & {\filter} & {\hlcella {\twof}} & {\hlcella {\twof}-{\iv}} & {\direct} & {\revise} & {\filter} & {\hlcella {\twof}} & {\hlcella {\twof}-{\iv}} \\
    \midrule
        GPT-OSS-120B & $89.23$\tiny{$\pm0.08$} & $75.96$\tiny{$\pm0.09$} & $79.06$\tiny{$\pm0.12$} & \hlcella $77.89$\tiny{$\pm0.06$} & \hlcella $83.54$\tiny{$\pm0.09$} & $77.34$\tiny{$\pm0.03$} & $72.27$\tiny{$\pm0.02$} & $78.81$\tiny{$\pm0.03$} & \hlcella $57.09$\tiny{$\pm0.05$} & \hlcella $57.78$\tiny{$\pm0.04$} \\
        GPT-OSS-20B & $61.34$\tiny{$\pm0.28$} & $61.90$\tiny{$\pm0.13$} & $44.68$\tiny{$\pm0.18$} & \hlcella $40.53$\tiny{$\pm0.12$} & \hlcella $44.18$\tiny{$\pm0.12$} & $78.30$\tiny{$\pm0.03$} & $56.07$\tiny{$\pm0.04$} & $76.37$\tiny{$\pm0.05$} & \hlcella $39.31$\tiny{$\pm0.06$} & \hlcella $40.02$\tiny{$\pm0.07$} \\
        Nemotron-32B & $77.06$\tiny{$\pm0.09$} & $72.61$\tiny{$\pm0.07$} & $71.34$\tiny{$\pm0.06$} & \hlcella $54.19$\tiny{$\pm0.09$} & \hlcella $51.80$\tiny{$\pm0.06$} & $79.54$\tiny{$\pm0.03$} & $74.41$\tiny{$\pm0.02$} & $81.35$\tiny{$\pm0.02$} & \hlcella $51.53$\tiny{$\pm0.03$} & \hlcella $50.63$\tiny{$\pm0.03$} \\
        Nemotron-7B & $43.55$\tiny{$\pm0.20$} & $47.91$\tiny{$\pm0.07$} & $38.36$\tiny{$\pm0.06$} & \hlcella $31.32$\tiny{$\pm0.09$} & \hlcella $36.10$\tiny{$\pm0.12$} & $75.35$\tiny{$\pm0.02$} & $70.55$\tiny{$\pm0.01$} & $73.62$\tiny{$\pm0.02$} & \hlcella $34.77$\tiny{$\pm0.02$} & \hlcella $39.11$\tiny{$\pm0.04$} \\
        Qwen3-32B & $75.83$\tiny{$\pm0.08$} & $69.32$\tiny{$\pm0.04$} & $62.23$\tiny{$\pm0.07$} & \hlcella $50.90$\tiny{$\pm0.08$} & \hlcella $51.66$\tiny{$\pm0.09$} & $78.48$\tiny{$\pm0.03$} & $56.99$\tiny{$\pm0.02$} & $78.18$\tiny{$\pm0.03$} & \hlcella $25.47$\tiny{$\pm0.03$} & \hlcella $41.54$\tiny{$\pm0.03$} \\
        Qwen3-8B & $67.49$\tiny{$\pm0.07$} & $53.86$\tiny{$\pm0.04$} & $49.80$\tiny{$\pm0.06$} & \hlcella $36.20$\tiny{$\pm0.08$} & \hlcella $36.20$\tiny{$\pm0.06$} & $76.29$\tiny{$\pm0.03$} & $43.78$\tiny{$\pm0.01$} & $76.21$\tiny{$\pm0.01$} & \hlcella $23.26$\tiny{$\pm0.02$} & \hlcella $29.24$\tiny{$\pm0.03$} \\
    \bottomrule\bottomrule
\end{tabular}
\end{adjustbox}
    
    \label{tab:context_manipulation_full_2_drafts}
\end{table}

\newpage
\section{Additional Information for Supervised Fine-tuning Experiments}
\label{app:training-details}

\begin{figure}[H]
    \centering
    \includegraphics[width=0.95\linewidth]{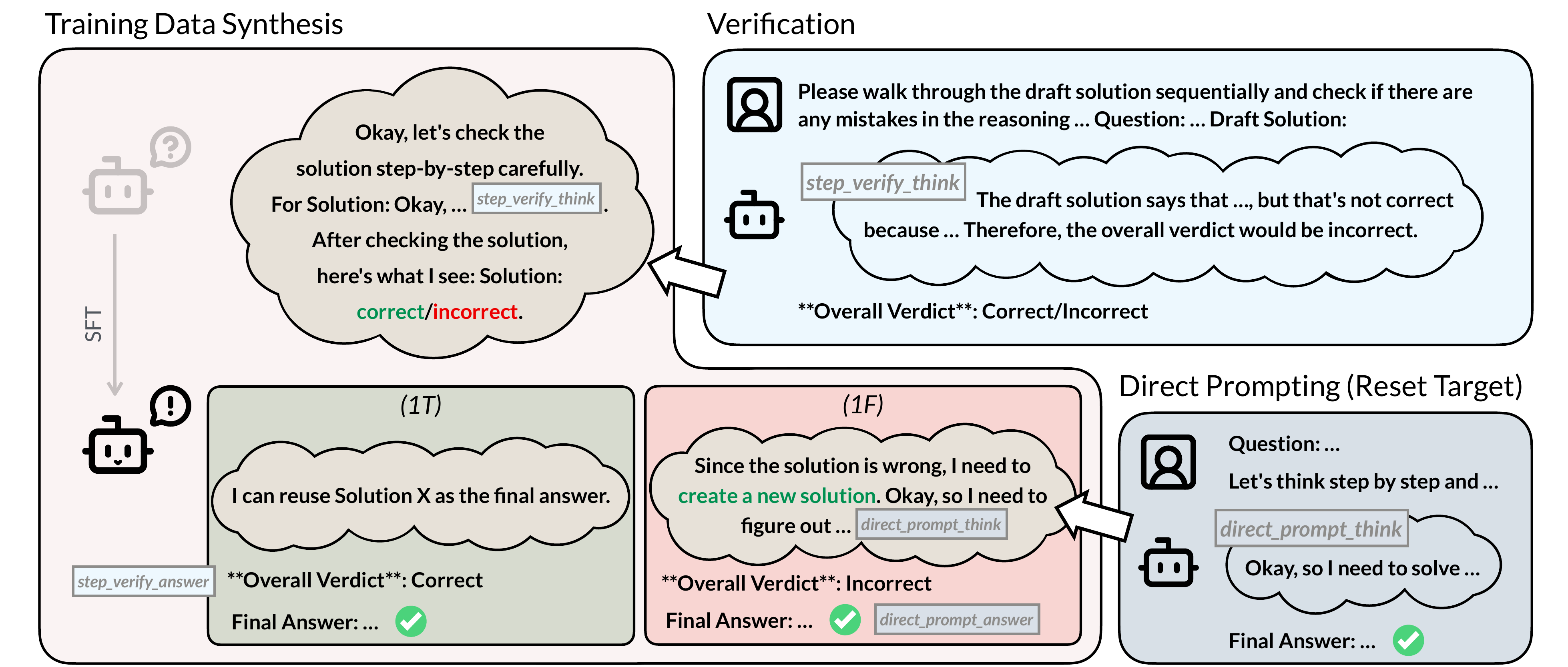}
    \caption{\textbf{Overview of the proposed training procedure}: Synthesized reasoning traces teach the model to verify in-context drafts before solving. If the model identifies the draft as incorrect, as in the case of 1F shown above, it is trained to ignore the revert to a reasoning trajectory samples from clean-slate generation. If the model verified the draft to be correct, it is trained to reuse the answer in draft.}
    \label{fig:mitigation-datasynth}
\end{figure}

In this section, we provide more details on the data curation and training setup for the supervised fine-tuning experiment described in \cref{sec:training}.

\subsection{Question Selection from BigMathRL}
\label{app:question_selection}

To obtain a set of questions on which we can achieve reliable verification, we draw problems from BigMathRL \cite{albalak2025bigmathlargescalehighqualitymath} and restrict to a reasonably hard subset using an external solver filter: we retain only questions whose solve rate under Llama3.1-8B (provided in the dataset) is below $0.5$. From this filtered pool, we subsample $40{,}000$ questions with a fixed source mixture across the underlying BigMathRL data sources.

We report the data composition of the question pool in \cref{tab:bigmathrl_source_mix}. We conducted balanced subsampling such that for small sources, we include the full available set.

\begin{table}[H]
\centering
\small
\caption{Source mixture of the $40\text{k}$ question subset used for synthesis. ``Available'' denotes the number of filtered questions per source after applying the Llama 3.1 8B solve-rate $<0.5$ criterion.}
\begin{tabular}{lrrr}
\toprule
\toprule
\textbf{Source} & \textbf{Available} & \textbf{Sampled} & \textbf{Sample rate} \\
\midrule
cn\_k12      & $37{,}228$ & $6{,}729$ & $18.1\%$ \\
orca\_math   & $27{,}153$ & $6{,}728$ & $24.8\%$ \\
olympiads    & $29{,}941$ & $6{,}728$ & $22.5\%$ \\
aops\_forum  & $5{,}067 $ & $5{,}067$ & $100.0\%$ \\
gsm8k        & $159$      & $159$     & $100.0\%$ \\
amc\_aime    & $71$       & $71$      & $100.0\%$ \\
math         & $2{,}780$  & $2{,}780$ & $100.0\%$ \\
omnimath     & $2{,}033$  & $2{,}033$ & $100.0\%$ \\
openmath     & $281$      & $281$     & $100.0\%$ \\
harp         & $2{,}696$  & $2{,}696$ & $100.0\%$ \\
big\_math    & $27{,}094$ & $6{,}728$ & $24.8\%$ \\
\midrule
\textbf{Total} & $-$ & \textbf{$40{,}000$} & $-$ \\
\bottomrule
\end{tabular}
\label{tab:bigmathrl_source_mix}
\end{table}

\subsection{Draft Generation and Verifiable Correctness}
\label{app:draft_generation}

For each selected question, we sample candidate draft solutions from a family of open-weight reasoning models: GPT-OSS-20B (with medium thinking effort), Qwen3-8B (with thinking), OpenReasoning-Nemotron-7B, LlamaR1-8B, and QwenR1-7B. Concretely, we sample $4$ independent {\direct} responses per (question, model) pair, yielding up to $20$ drafts per question. We then determine draft correctness using the \texttt{math\_verify} package \cite{kydlicek2024mathverify}, which provides verifiable answer checking for math problems.

\subsection{Verifier Trajectory Curation}
\label{app:verifier_generation}

To obtain supervision for verification behavior, we prompt GPT-OSS-20B (our target model for training) to verify each draft solution with the following prompt template, producing a verdict and accompanying verification rationale. 

\begin{promptbox}{template-training-verification}{Verification Prompt: Training Data Curation}
~\\
In this task you will be given a question and a draft solution to that question including the thinking steps. Please walk through the draft solution sequentially and check if there are any fatal mistakes in the reasoning (e.g. incorrect use of theorems / knowledge, incorrect calculations, etc.). Please note that once you spot an fatal error reasoning process, you **MUST** stop checking the remaining steps and directly output the verdict. **DO NOT ATTEMPT TO RESOLVE THE QUESTION**. If you can find a potential fix to the fatal step, you can also output it, but you **MUST NOT** attempt to further resolve the question.  If all steps appears to be correct, you should check the entire draft. In the end you must output an overall verdict to the correctness of the provided draft in the form of:

\emph{}

<overall\_verdict>[Correct / Incorrect]</overall\_verdict>

<overall\_confidence>[High / Medium / Low]</overall\_confidence>

<fix>[Any potential fix to the incorrect step if there is any]</fix>

\emph{}

Please be strict with algebra, logic, and theorem use. Please only verify and provide feedback to the provided draft. You **MUST NOT** attempt to re-solving the problem.

\emph{}

The question provided is:

-- beginning of question --

\textbf{\{problem\}}

-- end of question --

\emph{}

Please follow the instructions above and examine the draft solution:

-- beginning of draft --

\textbf{\{draft\}}

-- end of draft --

\emph{}

Please note that once you spot a fatal error reasoning process, you **MUST** stop checking the remaining steps and directly output the verdict. If all steps appears to be correct, you should check the entire draft. **DO NOT ATTEMPT TO RESOLVE THE QUESTION**
\end{promptbox}

The prompt above is designed such that we can accurately parse the model's final verification verdict wrapped between \texttt{<overall\_verdict>} and \texttt{<\textbackslash overall\_verdict>}.

We compare the model's verification verdict with the symbolic verifier's result and retain only verification trajectories that agree with the verifier outcome. This produces a set of correct verification trajectories for both correct and incorrect drafts.

\subsection{Balanced training set construction}
\label{app:training_pairs}

We construct the SFT dataset by sampling:
(i) $40{,}000$ {(question, correct draft, correct verification)} triples and
(ii) $40{,}000$ {(question, incorrect draft, correct verification)} triples.
This yields $80{,}000$ total training examples with a balanced mixture of correct / incorrect drafts.
Each example, depending on whether it contains a correct or incorrect draft, is formatted using a dedicated template (as shown below). \texttt{verification\_reasoning\_trace} is the reasoning trace collected in \cref{app:verifier_generation} under the verification prompt, and \texttt{clean\_slate\_reasoning\_trace} is the reasoning trace generated by GPT-OSS-20B when correctly solving the problem under the {\direct} setting with no additional context.
\
\begin{promptbox}{template-training-reuse}{Synthetic Reasoning Trace: Correct Draft}
~\\
<|channel|>analysis<|message|>We need to check the draft. If it is correct, I can reuse it as the final answer. If it is wrong, I need to come up with a correct solution myself, reusing any useful pieces if possible. \textbf{\{verification\_reasoning\_trace\}}\\

We can see that the draft is \textbf{Correct}. I can reuse the draft solution as the final answer.\\
<|end|><|start|>assistant<|channel|>final<|message|><overall\_verdict\_1>

\textbf{Correct}</overall\_verdict\_1>\\

The final answer is:\\

\textbf{\{final\_answer\}}

\end{promptbox}

\begin{promptbox}{template-training-reset}{Synthetic Reasoning Trace: Incorrect Draft}
~\\
<|channel|>analysis<|message|>We need to check the draft. If it is correct, I can reuse it as the final answer. If it is wrong, I need to come up with a correct solution myself, reusing any useful pieces if possible. \textbf{\{verification\_reasoning\_trace\}}\\

We can see that the draft is \textbf{Incorrect}. Since the draft is not correct, I need to create a new solution. I need to think about this problem again.\\

\textbf{\{clean\_slate\_reasoning\_trace\}}\\

<|end|><|start|>assistant<|channel|>final<|message|><overall\_verdict\_1>

\textbf{Incorrect}</overall\_verdict\_1>\\

The final answer is:\\

\textbf{\{final\_answer\}}
\end{promptbox}

\subsection{Training setup}
\label{app:sft_setup}

We perform full supervised fine-tuning for one epoch on the synthesized dataset using cosine learning rate decay with a peak learning rate of $5\times 10^{-5}$ and a $10\%$ warmup fraction. Training was conducted on $8\times$H100 nodes with Llama-factory \cite{zheng2024llamafactory}.

\subsection{Additional Results: {\oneT} Evaluation After SFT}
\label{app:sft_results_1T}

In \cref{sec:training} we fine-tune GPT-OSS-20B to (i) produce accurate verification trajectories and (ii) ``reset'' to its {\direct} reasoning trace when verification indicates that the in-context draft is incorrect. Here we report additional evaluation results on the {\oneT} setting (when there is a correct draft in context) after training.
We evaluate the trained model using the same set of questions and draft-construction pipeline as in our main {\oneT} and {\direct} evaluations, with the only exception being that the in-context draft is sampled to be a correct, as determined by the symbolic verifier.

\begin{table}[H]
    \centering
    \small
    \caption{\textbf{Targeted SFT decreases \oneT{} performance.} Pass@1 accuracy on competitive math benchmarks under \oneT{} (verify-then-answer) for GPT-OSS-20B baseline and our finetuned checkpoint. Deltas in parentheses are absolute percentage-point changes vs.\ baseline.}
\begin{tabular}{l|cccc}
\toprule
\toprule
\textsc{Setting} & \textsc{\aime{24}} & \textsc{\aime{25}} & \textsc{\hmmt{24}} & \textsc{\hmmt{25}} \\
\midrule
\multicolumn{5}{l}{\textsc{1T (Pass@$1$)}} \\
\midrule
Baseline  & $73.12$ & $73.56$ & $70.31$ & $76.79$ \\
Finetuned & $50.63$ (\perfdecrease{-$22.49$}) & $49.04$ (\perfdecrease{-$24.52$}) & $44.92$ (\perfdecrease{-$25.39$}) & $49.11$ (\perfdecrease{-$27.68$}) \\
\midrule
\multicolumn{5}{l}{\textsc{1T (Pass@$5$)}} \\
\midrule
Baseline  & $100.00$ & $94.71$ & $95.31$ & $97.02$ \\
Finetuned & $83.75$ (\perfdecrease{-$16.25$}) & $90.87$ (\perfdecrease{-$3.84$}) & $85.94$ (\perfdecrease{-$9.37$}) & $85.42$ (\perfdecrease{-$11.60$}) \\
\bottomrule
\end{tabular}
    \label{tab:sft_result_1t}
\end{table}

\cref{tab:sft_result_1t} shows that while our targeted SFT improves robustness in the {\onef} setting (\cref{sec:training}), it substantially reduces {\oneT} pass@$1$ and pass@$5$ performance across benchmarks. A plausible explanation is a \emph{robustness-utilization tradeoff}: training the model to ``reset'' after error detection may also make it less able to capitalize on useful information in the provided draft, even when that context is partially or fully correct. In {\oneT}, this can manifest as overly conservative behavior (discarding potentially helpful structure) or increased token/attention budget spent on verification-style trajectories rather than constructive problem solving, leading to lower pass rates. Together, these results suggest targeted SFT is a proof-of-concept mitigation that improves robustness to incorrect context but is not yet a complete intervention for settings where correct context utilization is important.

\subsection{Training with GRPO}
\label{sec:apx-grpo}

In addition to the curated supervised fine-tuning (SFT) recipe in
\cref{sec:training}, we also experimented with standard Group Relative Policy
Optimization (GRPO; \citealt{shao2024deepseekmath}) as a post-training alternative.
We train on the \emph{same} set of math problems used for SFT, but optimize
only a final-answer correctness reward. 

Concretely, for each prompt we
sample a group of $n{=}8$ rollouts from the current policy, compute a binary
reward based on whether the final answer matches the ground truth, and apply the
GRPO update. We use batchsize $64$,
maximum context length $16{,}384$, and otherwise follow standard GRPO defaults
(e.g., on-policy sampling, per-step KL regularization to the reference policy,
and reward normalization within each group).

\begin{table}[H]
    \centering
    \caption{\textbf{GRPO results on competitive math under contextual drag (1F).}
    We report accuracy (\%) in the 1F setting. ``Finetuned'' denotes our targeted
    SFT model (Section~\ref{sec:training}); ``GRPO'' denotes the model trained with
    standard GRPO. Deltas in parentheses are absolute percentage-point changes
    relative to the \textsc{Baseline} in the same block.}
    \label{tab:grpo_results}
    \vspace{0.25em}
    \begin{adjustbox}{width=0.6\linewidth}
    \begin{tabular}{lcccc}
        \toprule
        \textsc{Setting} & \textsc{\aime{24}} & \textsc{\aime{25}} & \textsc{\hmmt{24}} & \textsc{\hmmt{25}} \\
        \midrule
        \multicolumn{5}{l}{\textsc{1F}} \\
        \midrule
        Baseline  & $17.5$ & $18.8$ & $13.3$ & $21.2$ \\
        Finetuned & $40.6$ (\perfincrease{+$23.1$}) & $20.7$ (\perfincrease{+$1.9$})  & $19.2$ (\perfincrease{+$5.9$}) & $27.8$ (\perfincrease{+$6.6$}) \\
        GRPO      & $22.5$ (\perfincrease{+$5.0$})  & $20.7$ (\perfincrease{+$1.9$})  & $17.6$ (\perfincrease{+$4.3$}) & $24.4$ (\perfincrease{+$3.2$}) \\
        \midrule
        \multicolumn{5}{l}{\textsc{1F (Conditioned on Self-Detected Error Signal)}} \\
        \midrule
        Baseline  & $28.5$ & $43.3$ & $16.4$ & $33.9$ \\
        Finetuned & $52.9$ (\perfincrease{+$24.4$}) & $38.7$ (\perfdecrease{-$4.6$}) & $30.5$ (\perfincrease{+$14.1$}) & $45.3$ (\perfincrease{+$11.4$}) \\
        \bottomrule
    \end{tabular}
    \end{adjustbox}
\end{table}

\paragraph{Discussion.}
\cref{tab:grpo_results} shows that standard GRPO yields  modest gains under
contextual drag, improving over the baseline by $+1.9$ to $+5.0$ points across
benchmarks, and substantially underperforming targeted SFT.
This gap is consistent with GRPO’s reward being defined solely on final-answer
correctness: while it can nudge the policy toward better outcomes on average, it
does not explicitly teach the model to \emph{reset} its reasoning trajectory after
detecting errors in the in-context draft.
In contrast, the supervised objective directly trains the post-verification
trajectory to match the \textsc{Direct} reasoning trace, which appears more aligned
with mitigating contextual drag.
Overall, these results suggest that na\"ively applying outcome-based RL is not a
drop-in replacement for targeted supervision in this setting, and that more
structured rewards (e.g., verifier-aware or trajectory-level signals) may be
needed for RL-style training to reliably counteract erroneous context.
%%%%%%%%%%%%%%%%%%%%%%%%%%%%%%%%%%%%%%%%%%%%%%%%%%%%%%%%%%%%%%%%%%%%%%%%%%%%%%%
%%%%%%%%%%%%%%%%%%%%%%%%%%%%%%%%%%%%%%%%%%%%%%%%%%%%%%%%%%%%%%%%%%%%%%%%%%%%%%%

\end{document}